\documentclass[]{fairmeta}

\usepackage{lmodern}
\usepackage{amsmath,amssymb}
\usepackage{pifont}
\usepackage{algorithm}
\usepackage{wrapfig}
\usepackage{algorithmic}
\usepackage{caption}
\usepackage{listings}
\usepackage{url}

\usepackage[table]{xcolor}

\definecolor{BuilderBG}{HTML}{F5F6F8}
\definecolor{DesignerBG}{HTML}{EEF4FC}
\definecolor{CriticBG}{HTML}{E4EFFB}
\definecolor{GainBG}{HTML}{DEF3E7}

\definecolor{DeltaGreen}{HTML}{146C43}
\definecolor{DeltaRed}{HTML}{B42318}

\newcommand{\upchg}[1]{%
    \textsuperscript{\textcolor{DeltaGreen}{\tiny +#1}}%
}

\newcommand{\gaincell}[1]{\cellcolor{GainBG}#1}

\newcommand{\cmark}{\ding{51}}
\newcommand{\xmark}{\ding{55}}

\title{Scene-Designer: Learning Design Intelligence for Agentic 3D Scene Generation}
\author[1,*]{Xuanyu~Liu}
\author[1,*]{Hongyan~Feng}
\author[1]{Miao~Pan}
\author[1]{Sunlai~Chen}
\author[1]{Lizhou~Liang}
\author[2]{Bing~Sun}
\author[1]{Wenqi~Zhang}
\author[1]{Xuhong~Zhang}
\affiliation[1]{Zhejiang University\protect\\}
\affiliation[2]{Universal Identification Technology (Hangzhou) Co.,Ltd.}
\contribution[*]{Equal contribution.}
\newcommand{\resourceicon}[1]{\raisebox{-0.18em}{\includegraphics[height=1.05em]{resource/#1.pdf}}\hspace{0.45em}}
\renewcommand{\metadataformat}[2][]{{\small\strut{\sffamily\bfseries #1:}\hspace{0.35em}#2}}
\metadata[\protect\resourceicon{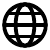}Project Page]{\protect\url{https://zju-omniai.github.io/Scene-Designer/}}
\metadata[\protect\resourceicon{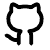}GitHub]{\protect\url{https://github.com/ZJU-OmniAI/Scene-Designer}}
\metadata[\protect\resourceicon{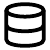}Hugging Face]{\protect\url{https://huggingface.co/XuanyuLiu}}
\metadata[\protect\resourceicon{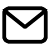}Email]{\protect\email{22651306@zju.edu.cn}}

\abstract{Generating complete, coherent, and usable 3D indoor scenes from natural language requires strong \textbf{design intelligence}: the ability to design high-quality scenes and critique their quality. However, existing approaches either rely on general-purpose models lacking domain-specific design experience or learn from static scenes. In this work, we introduce \textbf{Scene-Designer}, an agentic framework that learns design intelligence through complementary design and critique experiences. We construct two datasets: \textbf{Scene-OTA-35K}, which provides Observation--Thought--Action supervision for learning scene design and efficient error correction, and \textbf{Scene-OTC-15K}, which provides Observation--Thought--Critic supervision for learning scene quality critique across physical validity, functional correctness, and layout coherence. Based on these experiences, we propose \textbf{Scene-GRPO}, a Critic-guided GRPO framework that improves design decisions and reduces unnecessary trial-and-error corrections. Extensive experiments demonstrate that Scene-Designer achieves superior performance in physical validity, functional correctness, and layout coherence, while the learned Scene-Critic significantly outperforms general-purpose models in scene critique. Moreover, test-time Critic guidance further improves scene generation, showing that the Critic can provide useful guidance beyond training. Additional analysis suggests a trend of mutual enhancement between scene design and critique abilities.}

\begin{document}
\thispagestyle{firstheader}
\maketitle
\pagestyle{empty}

% Hide main-text headings in the appendix-only table of contents.
\addtocontents{toc}{\protect\setcounter{tocdepth}{-1}}
\section{Introduction}

\begin{tcolorbox}[
    colback=blue!3!white,
    colframe=blue!45!gray,
    boxrule=0.5pt,
    arc=2mm,
    left=2.5mm, right=2.5mm,
    top=2mm, bottom=2mm,
    before skip=6pt, after skip=10pt
]
\textit{``The clearer one's understanding becomes, the more steadfastly one acts;
the more steadfastly one acts, the clearer one's understanding becomes.''}
\unskip\nobreak\hfill\penalty50\hskip1em\null\nobreak\hfill
\mbox{\textit{--- Zhu Xi, Zhuzi yulei, vol.~14.}}\par
\end{tcolorbox}

\begin{figure*}[t]
    \centering
    \includegraphics[width=\textwidth]{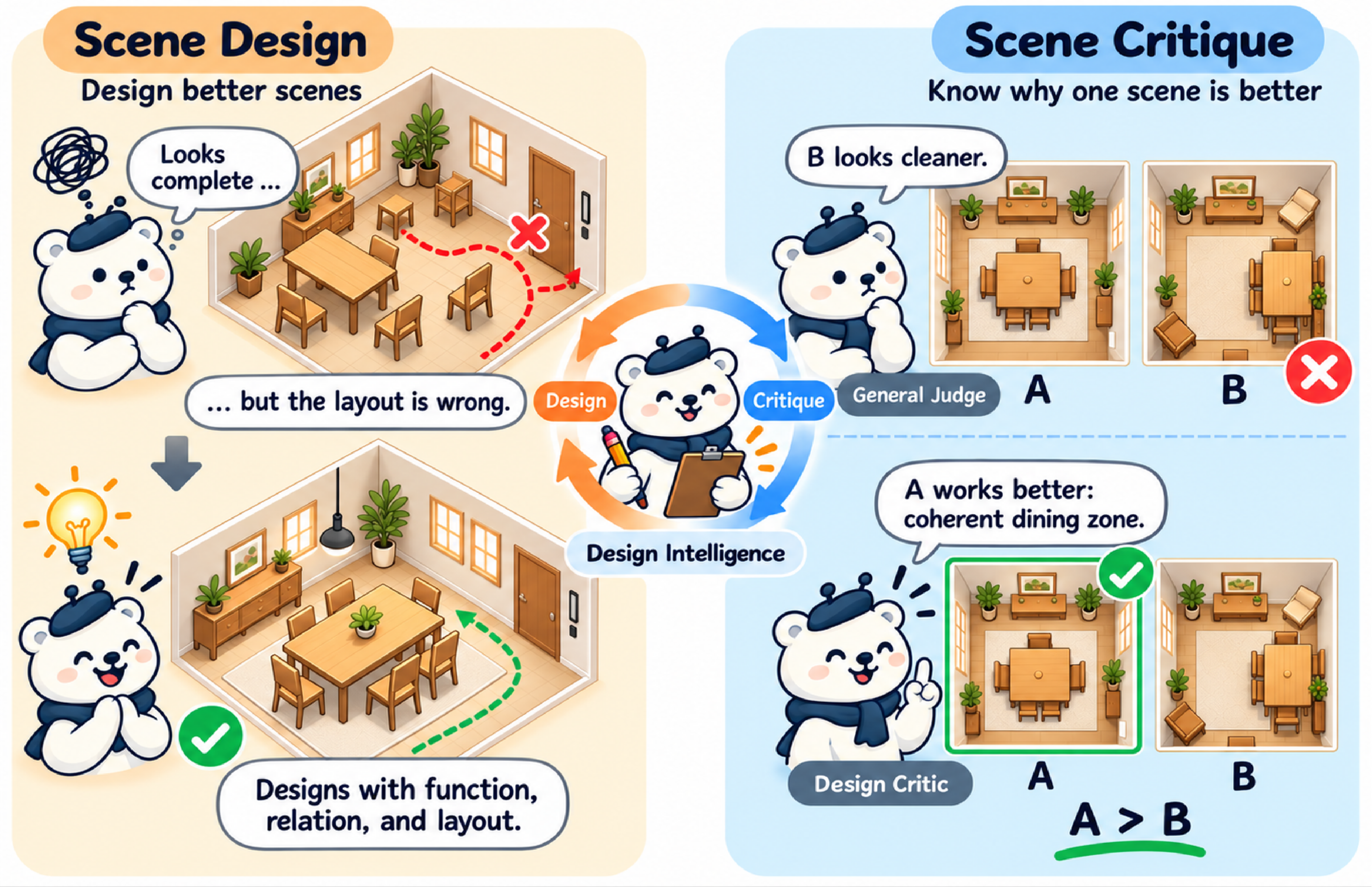}
    \caption{
    Overview of Design Intelligence.
    We define design intelligence as the ability to both \textbf{design} a scene and \textbf{critique} its quality.
    It consists of two complementary capabilities:
    \textbf{scene design}, which constructs physically valid, functionally correct, and layout-coherent scenes while making effective corrections,
    and \textbf{scene critique}, which assesses scene quality along the same dimensions and identifies better designs.
}
    \label{fig:teaser}
\end{figure*}
Automatically generating complete, coherent, and usable 3D scenes from natural-language instructions is becoming increasingly important with the rapid progress of embodied AI, automated spatial design, virtual content creation, and related applications. At its core, high-quality scene generation is not merely a problem of placing objects, but a problem of design. This calls for a model to reason both about how a scene should take shape and about the quality of the resulting scene. We refer to this capability as \textbf{Design Intelligence}.

We define design intelligence as the ability to both \textbf{design} a scene and \textbf{critique} its quality. These two capabilities correspond to two essential stages of the design process: making design decisions and evaluating their quality. \textbf{Scene design} concerns constructing physically valid, functionally correct, and layout-coherent scenes, while efficiently correcting errors without unnecessary trial and error when problems arise. \textbf{Scene critique} focuses on assessing scene quality along the same dimensions and distinguishing better designs from poorer ones. These two capabilities are closely connected, since effective design requires reliable critique to guide construction, while meaningful critique requires an understanding of what constitutes good scene design. However, even the strongest general-purpose models still exhibit clear limitations in both capabilities, as revealed by the following diagnostic studies.

\begin{figure}[!t]
    \centering
    \begin{minipage}{0.42\columnwidth}
        \centering
        \includegraphics[width=\linewidth]{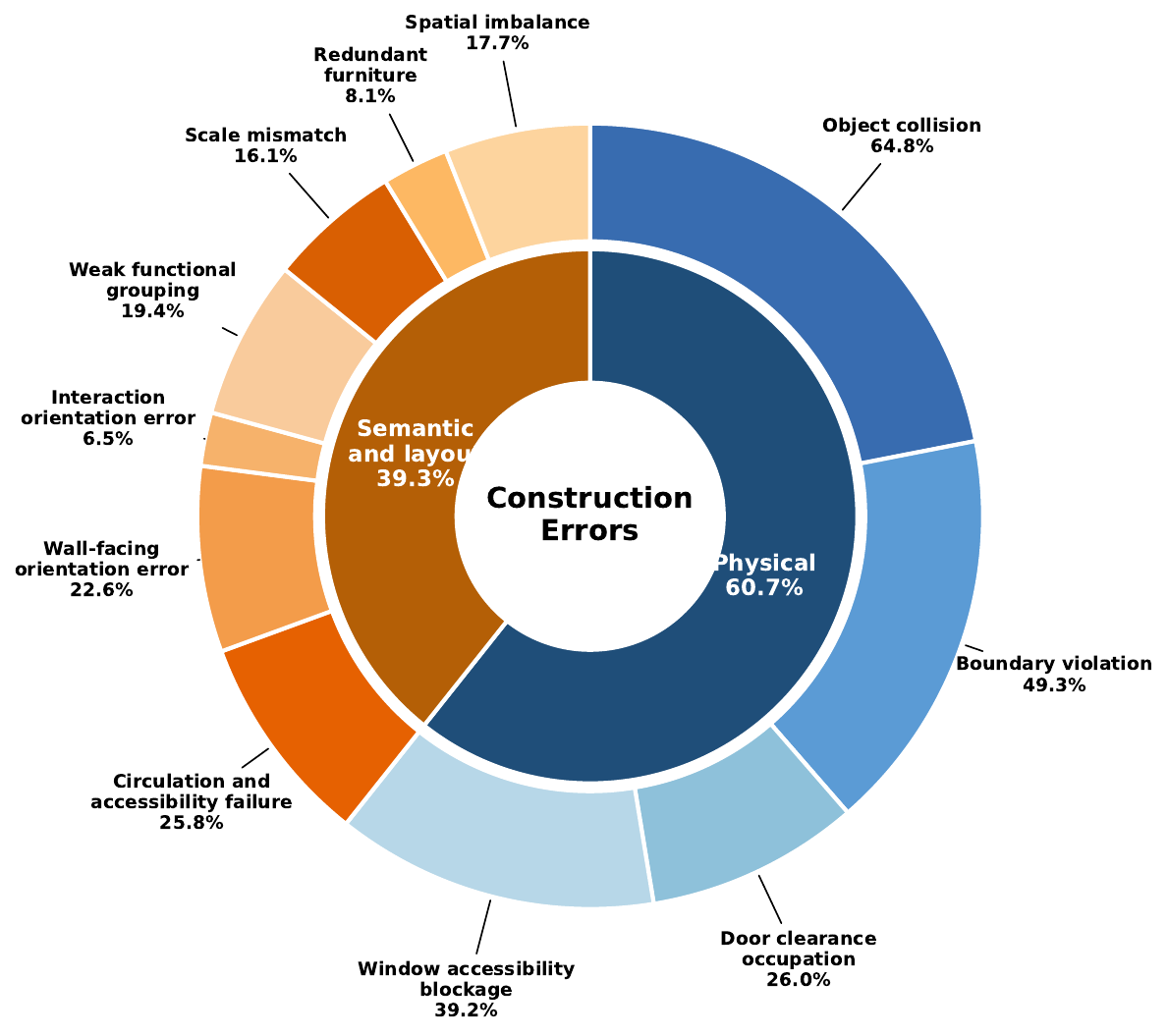}
    \end{minipage}
    \hfill
    \begin{minipage}{0.56\columnwidth}
        \centering
        \includegraphics[width=\linewidth]{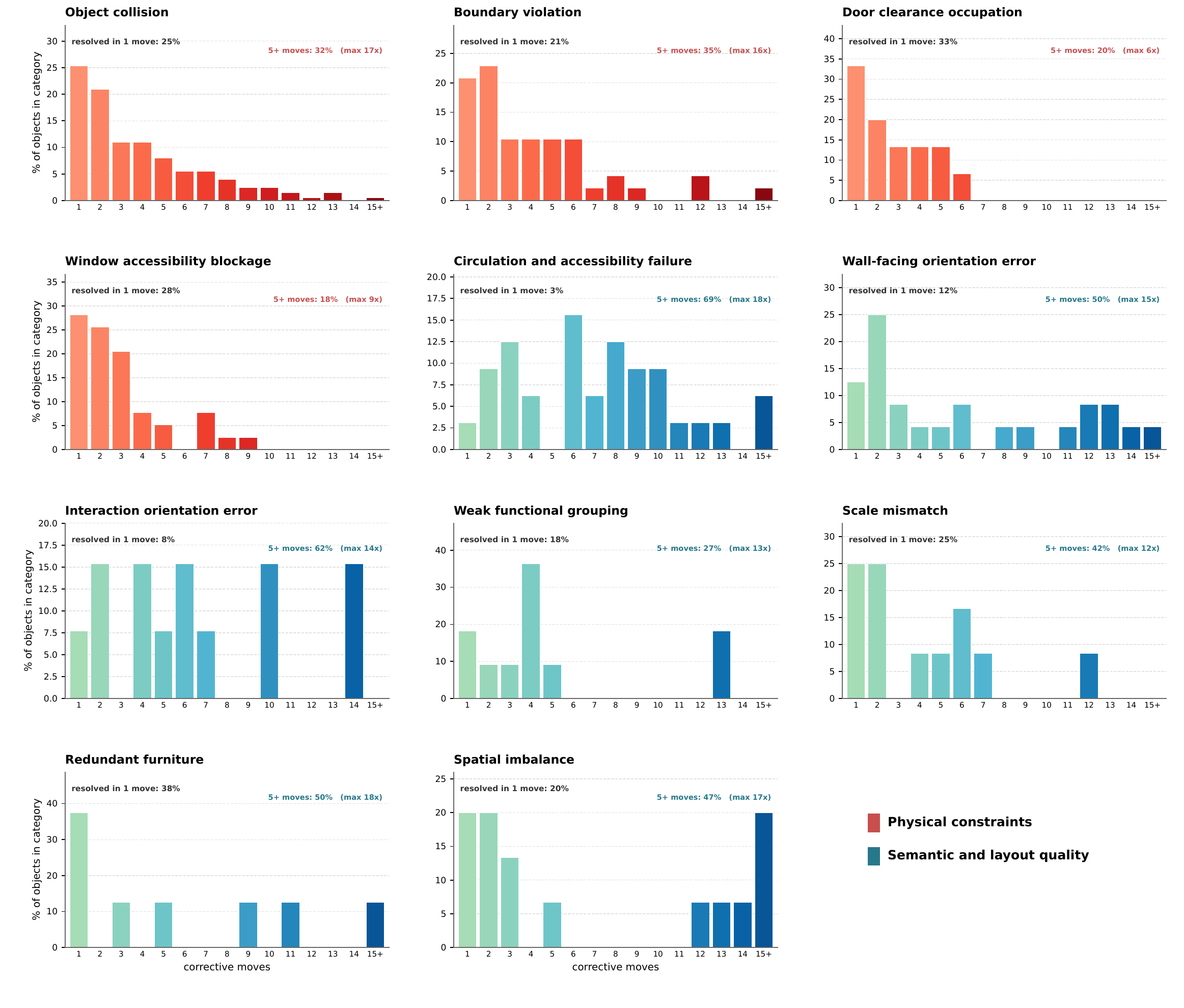}
    \end{minipage}
    \caption{
        Analysis of construction errors and correction behavior in 100 scenes generated by GPT-5.2.
        (a) \textbf{Left:} Distribution of construction errors across \textbf{physical errors} and \textbf{semantic and layout errors}.
        (b) \textbf{Right:} Distribution of corrective steps for each error category, showing that even straightforward errors often involve unnecessary trial and error before being resolved.
    }
    \label{fig:1}
\end{figure}

\begin{figure}[!t]
    \centering
    \begin{minipage}[t]{0.49\columnwidth}
        \centering
        \includegraphics[width=\linewidth]{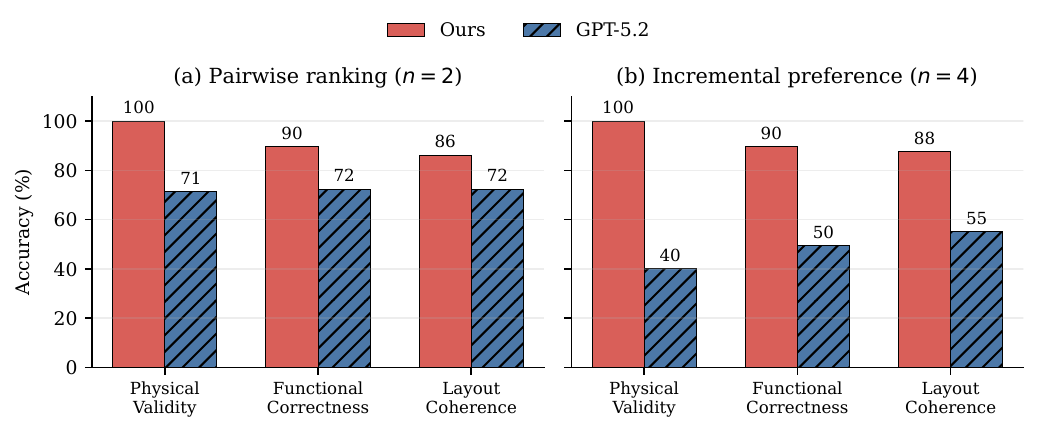}
    \end{minipage}
    \hfill
    \begin{minipage}[t]{0.49\columnwidth}
        \centering
        \includegraphics[width=\linewidth]{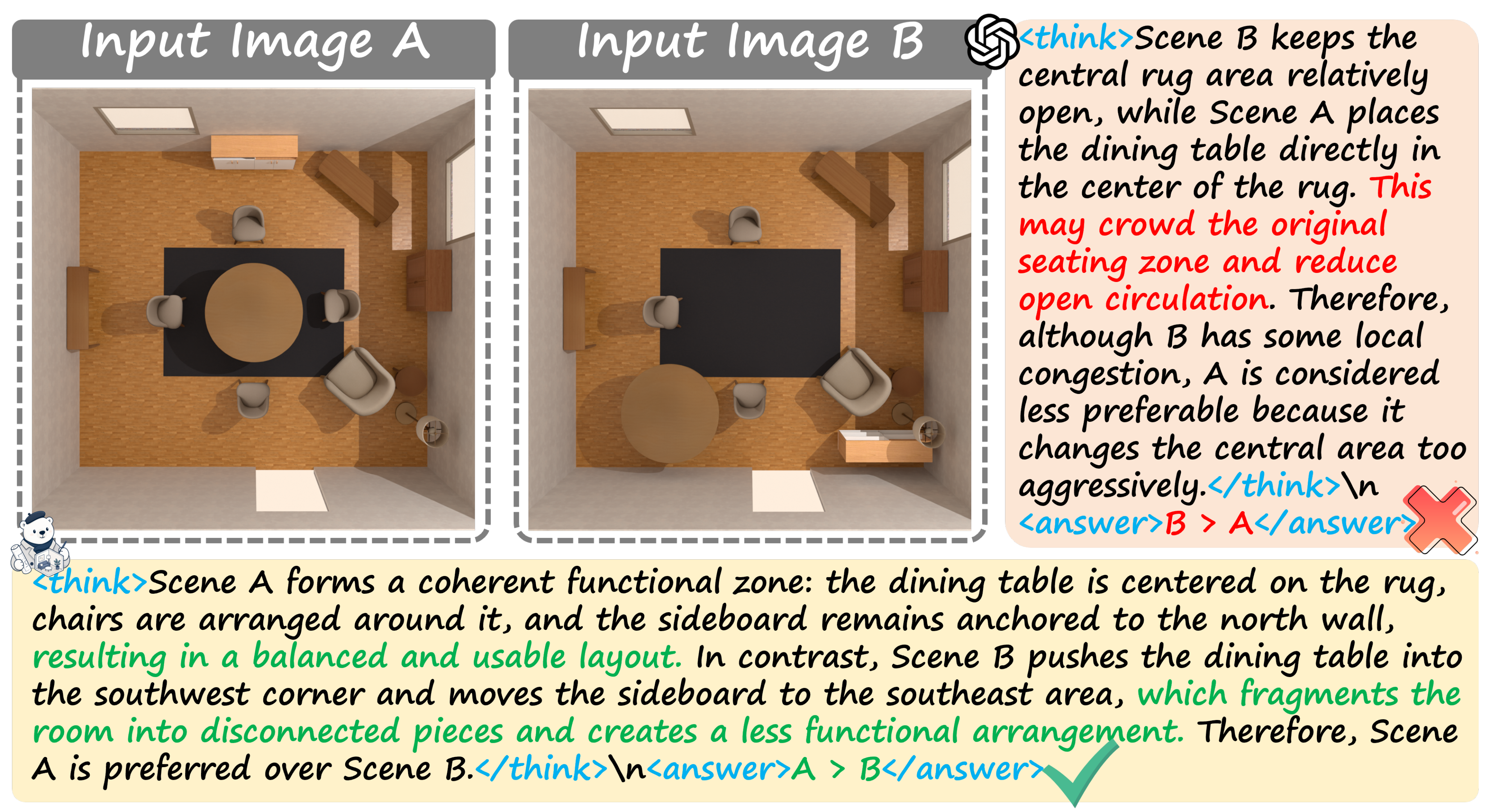}
    \end{minipage}

    \caption{
        Analysis of scene critique ability.
        (a) \textbf{Left:} Ours vs. GPT-5.2 on \textbf{physical validity}, \textbf{functional correctness}, and \textbf{layout coherence}: two-scene ranking accuracy ($n=2$) and adjacent-pair ordering accuracy within four-scene rankings ($n=4$).
        (b) \textbf{Right:} A representative failure case of GPT-5.2 in scene critique.
    }
    \label{fig:3}
\end{figure}

% Challenge callouts: visual styling only; original titles and prose preserved.
\begingroup
\definecolor{ChallengeBlue}{HTML}{397FB4}
\definecolor{ChallengeTeal}{HTML}{009C91}
\tcbset{introchallenge/.style={
    enhanced,
    colback=#1!4!white,
    colframe=#1,
    boxrule=0.65pt,
    arc=2.5mm,
    left=3mm, right=3mm,
    top=3mm, bottom=2.5mm,
    before skip=14pt, after skip=12pt,
    fonttitle=\sffamily\bfseries\normalsize,
    coltitle=black,
    attach boxed title to top left={xshift=3mm,yshift=-2mm},
    boxed title style={
        colback=#1!12!white,
        colframe=#1!25!white,
        boxrule=0.35pt,
        arc=2mm,
        left=2mm, right=2mm,
        top=1.5mm, bottom=1.5mm
    }
}}
\begin{tcolorbox}[introchallenge=ChallengeBlue,
    title={\parbox{\dimexpr\linewidth-14mm\relax}{\textcolor{ChallengeBlue}{Challenge 1:}\enspace Inefficient correction caused by unnecessary trial and error.}}]
Error correction is an essential part of scene design, but effective correction should avoid unnecessary trial and error. When the solution is straightforward, a capable model should correct the error directly. To examine this behavior, we analyze the generation trajectories of GPT-5.2 with a strong scene-construction scaffold. We categorize the observed construction errors into two major groups: \textbf{physical errors} and \textbf{semantic and layout errors}, each containing several fine-grained error types, as shown in Fig.~\ref{fig:1}. Our analysis of 100 generated scenes shows that GPT-5.2 can identify many design errors, yet often performs a sequence of unnecessary corrective attempts before reaching an effective solution. This behavior is particularly evident for errors that could be resolved with only one or two simple operations. For example, a window accessibility blockage can usually be corrected by directly moving the obstructing object, yet approximately 18\% of such ases still require more than five corrective moves. Similar patterns appear across multiple error categories. These results indicate that the main limitation is not merely error recognition, but the lack of design experience needed to avoid unnecessary trial and error and perform efficient correction.
\end{tcolorbox}

\begin{tcolorbox}[introchallenge=ChallengeTeal,
    title={\textcolor{ChallengeTeal}{Challenge 2:}\enspace Insufficient scene critique ability.}]
Scene critique is another essential component of design intelligence. Reliable critique requires more than recognizing obvious violations of \textbf{physical validity}; it must also capture whether a scene is functionally correct and whether its overall layout is coherent. To examine this capability, we compare GPT-5.2 with our domain-trained model on the Scene-OTC test sets across these three dimensions. As shown in Fig.~\ref{fig:3}, GPT-5.2 achieves around 72\% accuracy in pairwise scene comparison, but its performance drops to around 50\% in the more challenging four-scene preference ranking. In contrast, our model consistently achieves higher accuracy across all three dimensions. Fig.~\ref{fig:3} further illustrates a representative failure case. Although Scene A forms a more coherent dining area by grouping functionally related furniture, GPT-5.2 prefers Scene B because of its more open central space, overlooking the stronger functional organization and layout coherence of Scene A. Together, these results suggest that general-purpose models lack the scene-specific critique experience needed to reliably distinguish scene quality across physical validity, functional correctness, and layout coherence.
\end{tcolorbox}
\endgroup

These challenges cannot be directly addressed by existing scene data alone. Existing scene datasets, such as 3D-FRONT~\cite{fu20213d} and 3D-SynthPlace~\cite{yang2026optiscene}, mainly provide static scene states with object poses, but do not capture experience from the scene design process and critique. For scene design, construction trajectories provide richer supervision, but raw trajectories often contain substantial unnecessary trial and error, reflecting the exploration process rather than effective design strategies. Extracting useful construction and correction decisions from such trajectories is therefore essential for obtaining high-quality design experience. For scene critique, the challenge is different. Many scenes can satisfy physical constraints while still differing considerably in functional correctness and layout coherence, making reliable and scalable preference supervision difficult to obtain. Effective ways to obtain such design and critique experience are therefore essential for learning design intelligence.

To address these data-level challenges, we construct two complementary datasets to provide design and critique experience. For scene design, we introduce \textbf{Scene-OTA-35K}, containing 1K scenes and 35K Observation--Thought--Action interactions. To address trial-and-error behaviors in raw trajectories, we propose Causal Trajectory Consolidation, which identifies potential ineffective operations through state and error changes, evaluates their correction effects, and preserves important correction decisions while removing ineffective corrections. For scene critique, we introduce \textbf{Scene-OTC-15K}, a preference dataset containing 15K multi-scene comparison samples with Observation--Thought--Critic supervision. Unlike methods relying on subjective quality annotations, we generate ordered quality degradation through three types of scene perturbations with different strengths, and construct preference comparisons through three perturbation application strategies, where quality rankings can be automatically determined by perturbation strength.

With these complementary experiences, we adopt two-stage training to develop design intelligence. First, joint SFT teaches the model to critique scenes in physical validity, functional correctness, and layout coherence, while constructing physically valid scenes and imitating efficient corrections, yielding \textbf{Scene-Builder}. We then introduce \textbf{Scene-GRPO} to further enable the model to design functionally correct and layout-coherent scenes while improving its correction efficiency. At each editing step, Scene-Builder explores multiple candidate actions from the same parent scene. A frozen Critic provides two complementary signals by comparing candidate scenes to identify better design choices and comparing each child with its parent to assess scene improvement. These signals guide better decisions and reduce unnecessary trial and error, transforming Scene-Builder into \textbf{Scene-Designer}.

Extensive experiments demonstrate that Scene-Designer consistently improves scene quality across physical validity, functional correctness, and layout coherence, while making more effective correction decisions. For example, compared with Scene-Builder, the proportion of interaction-orientation errors requiring five or more corrective moves drops from 40\% to 7\%, while that of window-accessibility blockages decreases from 21\% to 5\%. The learned Scene-Critic also substantially outperforms general-purpose models, achieving 89\% accuracy in pairwise comparison and 76\% in four-scene ranking, compared with 72\% and 15\% for GPT-5.2, respectively. Moreover, beyond providing reward signals during training, the learned Scene-Critic can also guide scene construction at test time by selecting better editing decisions, further improving scene quality. Additional experiments reveal a promising trend of mutual reinforcement between scene design and critique.

Our contributions are summarized as follows:
(1) We construct two complementary datasets, \textbf{Scene-OTA-35K} and \textbf{Scene-OTC-15K}, providing effective design experience and reliable critique experience for learning design intelligence.
(2) We propose \textbf{Scene-GRPO}, a Critic-guided GRPO that leverages scene critique ability to optimize scene design decisions for high-quality scenes and reduce unnecessary trial-and-error processes.
(3) Extensive experiments demonstrate that our model generates scenes with superior physical validity, functional correctness, and layout coherence, while achieving substantially stronger scene critique ability than general-purpose models.

\section{Related Work}
\noindent\textbf{Foundation Model-based Scene Generation.}
Foundation models enable language-driven 3D scene generation through semantic understanding and spatial reasoning. Early methods, such as LayoutGPT~\cite{feng2023layoutgpt}, LayoutVLM~\cite{sun2025layoutvlm}, and Holodeck~\cite{yang2024holodeck}, use LLMs or VLMs to generate scene layouts or complete 3D scenes. Recent agentic frameworks, including I-Design~\cite{ccelen2024design}, SceneSmith~\cite{pfaff2026scenesmithagenticgenerationsimulationready}, and SceneWeaver~\cite{yang2026sceneweaver}, support iterative scene construction through tool execution and feedback. However, these approaches largely rely on general-purpose models for design decisions, which can lead to high interaction costs and inefficient error correction.

\noindent\textbf{Learning-based Scene Generation.}
Large-scale datasets, such as 3D-FRONT~\cite{fu20213d} and 3D-SynthPlace~\cite{yang2026optiscene}, provide scene layouts with object attributes and spatial relationships. Such scene representations have supported autoregressive~\cite{paschalidou2021atiss}, diffusion-based~\cite{tang2024diffuscene,yang2024physcene}, and graph-based~\cite{lin2024instructscene} generation methods. While these methods learn plausible spatial arrangements from scene collections, existing datasets mainly provide static scene states without capturing experience from the scene design process. In particular, they lack supervision on construction decisions, error corrections, and their effects on scene quality.

\noindent\textbf{Scene Evaluation and Critique.}
Existing scene evaluation methods commonly rely on geometric metrics, human preferences, or VLM-based judgments~\cite{feng2023layoutgpt,yang2024holodeck,yang2026sceneweaver}. More recent approaches introduce structured evaluation mechanisms. SceneEval~\cite{tam2026sceneeval} evaluates instruction fidelity and physical plausibility through predefined metrics, while SceneCritic~\cite{sengupta2026scenecritic} employs a symbolic evaluator based on statistical spatial priors to check object scales, co-occurrence, orientations, and overlaps. Although these approaches provide interpretable assessments, their predefined criteria remain limited in capturing holistic functional organization, layout coherence, and design preferences.

\begin{figure*}[!t]
    \centering
    \includegraphics[width=\textwidth]{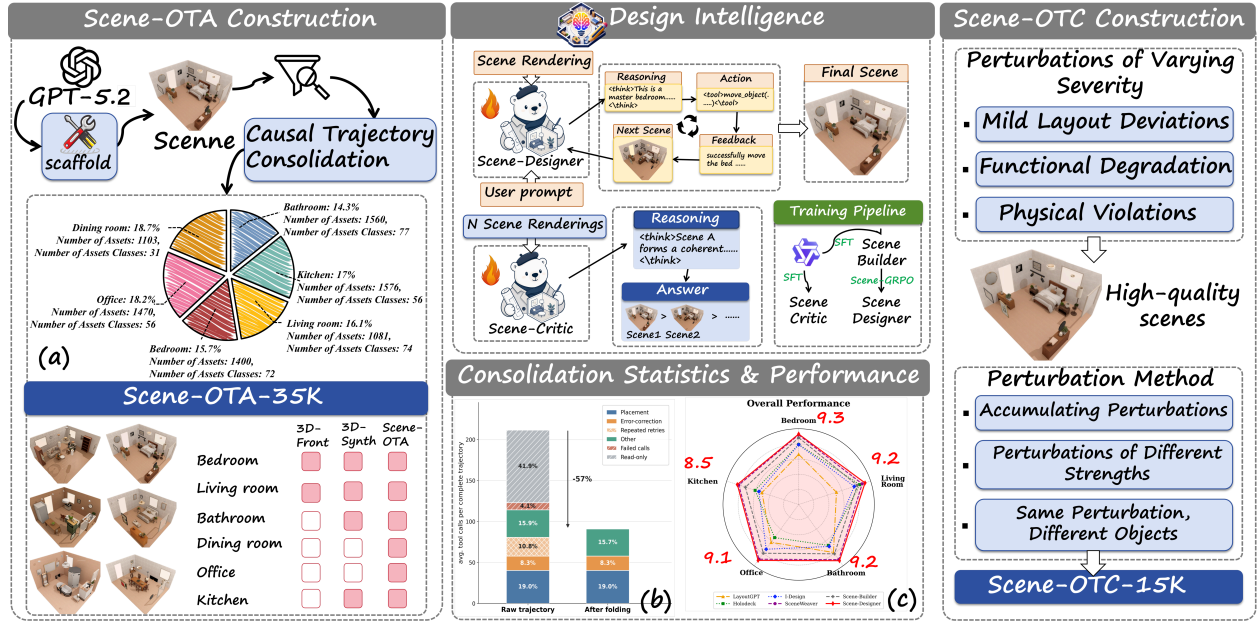}
    \caption{
    Overview of Scene-Designer.
    The framework learns \textbf{Design Intelligence} from two complementary experiences:
    \textbf{Scene-OTA Construction} extracts effective design experience from construction trajectories, while
    \textbf{Scene-OTC Construction} generates scene preference comparisons for learning critique.
    Together, they enable Scene-Designer to learn both scene design and critique abilities.
    \textbf{Consolidation Statistics \& Performance} presents
    (a) the scene categories, asset counts, and asset classes in Scene-OTA-35K;
    (b) trajectory statistics before and after consolidation;
    (c) overall scene quality comparisons across methods, with scores averaged over Real., Func., Lay., and Comp.
    }
    \label{fig:5}
\end{figure*}

\section{Method}
As shown in Fig.~\ref{fig:5}, we develop \textbf{Scene-Designer} in two stages. First, joint SFT on \textbf{Scene-OTA-35K} and \textbf{Scene-OTC-15K} teaches the model to construct physically valid scenes, learn effective corrections, and critique scene quality, yielding \textbf{Scene-Builder}. We then introduce \textbf{Scene-GRPO}, which uses a frozen Critic to compare candidate scenes and assess each child against its parent, guiding better design decisions and reducing unnecessary trial and error. This transforms Scene-Builder into \textbf{Scene-Designer}.

\subsection{Learning Design Intelligence: From General VLM to Scene-Builder}
To provide complementary experience for scene design and critique, we construct two datasets: \textbf{Scene-OTA-35K} and \textbf{Scene-OTC-15K}.

\noindent\textbf{Learning Scene Design with Scene-OTA.}
Collecting large-scale human scene-construction trajectories is costly. We therefore adopt the SceneSmith~\cite{pfaff2026scenesmithagenticgenerationsimulationready} scene-construction scaffold and use GPT-5.2 to generate scenes. After manually selecting high-quality scenes, we obtain their raw construction trajectories.

However, raw trajectories often contain unnecessary trial and error. To extract effective design experience, we propose \textbf{Causal Trajectory Consolidation}, which removes ineffective operations while preserving important construction and correction decisions. Given an original trajectory $T$, our goal is to obtain a shorter trajectory $T'$ that preserves effective correction strategies and produces the same final scene state:
\begin{equation}
T'\subseteq T,\quad S(T')=S(T).
\end{equation}

The consolidation process consists of two stages. First, \textbf{error-aware candidate identification} examines each operation through changes in scene state:
\begin{equation}
\Delta_i^{state}=D(S_{i-1},S_f)-D(S_i,S_f)
\end{equation}
and remaining errors:
\begin{equation}
\Delta_i^{error}=V_{i-1}-V_i,
\end{equation}
where $S_i$ is the state after operation $i$, $S_f$ is the final accepted state, $D(\cdot,\cdot)$ measures differences in object positions and orientations, and $V_i$ counts remaining errors. Operations that neither improve the scene state nor reduce errors are flagged as potentially ineffective.

Second, \textbf{correction-effect evaluation} uses GPT-5.2 to assess whether each candidate operation contributes to effective correction or necessary intermediate states, considering scene states, environment feedback, and construction context. Only ineffective operations without downstream contributions are removed.

The consolidated trajectories are then replayed to verify that the final scene states remain unchanged. As shown in Fig.~\ref{fig:5}(b), consolidation reduces the average trajectory length from approximately 110 to 30 steps. For each retained action, GPT-5.2 further generates step-level reasoning traces conditioned on the current scene state, interaction history, and target action. The resulting \textbf{Scene-OTA-35K} contains 1K scenes and 35K Observation--Thought--Action steps, providing supervision for learning effective scene design strategies. Detailed procedures and examples are provided in Appendix~\ref{app:scene-ota}.

\noindent\textbf{Learning Scene Critique with Scene-OTC.}
Based on high-quality scenes from Scene-OTA, we construct Scene-OTC to provide reliable preference supervision for scene critique. We introduce three levels of scene perturbations with increasing severity:
\ding{182}\textbf{Mild layout deviations} preserve physical validity and functional correctness but reduce layout coherence;
\ding{183}\textbf{Functional degradation} disrupts functional relationships while generally maintaining physical validity;
\ding{184}\textbf{Severe physical violations} introduce collisions, out-of-bound placements, and other physical constraint violations.

We construct ordered preference comparisons through three perturbation application strategies:
\ding{182}\textbf{accumulating perturbations} to progressively degrade scene quality;
\ding{183}\textbf{applying perturbations of different strengths to the same object};
and \ding{184}\textbf{applying the same perturbation to different objects}.
The first two strategies allow automatic ranking based on perturbation severity, while the third requires manual annotation, as the impact of a perturbation depends on the affected object's functional role.
We further use GPT-5.2 to generate reasoning traces from scene renderings and quality differences, forming Observation--Thought--Critic supervision for learning fine-grained scene quality distinctions across physical validity, functional correctness, and layout coherence.
Each example contains $N$ scene renderings, a reasoning trace, and a complete ranking from best to worst.
The resulting \textbf{Scene-OTC-15K} contains approximately 15K critique examples.
We also construct two test sets: 94 two-scene comparisons and 55 four-scene rankings.
Detailed perturbation configurations and examples are provided in Appendix~\ref{app:scene-otc}.

\noindent\textbf{Joint Supervised Fine-Tuning.}
Although general-purpose vision-language models possess strong visual understanding and language reasoning abilities, they lack task-specific experience for scene design and critique. We therefore jointly fine-tune the model on Scene-OTA and Scene-OTC. This SFT stage enables the model to construct physically valid scenes, imitate efficient corrections, and evaluate scene quality from the perspectives of physical validity, functional correctness, and layout coherence, yielding \textbf{Scene-Builder}.

\begin{figure}[!t]
    \centering
    \includegraphics[width=1\columnwidth]{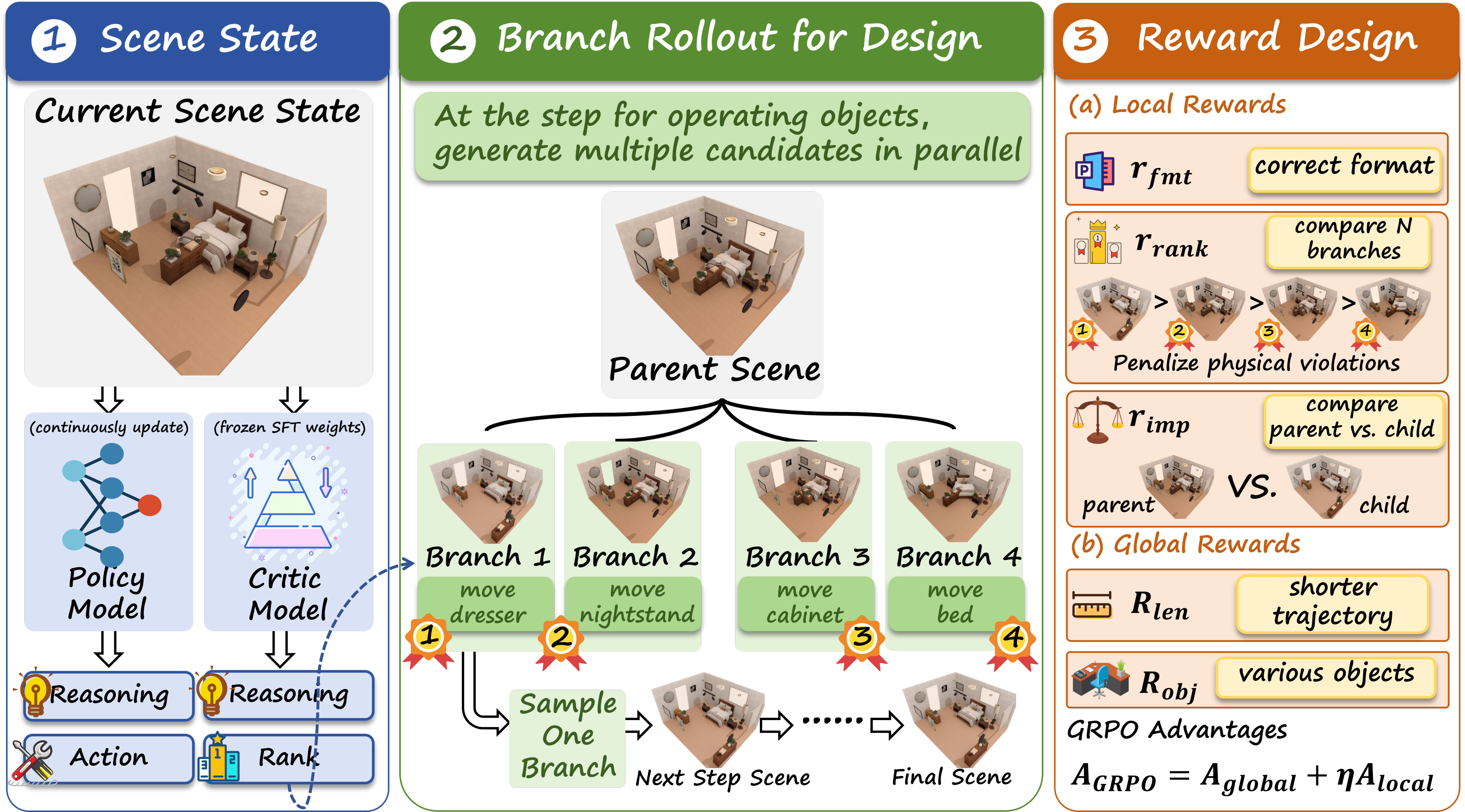}
    \caption{Overview of Scene-GRPO. At each editing step, the policy expands multiple candidate actions from the current scene and selects the best branch to continue the trajectory. A frozen Critic provides local rewards through branch ranking and parent--child comparison, while global rewards encourage efficient and complete scene design.}
    \label{fig:6}
\end{figure}

\begin{table*}[t]
\centering
\small
\setlength{\tabcolsep}{2pt}

\resizebox{\textwidth}{!}{%
\begin{tabular}{c|ccc|cccc|ccc|cccc}
\toprule

\multirow{3}{*}{Method}
& \multicolumn{7}{c|}{Bedroom}
& \multicolumn{7}{c}{Living Room}
\\

\cmidrule(lr){2-8}
\cmidrule(l){9-15}

& \multicolumn{3}{c|}{Physics}
& \multicolumn{4}{c|}{Visual \& Semantics}
& \multicolumn{3}{c|}{Physics}
& \multicolumn{4}{c}{Visual \& Semantics}
\\

\cmidrule(lr){2-4}
\cmidrule(lr){5-8}
\cmidrule(lr){9-11}
\cmidrule(l){12-15}

& \#Obj $\uparrow$
& \#OB $\downarrow$
& \#CN $\downarrow$
& Real.$\uparrow$
& Func.$\uparrow$
& Lay.$\uparrow$
& Comp.$\uparrow$

& \#Obj $\uparrow$
& \#OB $\downarrow$
& \#CN $\downarrow$
& Real.$\uparrow$
& Func.$\uparrow$
& Lay.$\uparrow$
& Comp.$\uparrow$
\\

\midrule

ATISS
& 3.9 & 0.5 & 0.6
& 7.4 & 7.1 & 6.6 & 4.2
& 7.8 & \underline{0.1} & \underline{0.7}
& 5.8 & 5.3 & 6.4 & 3.7
\\

DiffuScene
& 3.5 & \underline{0.1} & 1.1
& 6.5 & 7.0 & 6.7 & 3.6
& 6.9 & 0.5 & 1.2
& 5.5 & 4.9 & 5.2 & 3.5
\\

PhyScene
& 3.3 & \underline{0.1} & \underline{0.3}
& 5.7 & 6.3 & 5.7 & 4.0
& 8.0 & \textbf{0.0} & \underline{0.7}
& 5.2 & 5.3 & 5.1 & 3.3
\\

\midrule

LayoutGPT
& 5.4 & 1.0 & 1.3
& 7.5 & 8.1 & 6.7 & 4.2
& 8.4 & 1.1 & 2.8
& 6.4 & 5.8 & 5.2 & 3.6
\\

Holodeck
& \textbf{32.2} & \textbf{0.0} & 38.5
& 8.6 & 9.1 & 7.8 & 6.2
& \textbf{23.0} & \textbf{0.0} & 5.3
& 8.9 & 9.3 & 7.6 & 8.1
\\

I-Design
& 9.6 & \textbf{0.0} & \textbf{0.0}
& 8.6 & 9.3 & 7.6 & 6.1
& 9.7 & \textbf{0.0} & \textbf{0.0}
& 8.4 & 8.9 & 7.7 & 5.9
\\

SceneWeaver
& 14.0 & \textbf{0.0} & \textbf{0.0}
& 9.2 & \textbf{9.8} & 8.4 & \underline{9.4}
& 17.3 & \textbf{0.0} & \textbf{0.0}
& 9.1 & \textbf{9.5} & 8.0 & 8.7
\\

\midrule

\rowcolor{BuilderBG}
Scene-Builder
& 19.6
& \textbf{0.0}
& \textbf{0.0}
& 8.8
& 9.2
& 8.4
& 8.9
& 17.9
& \textbf{0.0}
& \textbf{0.0}
& 8.0
& 8.6
& 8.2
& 7.8
\\

\rowcolor{DesignerBG}
\textbf{Scene-Designer}
& 20.0\upchg{0.4}
& \textbf{0.0}
& \textbf{0.0}
& \underline{9.5}\upchg{0.7}
& \underline{9.5}\upchg{0.3}
& \underline{8.8}\upchg{0.4}
& \underline{9.4}\upchg{0.5}
& 20.3\upchg{2.4}
& \textbf{0.0}
& \textbf{0.0}
& \underline{9.6}\upchg{1.6}
& 9.0\upchg{0.4}
& \underline{9.3}\upchg{1.1}
& \underline{9.0}\upchg{1.2}
\\

\rowcolor{CriticBG}
\textbf{Scene-Designer w/ Critic}
& \gaincell{\underline{20.8}\upchg{0.8}}
& \textbf{0.0}
& \textbf{0.0}
& \gaincell{\textbf{9.8}\upchg{0.3}}
& \gaincell{\textbf{9.8}\upchg{0.3}}
& \gaincell{\textbf{9.2}\upchg{0.4}}
& \gaincell{\textbf{9.7}\upchg{0.3}}
& \gaincell{\underline{21.1}\upchg{0.8}}
& \textbf{0.0}
& \textbf{0.0}
& \gaincell{\textbf{10.0}\upchg{0.4}}
& \gaincell{\underline{9.4}\upchg{0.4}}
& \gaincell{\textbf{9.6}\upchg{0.3}}
& \gaincell{\textbf{9.5}\upchg{0.5}}
\\

\bottomrule
\end{tabular}%
}

\caption{
Quantitative comparison on common room types.
\textbf{Bold} and \underline{underlined} numbers indicate the best and second-best performance, respectively.
Shaded rows highlight our models, with green cells indicating improvements from test-time Critic guidance.
Green and red superscripts denote performance improvements and degradations relative to the preceding model variant, respectively.
}
\label{tab:1}

\end{table*}

\subsection{Critic-Guided Optimization: From Scene-Builder to Scene-Designer}
Although Scene-Builder acquires design and critique abilities through SFT, it mainly imitates existing design decisions without directly optimizing scene quality and correction efficiency. We therefore introduce \textbf{Scene-GRPO} to improve functional correctness, layout coherence, and correction efficiency. Specifically, we initialize two copies of Scene-Builder: a frozen Critic and a trainable policy.

\noindent\textbf{Branch-Sampled Rollout.}
To explore alternative design decisions, Scene-GRPO samples multiple candidate actions from the same parent scene at each editing step and executes them to obtain alternative child scenes. The frozen Critic ranks the valid candidate scenes based on physical validity, functional correctness, and layout coherence, and the highest-ranked branch is selected to continue the trajectory.

\noindent\textbf{Reward Design.}
To encourage higher-quality design decisions and reduce unnecessary trial and error, we introduce complementary local and global rewards. Two local rewards are central to Scene-GRPO. The \textbf{branch-ranking reward} compares sibling scenes generated from the same parent, encouraging the policy to favor better design choices. The \textbf{improvement reward} compares each child scene with its parent to determine whether the edit actually improves scene quality, discouraging ineffective modifications. Together, these rewards distinguish selecting better actions among alternatives from assessing whether an action leads to genuine improvement.

We further introduce auxiliary rewards to support stable and efficient training. The format reward encourages properly formatted reasoning and valid tool calls, while the global length and object-completeness rewards encourage efficient yet complete scene construction. Detailed reward definitions are provided in Appendix~\ref{app:scene-grpo-rewards}.

We optimize the policy using the standard GRPO objective, transforming Scene-Builder into \textbf{Scene-Designer} with improved scene design and correction capabilities.

\section{Experiment}
\noindent\textbf{Training Setup.}
We use Qwen3-VL-8B-Instruct as the base model and perform full-parameter fine-tuning. SFT is trained for 2 epochs with a learning rate of $1\times10^{-5}$, cosine scheduling, and 0.08 warmup ratio on 4 A800 GPUs. After SFT, we apply Scene-GRPO, where both policy and Critic are initialized from the SFT model but only the policy is updated. The policy samples 4 branches per editing step with a GRPO group size of 4, a maximum trajectory length of 80 steps, and a learning rate of $1\times10^{-6}$.

\noindent\textbf{Evaluation Metrics.}
Following SceneWeaver~\cite{yang2026sceneweaver}, we use ``Design me a \{room\_type\}'' as the unified prompt for all LLM-based methods. For each room type, each method generates 10 scenes for evaluation. Generated scenes are evaluated from two aspects: physical validity and visual-semantic quality. Physical validity is measured using objective metrics, including the number of objects (\#Obj$\uparrow$), out-of-bound objects (\#OB$\downarrow$), and object collisions (\#CN$\downarrow$). Visual-semantic quality is assessed by GPT-4 from four dimensions: realism (Real.$\uparrow$), functionality (Func.$\uparrow$), layout quality (Lay.$\uparrow$), and completeness (Comp.$\uparrow$), with each dimension scored from 0 to 10.
% \begin{figure}[!t]
%     \centering
%     \includegraphics[width=\columnwidth]{Figures_new/fig7.pdf}
%     \caption{Quantitative comparison on other scenes.}
%     \label{fig:7}
% \end{figure}

\begin{table*}[t]
\centering
\small
\setlength{\tabcolsep}{2pt}

\resizebox{\textwidth}{!}{%
\begin{tabular}{c|ccc|cccc|ccc|cccc}
\toprule
\multirow{3}{*}{Method}
& \multicolumn{7}{c|}{Bathroom}
& \multicolumn{7}{c}{Office}
\\
\cmidrule(lr){2-8}
\cmidrule(l){9-15}
& \multicolumn{3}{c|}{Physics}
& \multicolumn{4}{c|}{Visual \& Semantics}
& \multicolumn{3}{c|}{Physics}
& \multicolumn{4}{c}{Visual \& Semantics}
\\
\cmidrule(lr){2-4}
\cmidrule(lr){5-8}
\cmidrule(lr){9-11}
\cmidrule(l){12-15}
& \#Obj $\uparrow$
& \#OB $\downarrow$
& \#CN $\downarrow$
& Real.$\uparrow$
& Func.$\uparrow$
& Lay.$\uparrow$
& Comp.$\uparrow$
& \#Obj $\uparrow$
& \#OB $\downarrow$
& \#CN $\downarrow$
& Real.$\uparrow$
& Func.$\uparrow$
& Lay.$\uparrow$
& Comp.$\uparrow$
\\
\midrule
LayoutGPT
& 7.7 & \underline{1.3} & \underline{1.0}
& 8.3 & 9.3 & 7.7 & 6.0
& 7.3 & \underline{0.3} & \textbf{0.0}
& 6.7 & 7.7 & 6.3 & 4.0
\\
Holodeck
& 12.0 & \textbf{0.0} & 1.7
& 7.7 & 6.7 & 7.0 & 5.3
& \underline{27.0} & \textbf{0.0} & \underline{4.7}
& 7.0 & 6.3 & 4.3 & 4.0
\\
I-Design
& 9.7 & \textbf{0.0} & \textbf{0.0}
& 7.4 & 7.2 & 7.4 & 5.4
& 11.7 & \textbf{0.0} & \textbf{0.0}
& 8.0 & 9.0 & 6.8 & 5.4
\\
SceneWeaver
& 19.7 & \textbf{0.0} & \textbf{0.0}
& 9.0 & \textbf{10.0} & 8.0 & 9.0
& \textbf{40.0} & \textbf{0.0} & \textbf{0.0}
& 9.0 & \textbf{10.0} & 8.0 & 8.7
\\
\midrule

\rowcolor{BuilderBG}
Scene-Builder
& 18.0 & \textbf{0.0} & \textbf{0.0}
& 8.7 & 7.9 & 8.2 & 7.4
& 18.6 & \textbf{0.0} & \textbf{0.0}
& 8.0 & 8.3 & 7.5 & 8.0
\\

\rowcolor{DesignerBG}
\textbf{Scene-Designer}
& \underline{21.0}\upchg{3.0}
& \textbf{0.0}
& \textbf{0.0}
& \underline{9.4}\upchg{0.7}
& 9.2\upchg{1.3}
& \underline{8.7}\upchg{0.5}
& \underline{9.4}\upchg{2.0}
& 23.0\upchg{4.4}
& \textbf{0.0}
& \textbf{0.0}
& \underline{9.5}\upchg{1.5}
& 9.0\upchg{0.7}
& \underline{8.7}\upchg{1.2}
& \underline{9.3}\upchg{1.3}
\\

\rowcolor{CriticBG}
\textbf{Scene-Designer w/ Critic}
& \gaincell{\textbf{22.0}\upchg{1.0}}
& \textbf{0.0}
& \textbf{0.0}
& \gaincell{\textbf{9.7}\upchg{0.3}}
& \gaincell{\underline{9.5}\upchg{0.3}}
& \gaincell{\textbf{9.1}\upchg{0.4}}
& \gaincell{\textbf{9.7}\upchg{0.3}}
& \gaincell{24.2\upchg{1.2}}
& \textbf{0.0}
& \textbf{0.0}
& \gaincell{\textbf{9.8}\upchg{0.3}}
& \gaincell{\underline{9.4}\upchg{0.4}}
& \gaincell{\textbf{9.2}\upchg{0.5}}
& \gaincell{\textbf{9.6}\upchg{0.3}}
\\

\bottomrule
\end{tabular}%
}

\caption{
Quantitative comparison on other scene types.
\textbf{Bold} and \underline{underlined} numbers indicate the best and second-best performance, respectively.
Shaded rows highlight our models, with green cells indicating improvements from test-time Critic guidance.
Green and red superscripts denote performance improvements and degradations relative to the preceding model variant, respectively.
}
\label{tab:other_scenes}
\end{table*}

\begin{figure}[t]
    \centering
    \begin{minipage}{0.43\columnwidth}
        \centering
        \includegraphics[width=\linewidth]{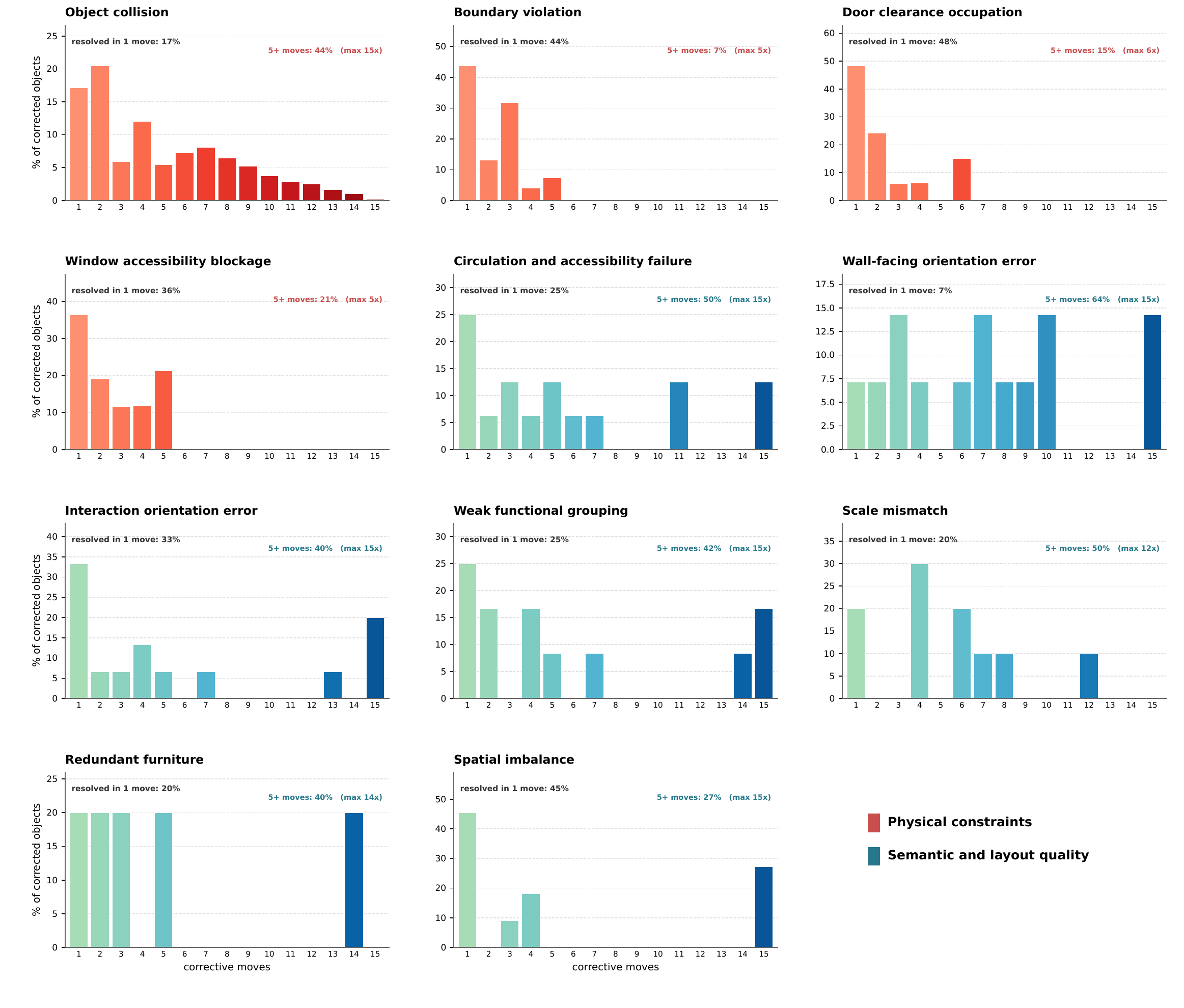}
        \caption*{(a) Scene-Builder}
    \end{minipage}
    \hfill
    \begin{minipage}{0.43\columnwidth}
        \centering
        \includegraphics[width=\linewidth]{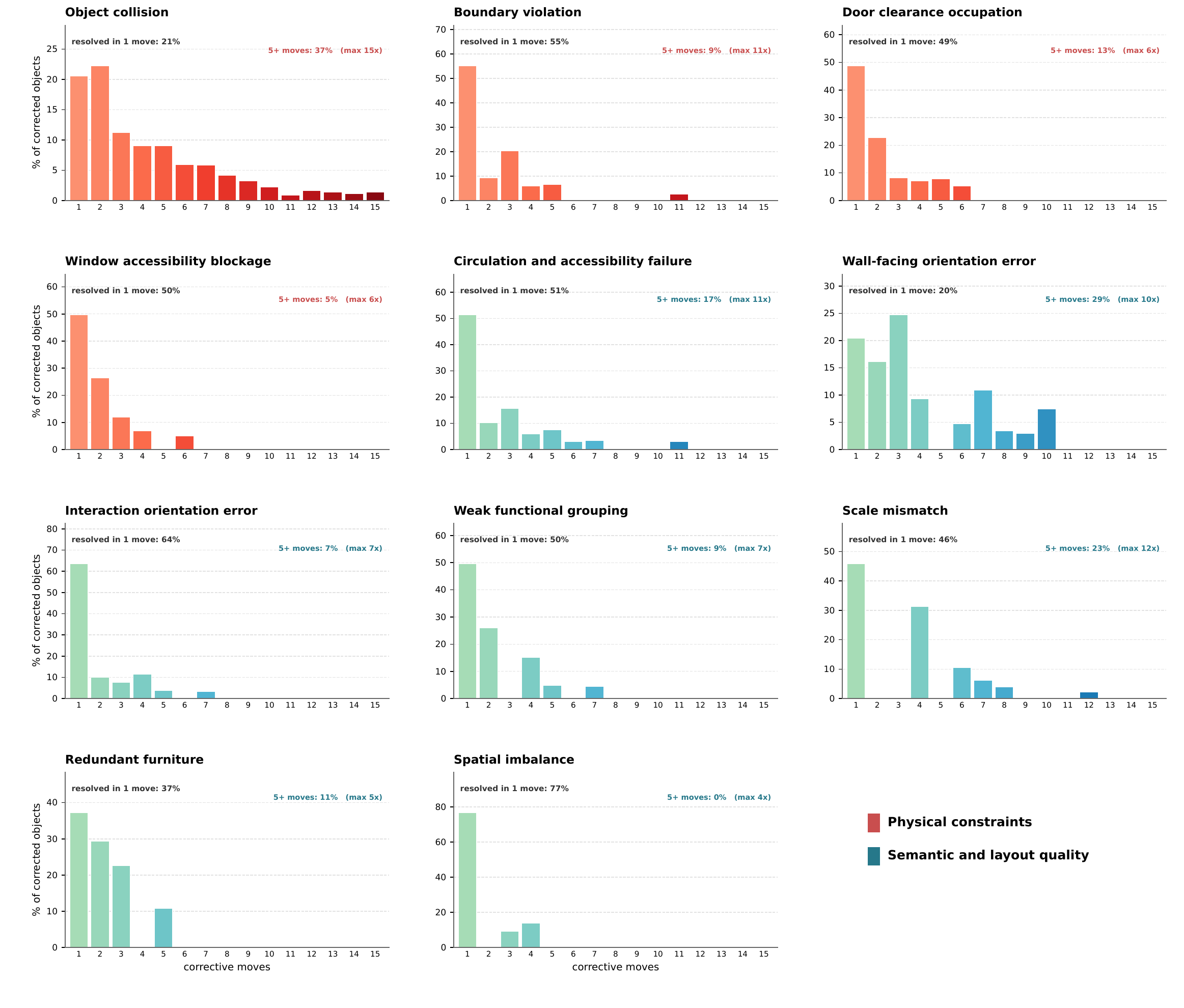}
        \caption*{(b) Scene-Designer}
    \end{minipage}

    \caption{
        Comparison of error correction efficiency across different error categories.
        Compared with (a) Scene-Builder, (b) Scene-Designer generally requires fewer corrective moves, reducing unnecessary trial and error.
        }
    \label{fig:10&11}
\end{figure}

% Measure each side figure and reserve its actual height inside a single page.
% This prevents wrapfigure settings from leaking onto the following page.
\newcommand{\MainBodySideFigure}[3]{%
\par\addvspace{3pt}%
\begingroup
\setbox0=\hbox{\begin{minipage}[t]{#1\textwidth}
\vspace{0pt}#2\end{minipage}}%
\count255=\dimexpr\ht0+\dp0+5pt\relax
\count254=\baselineskip
\advance\count255 by \count254
\advance\count255 by -1
\divide\count255 by \count254
% Keep only the lines beside the image together; later lines may cross a page.
\edef\MainLinePenalties{}
\count253=\count255
\loop\ifnum\count253>0
  \edef\MainLinePenalties{\MainLinePenalties\space10000}
  \advance\count253 by -1
\repeat
\setbox2=\vbox{%
\interlinepenalties\numexpr\count255+1\relax\MainLinePenalties\space0
\hsize=\textwidth
\noindent\makebox[0pt][l]{\hspace{\dimexpr\textwidth-#1\textwidth\relax}\raisebox{\ht\strutbox}[0pt][0pt]{\copy0}}%
\hangindent=-\dimexpr#1\textwidth+0.035\textwidth\relax
\hangafter=-\count255
#3\par}%
\dimen0=\dimexpr\ht0+\dp0\relax
\ifdim\dimexpr\ht2+\dp2\relax<\dimen0
  \vbox to \dimen0{\unvbox2\vfil}%
\else
  \unvbox2%
\fi
\endgroup
\par\addvspace{3pt}%
}

\subsection{Scene Design for Common Room Types}
\MainBodySideFigure{0.5}{
    \centering
    \includegraphics[width=\linewidth]{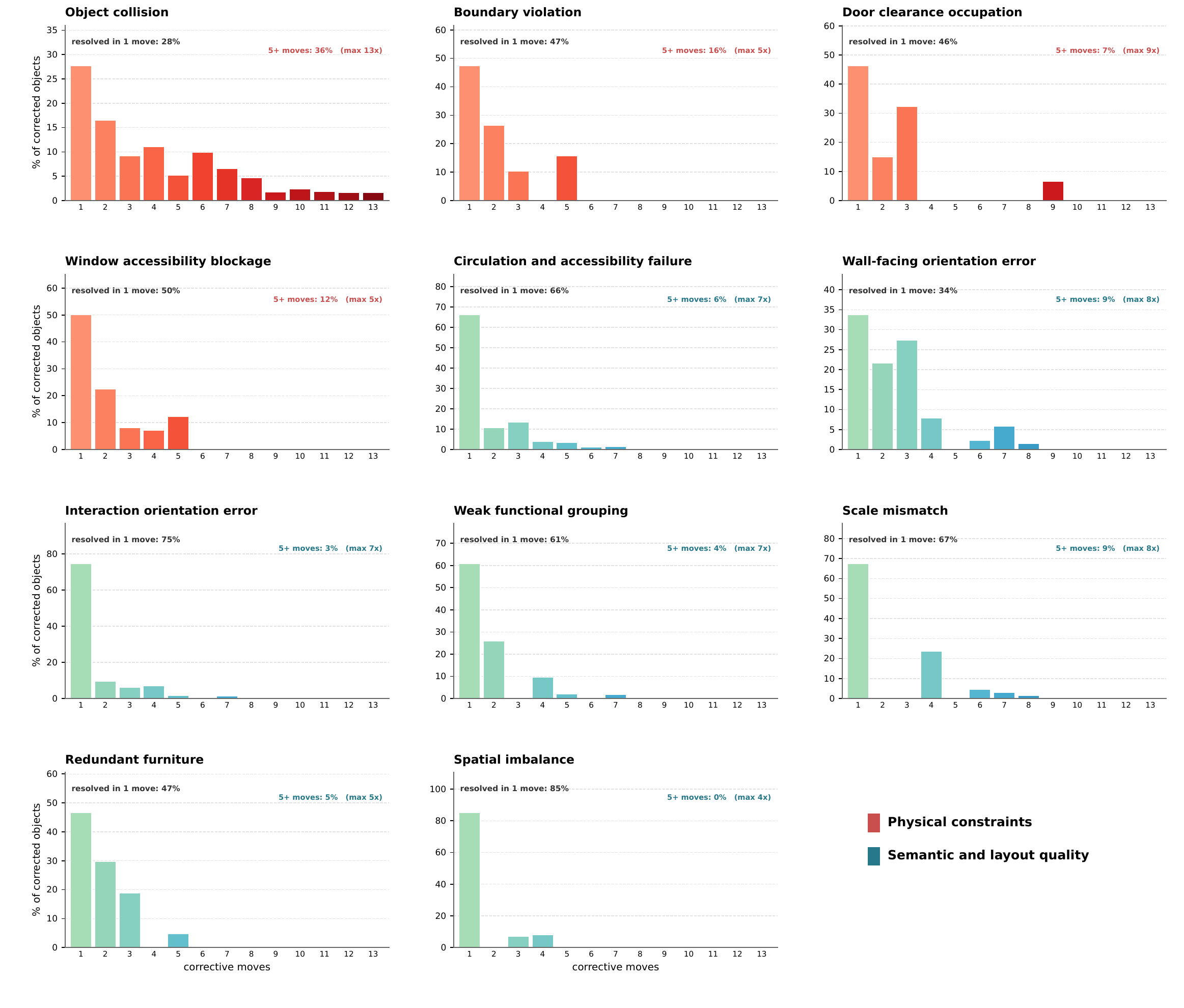}
    \captionof{figure}{
    Distribution of corrective steps for each error category of Scene-Designer w/ Critic.
    }
    \label{fig:12}
}{
\noindent\textbf{Main Results.}
As shown in Tab.~\ref{tab:1} and Tab.~\ref{tab:other_scenes}, Scene-Designer achieves strong overall performance across diverse room types. It generates scenes with high visual-semantic quality while maintaining physical validity, benefiting from the design experience learned from Scene-OTA. Furthermore, compared with learning-based methods and the SFT-only Scene-Builder, Scene-Designer achieves substantial improvements in functionality, layout quality, and realism through Critic-guided GRPO. These results demonstrate that Scene-GRPO enables the model to make better design decisions and generate physically valid, functionally correct, and layout-coherent scenes.
\hfil\break
\noindent\textbf{Test-Time Critic Guidance.}
Beyond providing reward signals during GRPO training, we investigate whether the learned Critic can directly guide scene design at inference time. At each editing step, the policy generates four candidate branches, and the Critic ranks them to select the most promising branch for continued construction. As shown in Tab.~\ref{tab:1} and Tab.~\ref{tab:other_scenes}, test-time Critic guidance consistently improves scene quality across different room types. These improvements suggest that the Critic learns meaningful design preferences that can guide better editing decisions beyond training. By filtering inferior candidate actions before committing them to the construction trajectory, Critic guidance further improves physical validity, functional correctness, and layout coherence.
}

\noindent\textbf{Correction Analysis.}
Beyond final scene quality, we analyze correction behavior across different error categories. Compared with general-purpose models, Scene-Builder already reduces corrective steps by learning effective correction strategies from Scene-OTA. However, it still requires unnecessary corrective attempts for certain errors, motivating further optimization through Scene-GRPO. Scene-Designer substantially reduces these attempts by learning more direct and effective correction strategies. Specifically, compared with Scene-Builder, Scene-Designer reduces the proportion of cases requiring five or more corrective moves for door clearance occupation, window accessibility blockage, interaction orientation error, and scale mismatch from 15\% to 13\%, 21\% to 5\%, 40\% to 7\%, and 50\% to 23\%, respectively. These results demonstrate that Scene-GRPO helps the model avoid unnecessary trial and error and make more effective correction decisions. Furthermore, Scene-Designer w/ Critic further reduces unnecessary corrections by filtering inferior editing decisions, demonstrating the additional benefits of test-time Critic guidance. Additional cases and analyses are provided in Appendix~\ref{app:case-study}.

\subsection{Scene-GRPO Analysis}

\MainBodySideFigure{0.50}{
\centering
\captionsetup{skip=4pt}
\includegraphics[width=0.7\linewidth]{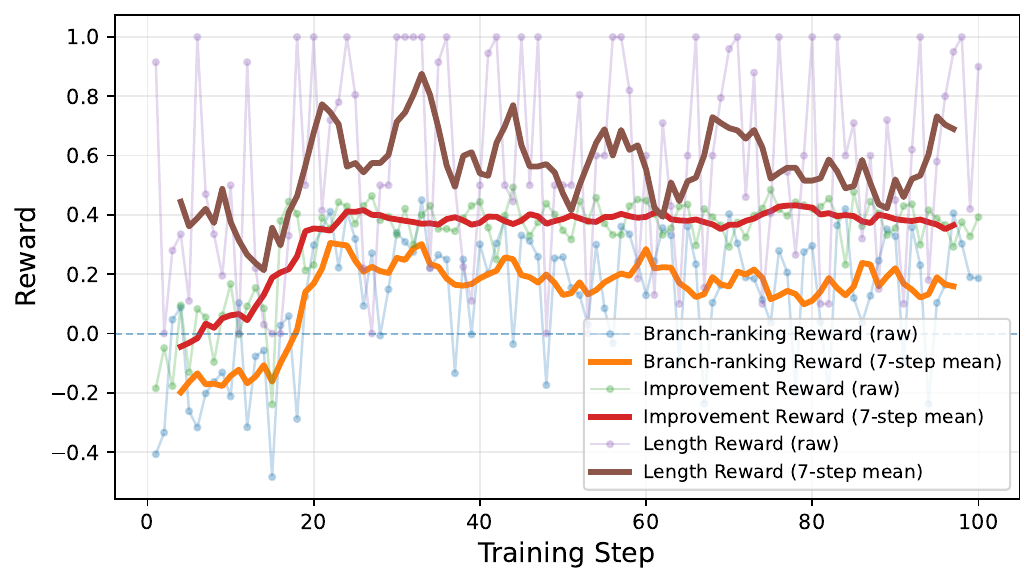}
    \captionof{figure}{Reward curves during Scene-GRPO training. 
    The curves show the evolution of branch-ranking, improvement, and length rewards during optimization.}
    \label{fig:8}
}{
\noindent\textbf{Training Curve Analysis.}
We analyze the reward dynamics during Scene-GRPO training in Fig.~\ref{fig:8}. All components exhibit overall upward trends, suggesting progressive policy improvement throughout optimization. Notably, the increases in local rewards, including branch-ranking and improvement rewards, are accompanied by steady growth in the global length reward. This suggests that the policy learns not only to select higher-quality editing decisions but also to reduce unnecessary interactions and construct scenes more efficiently.
}

\MainBodySideFigure{0.5}{
\centering
\includegraphics[width=\linewidth]{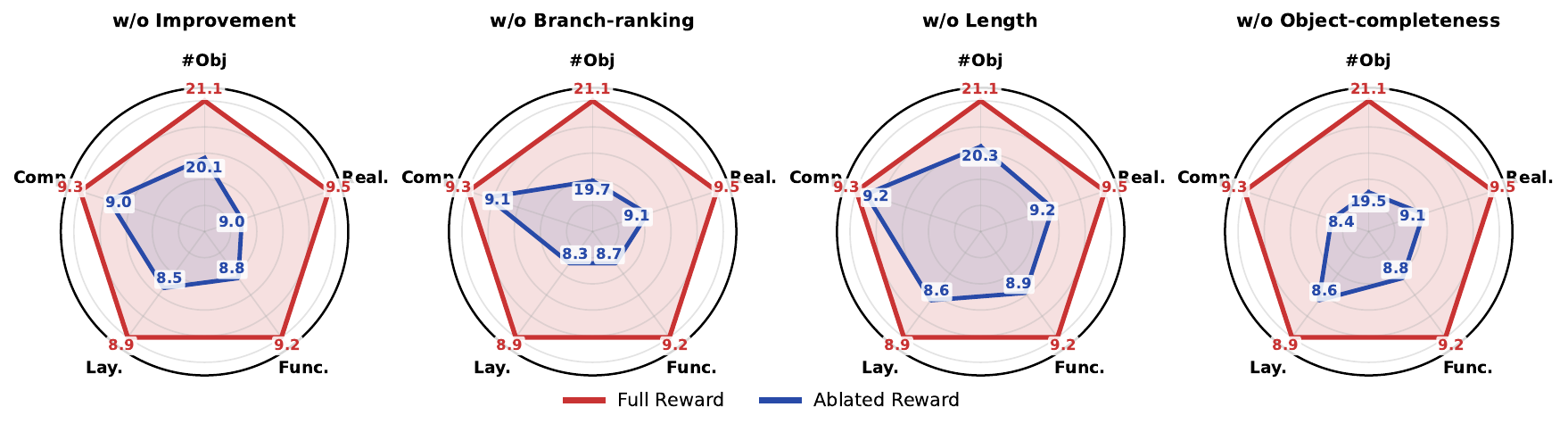}
    \captionof{figure}{Reward ablation in Scene-GRPO, comparing the full reward design with each ablated component, averaged over Bedroom, Living Room, Bathroom, and Office.}
    \label{fig:9}
}{
\noindent\textbf{Reward Component Analysis.}
We conduct an ablation study to evaluate the contribution of each reward component in Scene-GRPO. As shown in Fig.~\ref{fig:9}, removing any reward degrades performance, demonstrating their complementary contributions. Removing the branch-ranking reward causes a substantial drop, as the policy loses relative feedback among candidate actions and struggles to select better design decisions. Removing the improvement reward also reduces performance, since the policy lacks feedback on whether an edit actually improves its parent scene, leading to less effective corrections. These results highlight the complementary roles of the two rewards in guiding better design decisions and reducing unnecessary trial and error.
}

\subsection{Design and Critique: Two Sides of the Same Coin}

\noindent\textbf{Critic Reliability Evaluation.}
To assess Critic reliability, we evaluate pairwise scene comparison ($n=2$) and four-scene ranking ($n=4$) on the Scene-OTC test sets, comparing GPT-5.2, the base model, an OTC-only model, and our jointly trained Scene-OTA+OTC model. As shown in Fig.~\ref{fig:13&14}(a), the base model achieves 72\% and 2\% accuracy on $n=2$ and $n=4$, respectively, while GPT-5.2 reaches 72\% and 15\%. OTC fine-tuning improves performance to 86\% and 51\%, and joint training further raises it to 89\% and 76\%, demonstrating substantial gains in complex scene ranking.

\begingroup
\setlength{\intextsep}{6pt}
\begin{figure}[H]
    \centering
\begin{minipage}{0.94\textwidth}

    \begin{minipage}{0.42\columnwidth}
        \centering
        \includegraphics[width=1\linewidth]{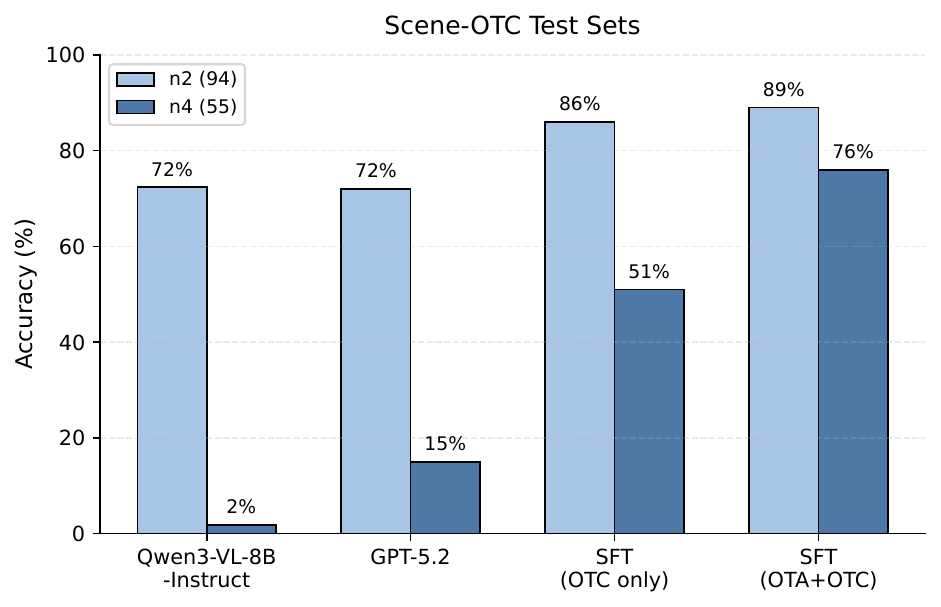}
        \caption*{(a) Overall preference accuracy.}
    \end{minipage}
    \hfill
    \begin{minipage}{0.56\columnwidth}
        \centering
        \includegraphics[width=1\linewidth]{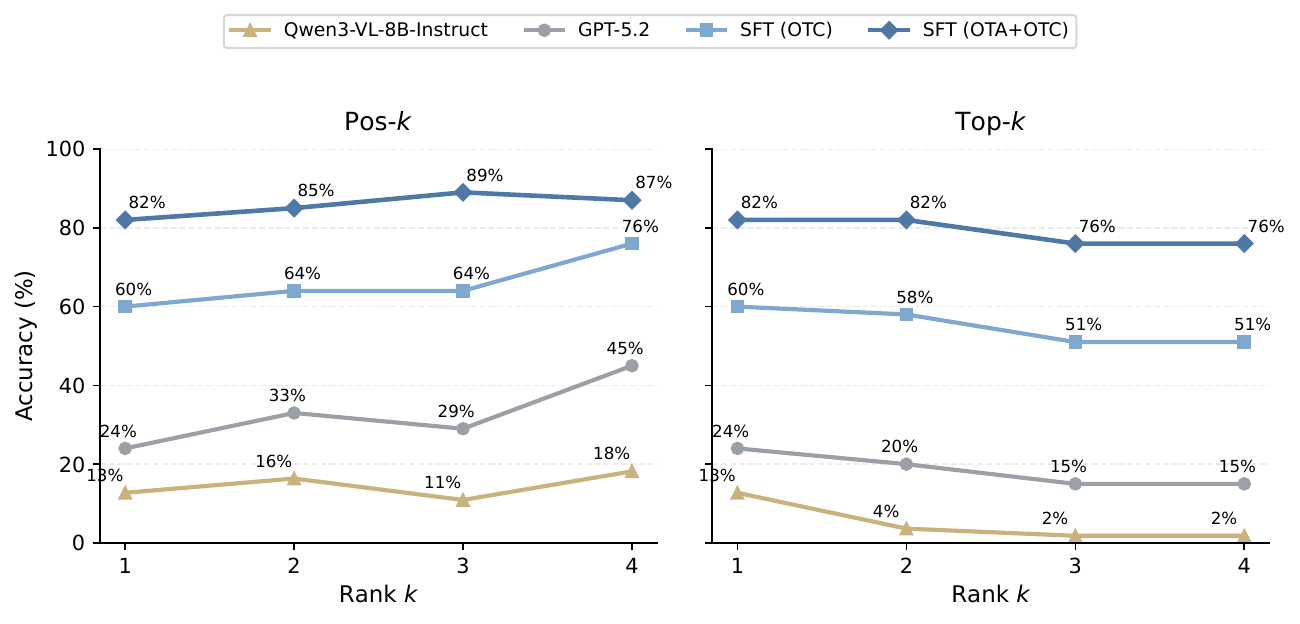}
        \caption*{(b) Detailed ranking consistency on $n4$.}
    \end{minipage}

    \caption{Evaluation of the Scene-OTC test sets. 
    (a) compares overall preference accuracy.
    (b) analyzes ranking consistency with Pos-$k$ and Top-$k$ metrics.}
    \label{fig:13&14}
\end{minipage}
\end{figure}
\endgroup

Fig.~\ref{fig:13&14}(b) further shows strong Pos-$k$ and Top-$k$ accuracy, indicating that our Critic can distinguish multiple quality levels rather than merely identify the best scene. To assess robustness, we shuffle the image-label order five times. As shown in Fig.~\ref{fig:15}, the Critic achieves $90.00\pm2.07\%$ on $n=2$ and $75.64\pm6.88\%$ on $n=4$, closely matching the original results. These findings demonstrate strong scene critique ability and consistent ranking performance, providing reliable preference supervision for Scene-GRPO.

\begingroup
\setlength{\intextsep}{8pt}
\begin{figure}[H]
\centering
\captionsetup{skip=4pt}
\includegraphics[width=0.7\linewidth]{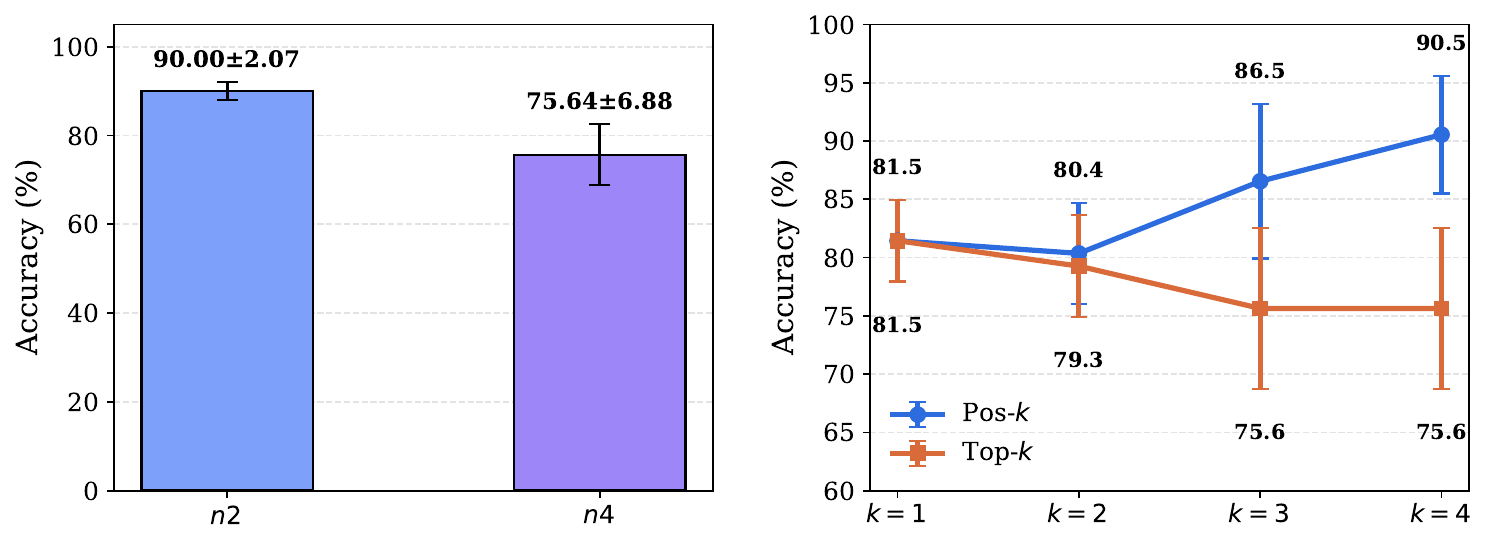}
    \caption{
Image-order robustness of the learned Critic. We independently shuffle the input image-label order five times. Left: accuracy on $n2$ and $n4$ across five randomized runs. Right: Pos-$k$ and Top-$k$ accuracy on $n4$.
    }
    \label{fig:15}
\end{figure}
\endgroup

\noindent\textbf{Design as Complement of Critique}
To further investigate what enables reliable scene evaluation, we analyze ranking consistency on the $n4$ test set. Fig.~\ref{fig:13&14}(b) reports Pos-$k$ and Top-$k$ accuracy, measuring whether the $k$-th best scene is correctly placed and whether the top-$k$ rankings are preserved. Scene-OTA+OTC consistently outperforms other evaluators, achieving 82\% Top-1 and 76\% Top-4 accuracy. These results show that scene critique is not merely a visual comparison task. A reliable Critic should understand both how to design and critique a scene: it must recognize not only which layout is preferred, but also why certain decisions lead to better physical validity, functional correctness, and layout coherence. This suggests that effective scene critique requires understanding the design process itself, with design knowledge serving as an essential complement to critique ability.

\MainBodySideFigure{0.50}{
\centering
\includegraphics[width=\linewidth]{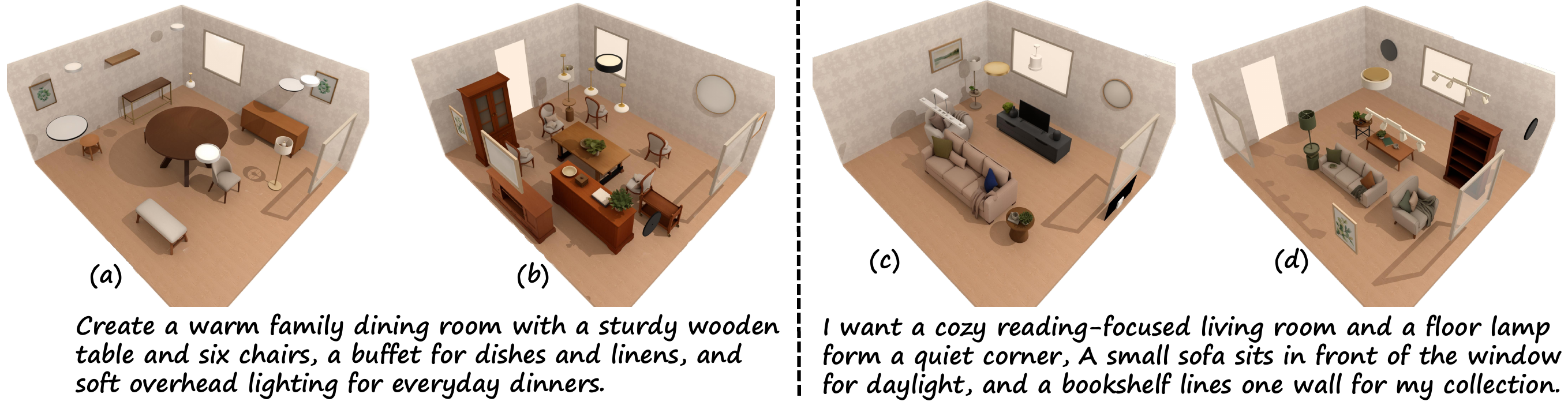}
    \captionof{figure}{
Qualitative comparison of SFT policies with different supervision.
(a,c) Scene-OTA only; (b,d) Scene-OTA+OTC.
}
    \label{fig:16}
}{
\noindent\textbf{Critique as Guidance for Design.}
Beyond evaluation, we naturally ask whether learning scene critique can also benefit scene design. Fig.~\ref{fig:16} compares SFT policies trained with Scene-OTA only and Scene-OTA+OTC under identical instructions. In the dining-room example (Fig.~\ref{fig:16}(a,b)), although the Scene-OTA+OTC policy may contain minor visual imperfections, it better captures the design intent by organizing the table, chairs, buffet, and lighting into a more coherent functional composition. Similarly, in the reading-focused living-room example (Fig.~\ref{fig:9}(c,d)), it establishes clearer relationships among the sofa, floor lamp, and bookshelf, forming a more recognizable reading corner, while the Scene-OTA-only policy mainly satisfies object requirements with weaker functional alignment. These examples reveal an interesting trend: learning scene critique can provide complementary guidance for scene generation beyond construction trajectories alone. By understanding why certain layouts are preferred, the policy can leverage critique knowledge to make more informed design decisions and better translate high-level instructions into coherent spatial organizations and functional compositions.
}

This suggests that design and critique are complementary aspects of design intelligence. A model capable of designing well must understand the principles behind preferred layouts, and these principles can in turn guide it toward creating more coherent and functional scenes.
\par
\begingroup
\titlespacing*{\section}{0pt}{10pt}{6pt}
\section{Conclusion}
In this work, we introduce \textbf{Scene-Designer}, an agentic framework that learns design intelligence for 3D scene generation. We construct \textbf{Scene-OTA} and \textbf{Scene-OTC} to provide complementary supervision for scene design and critique, and propose \textbf{Scene-GRPO} for Critic-guided optimization of design decisions. Extensive experiments demonstrate that Scene-Designer
achieves superior performance in scene generation, while suggesting that design and critique are complementary capabilities.
\par
\endgroup

\bibliographystyle{assets/acl_natbib}
\nocite{*}
\bibliography{aaai2027}
\clearpage
\beginappendix
\addtocontents{toc}{\protect\setcounter{tocdepth}{2}}
\renewcommand{\contentsname}{Appendix Contents}
\setcounter{tocdepth}{2}
% Link each entry to its printed page, independent of section-float anchors.
\begingroup
\let\AppendixOriginalContentsLine\contentsline
\renewcommand{\contentsline}[4]{%
  \AppendixOriginalContentsLine{#1}{#2}{#3}{page.#3}%
}
\tableofcontents
\endgroup
\clearpage

\section{Related Work}

% Right-side comparison table; body text resumes full width below it.
\noindent\begin{minipage}{\textwidth}
\begingroup
\setbox0=\hbox{\begin{minipage}[t]{0.52\textwidth}
\vspace{0pt}
\setlength{\tabcolsep}{3pt}
\centering
\captionof{table}{
Comparison of representative scene generation methods. Existing methods lack a unified framework that combines agentic tool execution, complete scene construction and scene critique ability.
}
\label{tab:method_comparison}
\resizebox{\linewidth}{!}{
\begin{tabular}{lccccc}
\toprule
\textbf{Method} & \textbf{API} & \textbf{Agentic} & \textbf{Tool} & \textbf{Complete} & \textbf{Critique} \\
\midrule
LayoutGPT 
& \cmark & \xmark & \xmark & \xmark & \xmark \\

LayoutVLM 
& \cmark & \xmark & \xmark & \xmark & \xmark \\

Holodeck 
& \cmark & \xmark & \xmark & \cmark & \xmark \\

OptiScene 
& \xmark & \xmark & \xmark & \xmark & \xmark \\

SceneWeaver 
& \cmark & \cmark & \cmark & \xmark & \cmark \\

SceneSmith 
& \cmark & \cmark & \cmark & \cmark & \cmark \\

\textbf{Ours} 
& \xmark & \textbf{\cmark} & \textbf{\cmark} & \textbf{\cmark} & \textbf{\cmark} \\

\bottomrule
\end{tabular}
}

\end{minipage}}
\count255=\dimexpr\ht0+\dp0+2pt\relax
\count254=\baselineskip
\advance\count255 by \count254
\advance\count255 by -1
\divide\count255 by \count254
\noindent\makebox[0pt][l]{\hspace{0.48\textwidth}\raisebox{\ht\strutbox}[0pt][0pt]{\copy0}}%
\hangindent=-0.56\textwidth
\hangafter=-\count255
\noindent\textbf{Foundation Model-based Scene Generation.}
Recent advances in large language models (LLMs) and vision-language models (VLMs) have enabled language-driven indoor scene generation by providing strong semantic understanding and planning capabilities. 
As summarized in Table~\ref{tab:method_comparison}, early approaches such as LayoutGPT~\cite{feng2023layoutgpt}, LayoutVLM~\cite{sun2025layoutvlm}, and Holodeck~\cite{yang2024holodeck} leverage foundation models to translate natural language instructions into structured layouts or complete 3D environments. 
These methods demonstrate the potential of foundation models for bridging language instructions and spatial reasoning, but typically rely on external general-purpose model APIs, resulting in substantial inference costs.
\par
\endgroup
\end{minipage}

Beyond one-shot generation, recent agentic frameworks introduce iterative construction through tool interaction. 
Methods including SceneCraft~\cite{hu2024scenecraft}, FlairGPT~\cite{littlefair2025flairgpt}, SceneTeller~\cite{ocal2024sceneteller}, I-Design~\cite{ccelen2024design}, SceneSmith~\cite{pfaff2026scenesmithagenticgenerationsimulationready}, and SceneWeaver~\cite{yang2026sceneweaver} enable models to progressively modify scenes through intermediate feedback and tool execution. 
However, these agentic approaches still rely on general-purpose models for design decisions and lack learned scene design and scene critique experience. 
Consequently, they may suffer from inefficient construction decisions and lengthy correction trajectories, limiting scalable scene generation.

\noindent\textbf{Learning-based Scene Generation and Dataset Limitations.}
Large-scale 3D datasets have played a central role in advancing indoor scene generation. 
Datasets such as 3D-FRONT~\cite{fu20213d} and 3D-SynthPlace~\cite{yang2026optiscene} provide diverse indoor layouts with object categories, geometries, and spatial relationships, enabling models to learn scene distributions and generate plausible arrangements. 
These datasets have supported a wide range of learning-based approaches, including autoregressive generation~\cite{paschalidou2021atiss,yang2024llplace}, diffusion-based synthesis~\cite{tang2024diffuscene,wei2023lego,yang2024physcene}, and graph-based reasoning~\cite{lin2024instructscene}. 
Such methods improve spatial plausibility by capturing statistical regularities from large-scale scene collections.

However, existing 3D datasets mainly describe static scene states rather than the design process behind them. 
Although they provide object categories, geometries, and spatial relationships, they do not record construction actions, error correction decisions, or quality improvements. 
Therefore, models trained on these datasets can learn what plausible scenes look like, but lack knowledge of how to construct and refine scenes through effective design strategies.

In contrast, our work explores trajectory-based supervision that captures construction decisions, correction operations, and scene improvements. 
This enables models to learn not only scene distributions but also effective strategies for designing and refining scenes.

\noindent\textbf{Scene Evaluation and Critique.}
As scene generation systems become increasingly capable, evaluating generated scenes beyond geometric correctness has become essential. 
Existing evaluation approaches rely on geometric measurements, human preference studies, or VLM-based judgments~\cite{feng2023layoutgpt,yang2024holodeck,yang2026optiscene,yang2026sceneweaver,zhao2026scenerevis}. 
These methods typically evaluate explicit properties such as object placement, instruction compliance, and physical validity.

Recent works including SceneEval~\cite{tam2026sceneeval} and SceneCritic~\cite{sengupta2026scenecritic} explore more structured scene evaluation beyond simple visual judgment. 
SceneEval combines text-scene fidelity metrics with physics-based plausibility checks, while SceneCritic introduces a symbolic evaluator grounded in a structured spatial ontology, verifying semantic coherence, orientation correctness, and geometric consistency through object-level and relationship-level constraints.

However, existing evaluation methods remain limited in capturing holistic design qualities. 
Although symbolic evaluators such as SceneCritic provide reliable assessments for predefined spatial constraints, they rely on explicitly modeled relationships and handcrafted criteria, limiting their ability to assess global functional organization, layout coherence, and human-centered design preferences.

In contrast, our work studies scene critique as a learned design capability rather than a predefined evaluation procedure. 
We train a domain-specific VLM Critic from construction and revision experiences, enabling it to evaluate scenes across physical validity, functional correctness, and layout coherence, and further leverage critique knowledge to guide iterative scene improvement.

\section{Problem Formulation}
\label{app:problem-formulation}

Given a natural-language instruction $U$, our goal is to generate a physically valid, functionally correct, and layout-coherent scene $S$. We formulate scene generation as a multi-turn tool-use process. At step $t$, the policy observes
\begin{equation}
o_t=(I_{t-1},M_t,f_{t-1}),
\end{equation}
where $I_{t-1}$ is the current scene rendering, $M_t$ stores the scene state and interaction history, and $f_{t-1}$ is the feedback from the previous tool call. The policy generates a reasoning trace $h_t$ and an executable tool call $u_t$:
\begin{equation}
a_t=(h_t,u_t)
\sim
\pi_\theta(\cdot|U,o_t),
\quad
(S_t,f_t)=\mathrm{Env}(S_{t-1},u_t).
\end{equation}

Repeating this process forms an Observation--Thought--Action trajectory:
\begin{equation}
T=\big((o_1,a_1),\ldots,(o_Y,a_Y)\big),
\end{equation}
where the final scene is $S=S_Y=S(T)$.

We further introduce a Critic $C_\phi$ to evaluate scene quality. Given an instruction and $N$ candidate scene renderings, the Critic outputs a preference ranking:
\begin{equation}
\sigma
=
C_\phi(U,I^{(1)},\ldots,I^{(N)})
=
(\sigma_1,\sigma_2,...,\sigma_N),
\end{equation}
where $\sigma$ denotes the ranking order from the best to the worst scene, and $\sigma_i$ denotes the index of the scene ranked at position $i$.

\section{Scene-GRPO Reward}
\label{app:scene-grpo-rewards}

This section provides the detailed formulation of the local and global rewards used in Scene-GRPO.

\subsection{Local Rewards}

For the $i$-th candidate branch expanded at editing step $t$, the local reward is defined as
\begin{equation}
r_{t,i}^{\mathrm{local}}
=
r_{t,i}^{\mathrm{fmt}}
+
r_{t,i}^{\mathrm{rank}}
+
r_{t,i}^{\mathrm{imp}}.
\end{equation}

\noindent\textbf{Format Reward.}
The format reward verifies whether the policy output contains a valid reasoning trace and an executable tool call:
\begin{equation}
r_{t,i}^{\mathrm{fmt}}
=
\begin{cases}
1.0,
& \text{if the output format and tool call are valid},\\
0.0,
& \text{otherwise}.
\end{cases}
\end{equation}

\noindent\textbf{Branch-Ranking Reward.}

Let $\mathcal{L}_t$ denote the set of candidate branches that satisfy geometric constraints and are successfully rendered. The frozen Critic ranks their resulting scene renderings:
\begin{equation}
\sigma_t^{\mathrm{rank}}
=
C_\phi\!\left(
U,
\left\{I_t^{(j)}\right\}_{j\in\mathcal{L}_t}
\right).
\end{equation}

Let $p_{t,i}$ denote the position of branch $i$ in the resulting ranking, where $p_{t,i}=0$ denotes the best branch and $p_{t,i}=|\mathcal{L}_t|-1$ denotes the worst. The branch-ranking reward is
\begin{equation}
r_{t,i}^{\mathrm{rank}}
=
\begin{cases}
-0.5,
& i\notin\mathcal{L}_t,\\[4pt]
\dfrac{|\mathcal{L}_t|-1-p_{t,i}}
{|\mathcal{L}_t|-1},
& i\in\mathcal{L}_t,\ |\mathcal{L}_t|\geq 2,\\[10pt]
1.0,
& i\in\mathcal{L}_t,\ |\mathcal{L}_t|=1.
\end{cases}
\end{equation}

Non-renderable branches receive a negative reward, while valid branches receive a normalized reward according to their relative ranking.

\noindent\textbf{Improvement Reward.}

For each renderable branch, the Critic compares the parent rendering $I_{t-1}$ with the child rendering $I_t^{(i)}$:
\begin{equation}
\sigma_{t,i}^{\mathrm{imp}}
=
C_\phi\!\left(
U,
I_{t-1},
I_t^{(i)}
\right).
\end{equation}

The parent and child scenes are indexed by $1$ and $2$, respectively. The improvement reward is defined as
\begin{equation}
r_{t,i}^{\mathrm{imp}}
=
\begin{cases}
+0.5,
& \sigma_{t,i}^{\mathrm{imp}}=(2,1),\\[3pt]
-0.5,
& \sigma_{t,i}^{\mathrm{imp}}=(1,2),\\[3pt]
0.0,
& i\notin\mathcal{L}_t.
\end{cases}
\end{equation}

Therefore, a branch receives a positive reward when the edit improves scene quality and a negative reward when it degrades the parent scene, encouraging meaningful modifications rather than unnecessary edits.

\subsection{Global Rewards}

For each completed trajectory $T_g$, let $Y_g$ denote its number of interaction steps and $n_g$ denote the number of objects in its final scene. The global reward is
\begin{equation}
R_g^{\mathrm{global}}
=
R_g^{\mathrm{len}}
+
R_g^{\mathrm{obj}}.
\end{equation}

\noindent\textbf{Trajectory-Length Reward.}

To encourage efficient yet complete scene construction, we assign higher rewards to trajectories with fewer interaction steps:
\begin{equation}
R_g^{\mathrm{len}}
=
\mathrm{clip}
\left(
\frac{Y_{\mathrm{zero}}-Y_g}
{Y_{\mathrm{zero}}-Y_{\mathrm{full}}},
0,1
\right),
\end{equation}
where $Y_{\mathrm{full}}$ denotes the target number of steps for an efficient and complete trajectory, and $Y_{\mathrm{zero}}$ denotes the step threshold above which no length reward is assigned. In our implementation, we set $Y_{\mathrm{full}}=25$ and $Y_{\mathrm{zero}}=45$ during training.

\noindent\textbf{Object-Completeness Reward.}

To discourage premature termination, we reward trajectories according to the number of objects in the final scene:
\begin{equation}
R_g^{\mathrm{obj}}
=
\mathrm{clip}
\left(
\frac{n_g-n_{\mathrm{zero}}}
{n_{\mathrm{full}}-n_{\mathrm{zero}}},
0,1
\right),
\end{equation}
where $n_{\mathrm{full}}$ denotes the target object count for a complete scene, and $n_{\mathrm{zero}}$ denotes the minimum object count below which no completeness reward is assigned. In our implementation, we set $n_{\mathrm{full}}=25$ and $n_{\mathrm{zero}}=10$ during training.

\subsection{Advantage Estimation}

For the $K$ sampled candidate branches expanded from the same parent scene, the local advantage is computed through group-wise normalization:
\begin{equation}
A_{t,i}^{\mathrm{local}}
=
\frac{
r_{t,i}^{\mathrm{local}}
-
\mu_t^{\mathrm{local}}
}{
\sigma_t^{\mathrm{local}}+\epsilon
},
\end{equation}
where
\begin{equation}
\mu_t^{\mathrm{local}}
=
\frac{1}{K}
\sum_{j=1}^{K}
r_{t,j}^{\mathrm{local}},
\end{equation}
and $\sigma_t^{\mathrm{local}}$ is the corresponding standard deviation. In our implementation, we set $K=4$ during training.

Similarly, for the $G$ completed trajectories generated from the same instruction, the global advantage is
\begin{equation}
A_g^{\mathrm{global}}
=
\frac{
R_g^{\mathrm{global}}
-
\mu^{\mathrm{global}}
}{
\sigma^{\mathrm{global}}+\epsilon
},
\end{equation}
where
\begin{equation}
\mu^{\mathrm{global}}
=
\frac{1}{G}
\sum_{q=1}^{G}
R_q^{\mathrm{global}}.
\end{equation}

In our implementation, we set $G=4$ during training.

The final advantage assigned to branch $i$ at step $t$ along trajectory $g$ is
\begin{equation}
A_{t,i}
=
A_{t,i}^{\mathrm{local}}
+
\eta A_g^{\mathrm{global}}.
\end{equation}

The global advantage is shared across all editing steps within the same trajectory, providing complementary trajectory-level supervision. In our implementation, we set $\eta=1$ during training.

% Allow large appendix figures to share a page with their discussion.
\renewcommand{\topfraction}{0.92}
\renewcommand{\bottomfraction}{0.90}
\renewcommand{\textfraction}{0.06}
\renewcommand{\floatpagefraction}{0.90}

% Side figures use measured heights; text and captions are unchanged.
\newsavebox{\AppendixSideBox}
\newcount\AppendixSideLines
\newlength{\AppendixSideIndent}
\newcommand{\AppendixSideParagraph}[1]{%
  \ifnum\AppendixSideLines>0
    \hangindent=-\AppendixSideIndent
    \hangafter=-\AppendixSideLines
  \fi
  #1\par
  \advance\AppendixSideLines by -\prevgraf
}
\newcommand{\AppendixSideFigure}[4][1]{%
  \FloatBarrier
  \ifnum#1=0\noindent\begin{minipage}{\textwidth}\else\par\fi
  \begingroup
  \setlength{\emergencystretch}{1em}
  \setlength{\AppendixSideIndent}{#2\textwidth}
  \addtolength{\AppendixSideIndent}{0.04\textwidth}
  \sbox{\AppendixSideBox}{\begin{minipage}[t]{#2\textwidth}
    \vspace{0pt}\centering\captionsetup{type=figure} #3
  \end{minipage}}
  \count255=\dimexpr\ht\AppendixSideBox+\dp\AppendixSideBox+2pt\relax
  \count254=\baselineskip
  \advance\count255 by \count254
  \advance\count255 by -1
  \divide\count255 by \count254
  \AppendixSideLines=\count255
  \par\penalty0
  \ifnum#1=1
    \ifdim\dimexpr\ht\AppendixSideBox+\dp\AppendixSideBox+\baselineskip\relax
      >\dimexpr\pagegoal-\pagetotal\relax
      \newpage
    \fi
  \fi
  \noindent\makebox[0pt][l]{%
    \hspace{\dimexpr\textwidth-#2\textwidth\relax}%
    \raisebox{\ht\strutbox}[0pt][0pt]{\usebox{\AppendixSideBox}}}%
  #4
  \ifnum\AppendixSideLines>0
    \vspace{\dimexpr\AppendixSideLines\baselineskip\relax}
  \fi
  \endgroup
  \ifnum#1=0\end{minipage}\fi\par
}

\section{Data Construction}

\subsection{Analysis of Scene Construction Errors, Correction Processes, and Editing Behaviors}

\AppendixSideFigure{0.48}{
\includegraphics[width=\linewidth]{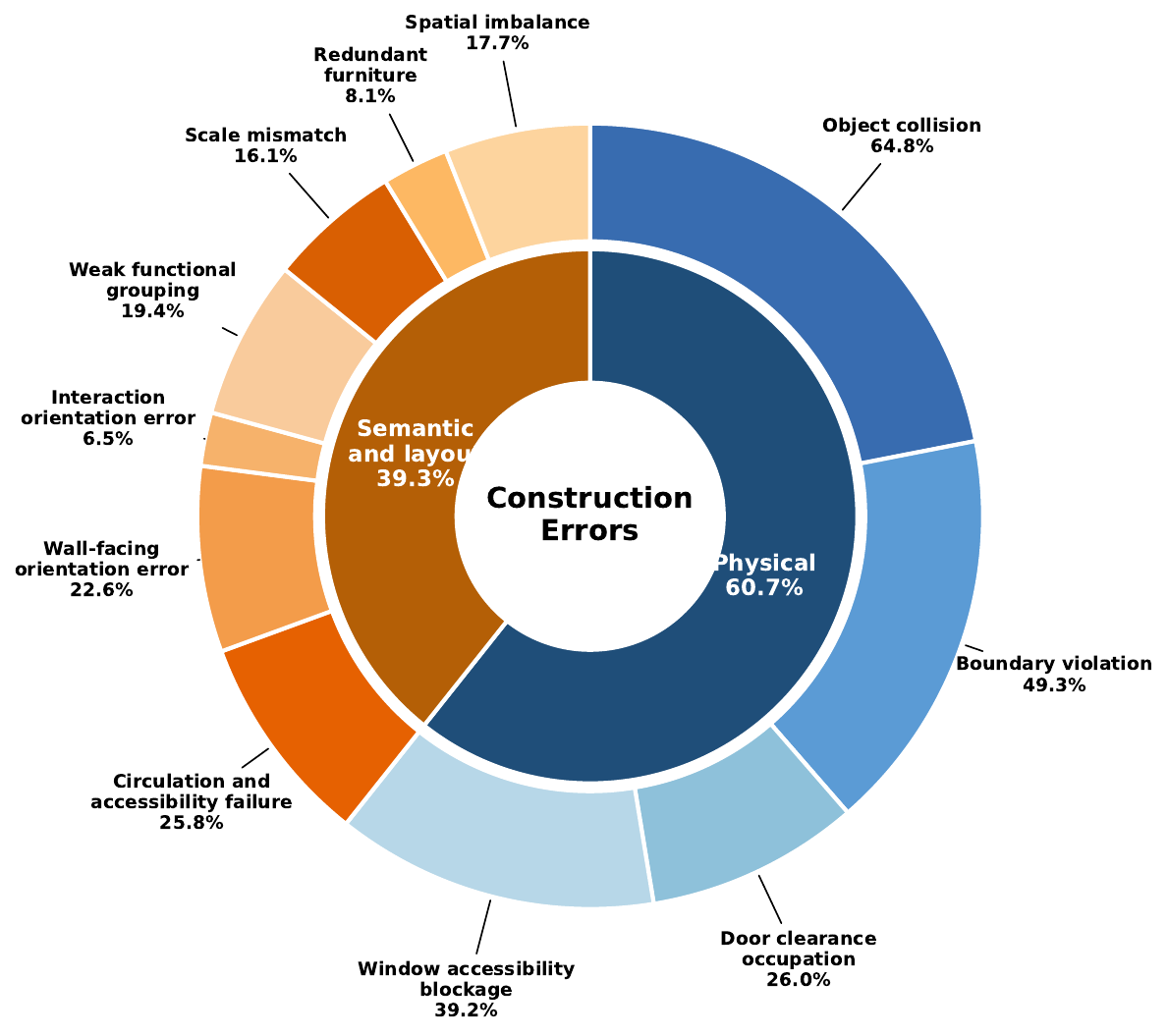}
    \captionof{figure}{Category of Errors Causing Corrections.}
    \label{fig:scene_ota_error_taxonomy}
}{
\AppendixSideParagraph{
\noindent\textbf{Category of Errors Causing Corrections.}
We analyze the types of errors that trigger corrections during scene construction. To understand the challenges faced by existing scene generation models, we examine failure cases and their corresponding correction operations in the original construction trajectories.
}

\AppendixSideParagraph{
We categorize the correction-triggering errors into two major families: \textbf{physical errors} and \textbf{semantic and layout errors}. Physical errors violate explicit geometric or environmental constraints, whereas semantic and layout errors occur when a scene remains physically valid but fails to satisfy higher-level functional or compositional requirements. Fig.~\ref{fig:scene_ota_error_taxonomy} summarizes the distribution of these error categories identified from 100 scenes generated by GPT-5.2. Physical errors account for 60.7\% of all observed errors, while semantic and layout errors account for the remaining 39.3\%.
}
}

Physical errors consist of four categories. \textbf{Object collision} refers to interpenetration between furniture assets caused by inaccurate placement or insufficient spacing. \textbf{Boundary violation} occurs when objects intersect room boundaries due to improper positions, orientations, or scales. \textbf{Door clearance occupation} describes cases in which furniture enters a door-opening or entrance area and obstructs normal access. \textbf{Window accessibility blockage} refers to objects placed in front of windows, restricting lighting, opening, or maintenance space. These errors primarily reflect inaccurate spatial estimation and incomplete satisfaction of geometric and environmental constraints.

Semantic and layout errors consist of seven categories. \textbf{Circulation and accessibility failure} refers to obstructed walking paths or disconnected reachable regions caused by inappropriate furniture placement. \textbf{Wall-facing orientation error} occurs when wall-adjacent furniture exposes its usable side toward the wall rather than the room interior. \textbf{Interaction orientation error} describes interactive objects, such as chairs, facing away from their functional targets. \textbf{Weak functional grouping} refers to related assets that remain individually valid but fail to form a coherent activity zone. \textbf{Scale mismatch} occurs when object dimensions deviate from realistic usage requirements. \textbf{Redundant furniture} describes unnecessary assets that introduce crowding or weaken the intended spatial focus. \textbf{Spatial imbalance} refers to an uneven distribution of furniture that disrupts the overall composition or local spacing relationships.

\AppendixSideFigure{0.56}{
\begin{minipage}{0.49\linewidth}
        \centering
        \includegraphics[width=0.8\linewidth]{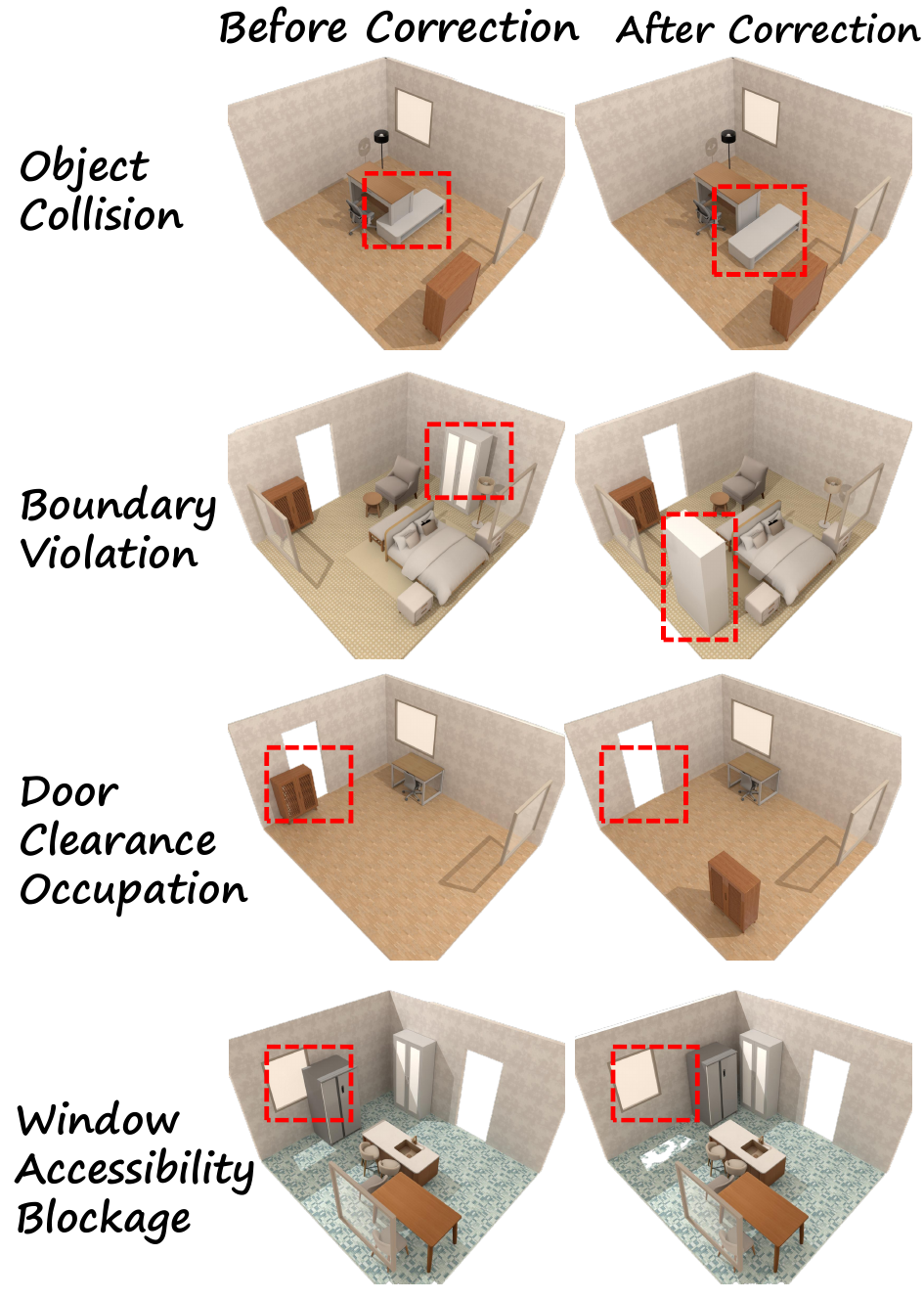}
        \caption*{(a) Representative rework cases for physical errors.}
    \end{minipage}
    \hfill
    \begin{minipage}{0.49\linewidth}
        \centering
        \includegraphics[width=0.8\linewidth]{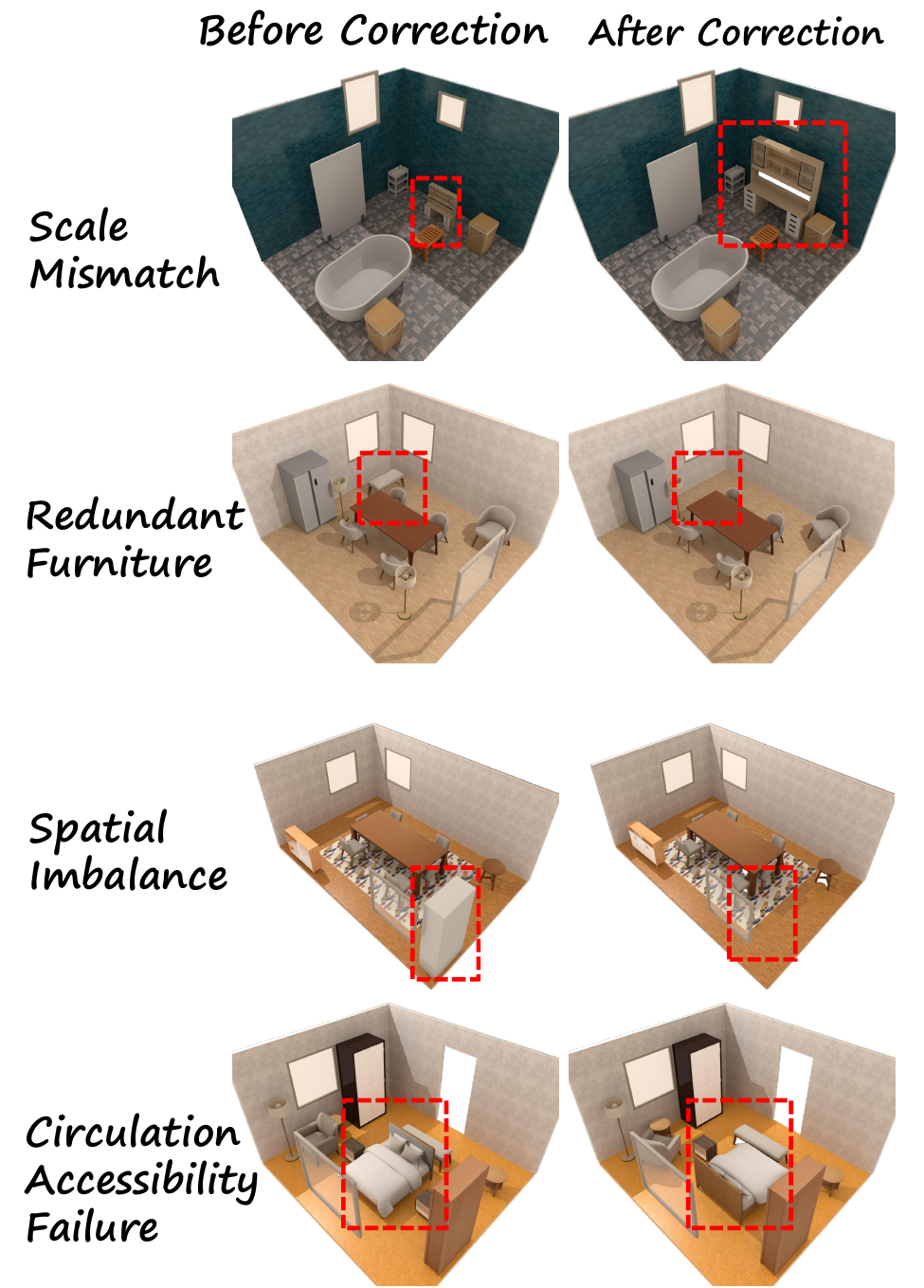}
        \caption*{(b) Representative rework cases for selected semantic and layout errors.}
    \end{minipage}
    \captionof{figure}{Representative rework cases for construction errors.}
    \label{fig:ffig6&ffig8}
}{
\AppendixSideParagraph{
Representative correction cases for the above error categories are presented in Figs.~\ref{fig:ffig6&ffig8} and~\ref{fig:ffig7}, where each case illustrates the scene before and after correction.
}

\AppendixSideParagraph{
As shown in Fig.~\ref{fig:ffig6&ffig8}(a), the object-collision case contains a 4.9\,cm interpenetration between a desk and a table, which is resolved by repositioning the table. In the boundary-violation case, a wardrobe intersects the room boundary and is moved and rotated back into the valid interior region. The door-clearance case shows a cabinet entering the door-opening area; relocating it restores unobstructed access. In the window-accessibility case, a refrigerator blocks a window, and relocating it recovers the usable window area.
}
}

\AppendixSideFigure[1]{0.36}{
\includegraphics[width=\linewidth]{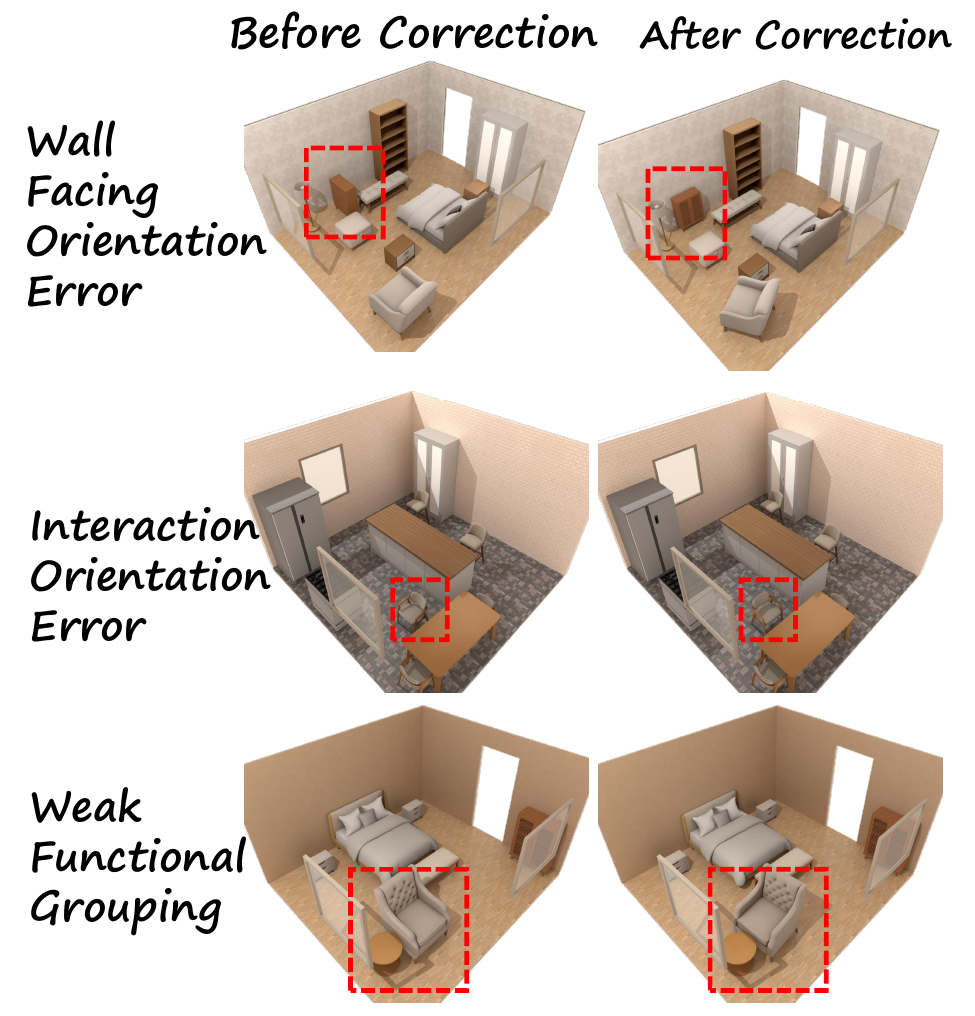}
    \captionof{figure}{Additional representative correction cases for semantic and layout errors.}
    \label{fig:ffig7}
}{
\AppendixSideParagraph{
Figs.~\ref{fig:ffig6&ffig8}(b) and~\ref{fig:ffig7} present representative semantic and layout corrections. In the circulation and accessibility case, the bed, nightstand, and reading-corner furniture constrain the central passage and divide the room into disconnected reachable regions. Reorganizing the bedside and reading areas and removing the redundant nightstand restore continuous circulation.
}

\AppendixSideParagraph{
In the wall-facing orientation case, a dresser is placed against the wall with its functional side facing away from the room interior; rotating and realigning the dresser restore its intended usage direction. The interaction-orientation case shows a dining chair facing away from the dining table, which is corrected by rotating the chair by approximately $90^\circ$.
}

\AppendixSideParagraph{
In the weak functional-grouping case, an armchair and side table initially fail to form a coherent reading area, whereas adjusting the chair orientation establishes a clearer functional relationship. The scale-mismatch case contains an unrealistically small double vanity, which is enlarged by a factor of 2.2 to obtain a plausible and usable size.
}

\AppendixSideParagraph{
In the redundant-furniture case, an additional armchair and bench weaken the dining table as the primary spatial focus; removing unnecessary furniture produces a cleaner composition. Finally, the spatial-imbalance case places a sideboard and display cabinet along the same side of the dining room, creating excessive visual weight. Removing the display cabinet and repositioning the sideboard yield a more balanced layout.
}
}

\noindent\textbf{Analysis of Correction.}
We analyze correction behaviors from two complementary perspectives: how frequently different errors occur in construction trajectories and how many corrective moves are required to resolve them.

\AppendixSideFigure{0.48}{
\includegraphics[width=\linewidth]{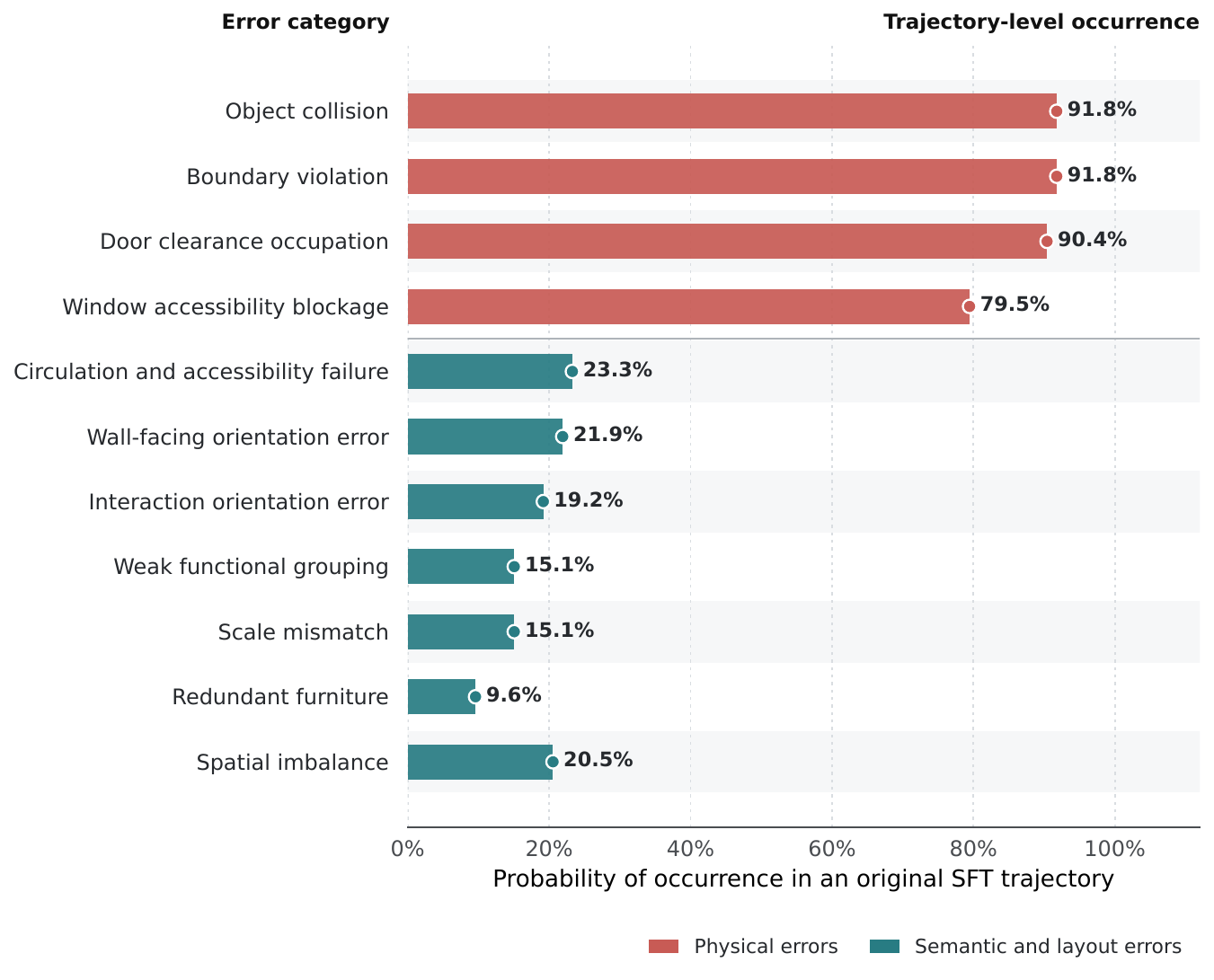}
    \captionof{figure}{Trajectory-level occurrence probability of each error category in the original trajectories.}
    \label{fig:trajectory_error_probabilities}
}{
\AppendixSideParagraph{
Fig.~\ref{fig:trajectory_error_probabilities} reports the trajectory-level occurrence frequency of each error category across the 100 scenes generated by GPT-5.2. We observe that construction errors occur frequently during scene generation. Physical errors appear in the majority of trajectories, with object collision, boundary violation, and door clearance occupation occurring in 91.8\%, 91.8\%, and 90.4\% of trajectories, respectively. Window accessibility blockage also occurs in 79.5\% of trajectories. Semantic and layout errors are less frequent but still appear regularly, including circulation and accessibility failure (23.3\%), wall-facing orientation error (21.9\%), spatial imbalance (20.5\%), and interaction orientation error (19.2\%).
}

\AppendixSideParagraph{
Importantly, our goal is not to require models to completely avoid all construction errors, as iterative refinement is an inherent part of scene design. Instead, effective design intelligence requires not only recognizing construction errors but also identifying efficient correction strategies. Since the same error can often be resolved through different numbers of corrective moves, correction efficiency directly affects the overall trajectory length.
}
}

Fig.~\ref{fig:ffig9} further reports the number of corrective moves required for each fine-grained error type. For each category, we measure the proportion of cases resolved with different numbers of corrective operations. The results reveal that even relatively straightforward errors can require excessive corrective operations before being resolved.

\AppendixSideFigure[1]{0.48}{
\includegraphics[width=\linewidth]{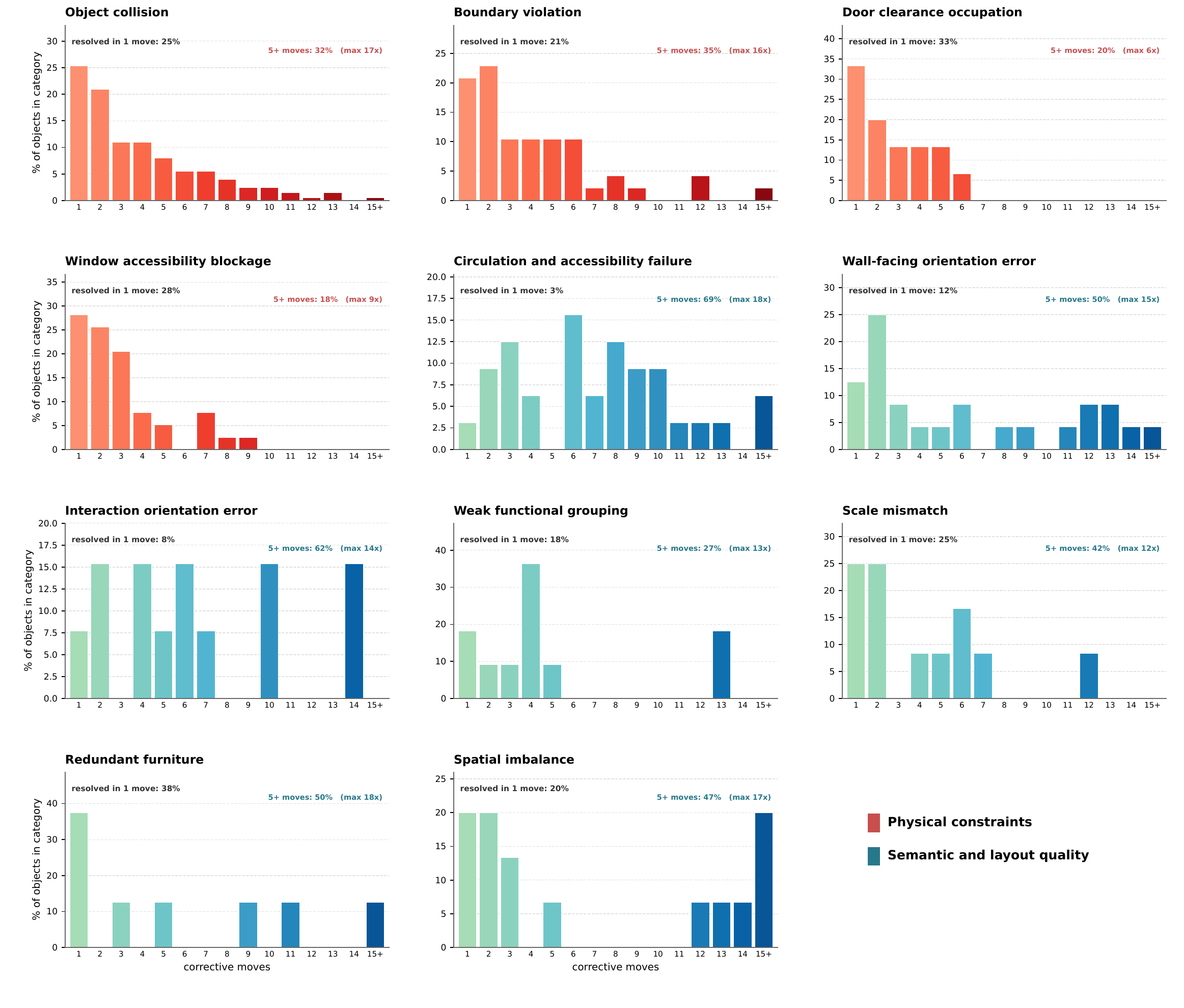}
    
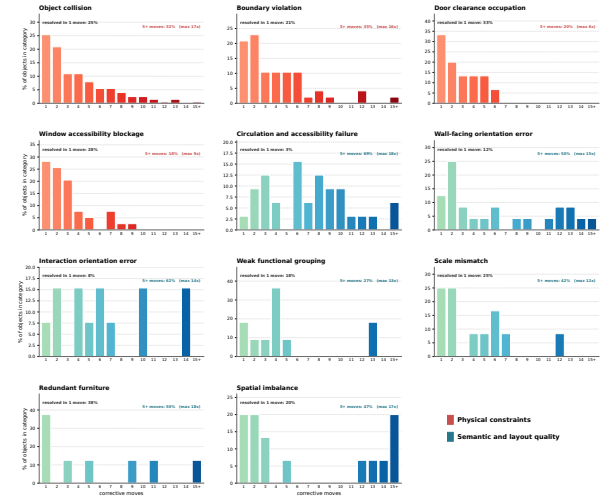
\captionof{figure}{Distribution of corrective moves required for each error category in the original trajectories.}
    \label{fig:ffig9}
}{
\AppendixSideParagraph{
We further analyze the number of corrective moves required for each error category. We focus on representative errors whose solutions often involve straightforward local modifications. For example, a door clearance occupation can often be resolved by directly moving the obstructing object away from the doorway. Nevertheless, only 33\% of such cases are corrected in a single move, while 20\% require at least five corrective moves. Similarly, a window accessibility blockage usually requires only relocating the object that blocks access to the window, yet 18\% of cases still involve five or more corrective moves.
}

\AppendixSideParagraph{
Orientation errors exhibit even greater inefficiency. These errors can often be corrected through directly rotating the affected object toward the room or its interaction target. However, only 12\% and 7.5\% of wall-facing orientation errors and interaction orientation errors are resolved in one move, respectively, while 50\% and 62\% require at least five moves. A similar pattern appears for scale mismatch, where resizing the affected object is often sufficient, yet 42\% of cases still require five or more corrective moves. More complex semantic and layout errors are even harder to resolve: 69\% of circulation and accessibility failures and 47\% of spatial imbalance cases involve at least five corrective moves.
}

\AppendixSideParagraph{
These results indicate that GPT-5.2 frequently fails to identify and execute the most direct correction strategy. Instead, it performs repeated edits of object positions, orientations, or scales before reaching a satisfactory solution. Since inefficient correction chains substantially extend generation trajectories, frequently occurring errors combined with ineffective corrections become a major source of unnecessary interaction. Therefore, although the model can recognize many construction errors, it lacks effective correction strategies and requires excessive corrective operations even for relatively straightforward fixes. Such inefficient trajectories also increase inference cost and latency when using closed-source model APIs.
}

}

\begin{figure}[H]
    \centering
    \includegraphics[width=0.9\columnwidth]{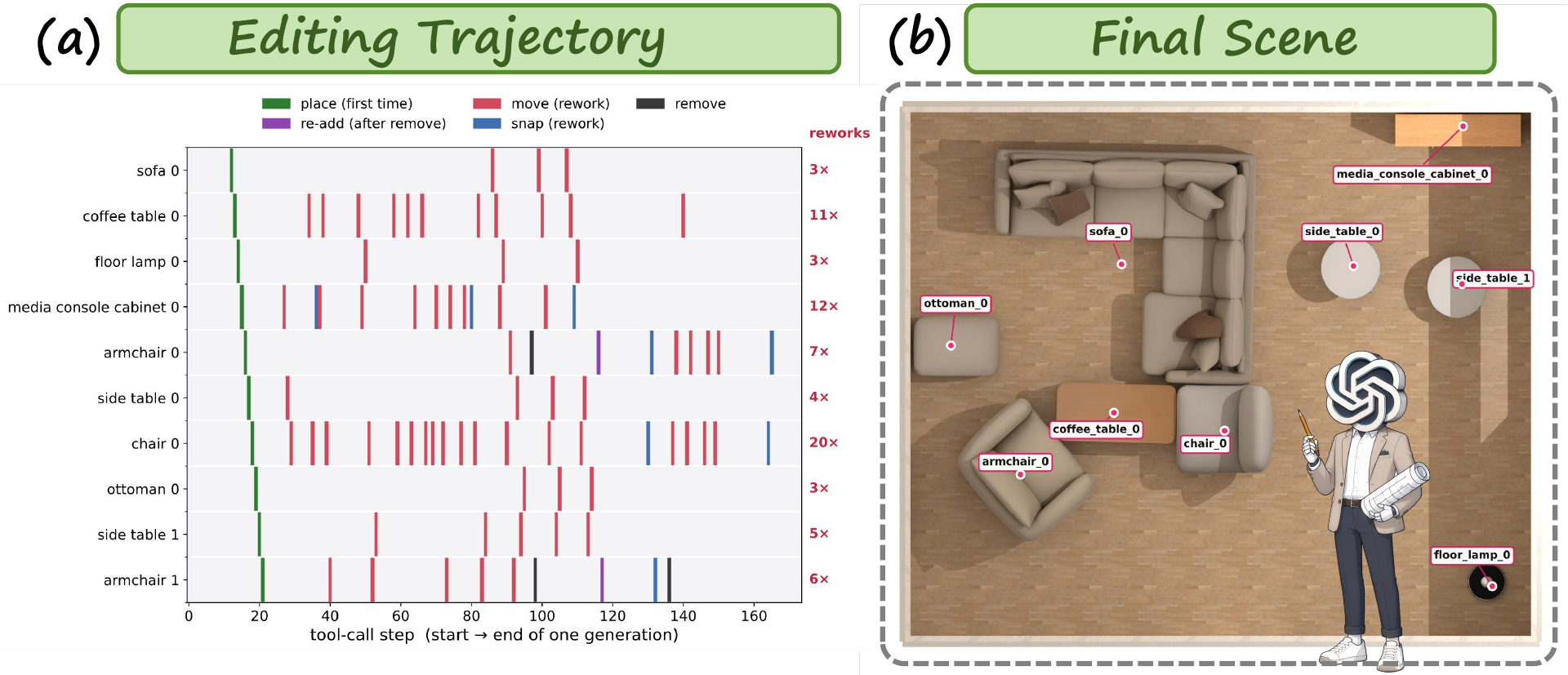}
    \caption{
        A representative asset-level editing trajectory of a GPT-5.2-generated scene.
        (a) The editing trajectory of each major asset over the generation process, where different colors indicate different editing operations, including initial placement and subsequent reworks.
        (b) The final generated scene corresponding to the trajectory in (a), with major assets highlighted.
        }
    \label{fig:ffig10}
\end{figure}

\AppendixSideFigure[1]{0.48}{
\includegraphics[width=\linewidth]{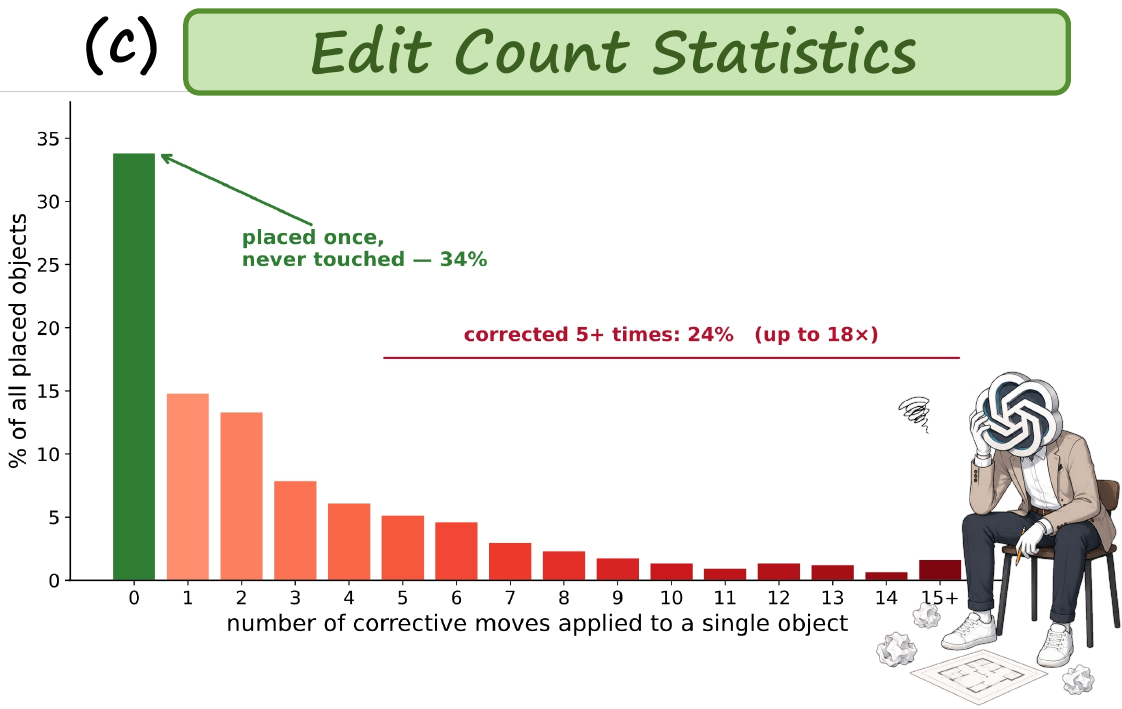}
    \captionof{figure}{
        Statistics of asset editing frequency in GPT-5.2-generated construction trajectories.
        The distribution shows that while many assets are placed once and never modified, a considerable proportion of assets undergo repeated corrections, with some assets requiring more than five edits.
        }
    \label{fig:ffig11}
}{
\AppendixSideParagraph{
\noindent\textbf{Analysis of Editing Behaviors.}
To further understand the underlying causes of inefficient corrections, we analyze how GPT-5.2 modifies scene assets during construction. Fig.~\ref{fig:ffig10} visualizes the editing trajectory of individual assets in a representative scene, where each node denotes an asset state and each edge represents an editing operation. We observe that the model often revisits the same assets multiple times before reaching the final configuration, indicating that many operations correspond to exploratory adjustments rather than effective correction decisions.
}

\AppendixSideParagraph{
For example, Fig.~\ref{fig:ffig10}(b) presents a representative scene generated by GPT-5.2, and Fig.~\ref{fig:ffig10}(a) visualizes the editing trajectories of the major assets in this scene. We observe that several assets undergo repeated modifications before reaching their final states. Instead of applying a targeted correction after identifying the underlying error, the model repeatedly modifies the same assets through trial and error to explore different configurations. This behavior indicates that GPT-5.2 often lacks efficient correction strategies and relies on repeated asset-level editing to gradually approach a satisfactory scene.
}

\AppendixSideParagraph{
We further summarize the editing frequency of assets across generated trajectories in Fig.~\ref{fig:ffig11}. The distribution reveals a highly imbalanced editing pattern: while 34\% of assets are placed once and never modified, a considerable proportion of assets undergo repeated corrections. Specifically, 24\% of assets require five or more corrective moves, with some assets being edited up to 18 times. These repeatedly modified assets contribute disproportionately to the overall correction operations and substantially extend the trajectory length. This observation suggests that the main bottleneck is not simply the occurrence of construction errors, but the model's inability to identify effective correction strategies and avoid repeated edits on the same assets.
}

}

\subsection{Scene-OTA Construction}
\label{app:scene-ota}

\noindent\textbf{Overview.}
Scene-OTA is constructed from raw scene-construction trajectories generated by GPT-5.2 with the SceneSmith~\cite{pfaff2026scenesmithagenticgenerationsimulationready} scaffold. Since raw trajectories contain many trial-and-error operations, we introduce Causal Trajectory Consolidation to remove ineffective corrections while preserving effective repair strategies. Given an original trajectory $T$, our objective is to obtain a shorter subsequence $T'$ satisfying:

\begin{equation}
T'\subseteq T,\quad S(T')=S(T),
\end{equation}

where $S(T)$ denotes the final scene state produced by trajectory $T$. Different from fixed-ratio trajectory shortening, our consolidation procedure evaluates each operation according to its contribution to scene improvement and error correction.

\noindent\textbf{Trajectory State Reconstruction.}
Before consolidation, we reconstruct intermediate scene states from tool interactions. Each asset is represented by its geometric state:

\begin{equation}
s_j=(x_j,y_j,\theta_j,r_j),
\end{equation}

where $(x_j,y_j)$ denotes the planar position, $\theta_j$ denotes the rotation angle, and $r_j$ denotes the scale factor of asset $j$.

For each successful tool operation, we update the corresponding asset state. Specifically, move operations update the asset position and orientation, snap operations update the resulting position and orientation after applying the spatial relation, and rescale operations update the scale factor. Failed tool calls are ignored since they do not modify the actual scene state.

\noindent\textbf{Error-aware Candidate Identification.}
We first identify potentially ineffective operations using two complementary signals: geometric progress toward the final state and reduction of construction errors.

Given an operation $a_i$, we compare the scene states before and after the operation with the final accepted scene state $S_f$. Here, $S_i$ denotes the complete scene state after operation $i$, which consists of all asset states. Specifically, we define the state progress as:

\begin{equation}
\Delta_i^{state}
=
D(S_{i-1},S_f)-D(S_i,S_f),
\end{equation}

where $S_{i-1}$ and $S_i$ denote the scene states before and after operation $i$, respectively. 
$D(\cdot,\cdot)$ measures the difference between two scene states.

For each asset state, we first define the asset-level distance as:

\begin{equation}
D_a(s_i,s_f)
=
\lambda_p d_p(s_i,s_f)
+
\lambda_\theta d_\theta(s_i,s_f),
\end{equation}

where $d_p$ is the Euclidean distance between asset positions:

\begin{equation}
d_p(s_i,s_f)
=
\sqrt{(x_i-x_f)^2+(y_i-y_f)^2},
\end{equation}

and $d_\theta$ measures the angular difference:

\begin{equation}
d_\theta(s_i,s_f)
=
|\mathrm{wrap}(\theta_i-\theta_f)|.
\end{equation}

The scene-level distance is computed by aggregating the distances of all assets:

\begin{equation}
D(S_i,S_f)
=
\sum_j D_a(s_{i,j},s_{f,j}),
\end{equation}

where $s_{i,j}$ and $s_{f,j}$ denote the states of asset $j$ before and after consolidation.

A positive $\Delta_i^{state}$ indicates that the operation moves the scene closer to its final configuration, while a negative value indicates that it deviates from the final state.

In addition to geometric progress, we evaluate whether an operation reduces existing construction errors:

\begin{equation}
\Delta_i^{error}=V_{i-1}-V_i,
\end{equation}

where $V_i$ denotes the number of remaining construction errors after operation $i$. The error set includes both physical errors and semantic and layout errors. Physical errors are obtained from environment feedback, which provides explicitly detectable constraint violations, including object collisions, boundary violations, door clearance occupation, and window accessibility blockage. Semantic and layout errors are identified through scene evaluation feedback.

An operation that neither improves the final state distance nor reduces existing construction errors is regarded as a suspicious trial-and-error operation and becomes a candidate for further causal evaluation.

\noindent\textbf{Correction-Effect Evaluation.}
The above rule-based criteria are designed to identify potentially ineffective operations with high recall, but they cannot determine whether an operation is truly unnecessary. Some operations may temporarily increase the distance to the final state while establishing intermediate configurations required by subsequent corrections. Therefore, we further employ GPT-5.2 as a causal verifier to evaluate the contribution of each candidate operation.

For each candidate operation, GPT-5.2 receives the following information: 
(1) the original room instruction; 
(2) the scene states before and after the candidate operation; 
(3) the final accepted scene state; 
(4) the operation description and tool parameters; 
(5) surrounding operations before and after the candidate operation; 
(6) available environment feedback or semantic scene-evaluation feedback; and 
(7) rule-based screening signals, including final-state progress $\Delta_i^{state}$, segment-level error reduction $\Delta_k^{error}$, operation reversal, and immediate overwrite. We additionally provide the definitions of construction-error categories, enabling the verifier to associate each operation with the corresponding error being corrected.

Given these inputs, the verifier evaluates whether each operation makes a meaningful contribution to the correction process. Specifically, each candidate operation is assigned one of three labels:

\begin{equation}
y_i \in
\left\{
\begin{aligned}
&\texttt{redundant\_trial},\\
&\texttt{useful\_correction},\\
&\texttt{uncertain}.
\end{aligned}
\right.
\end{equation}

An operation is classified as \texttt{redundant\_trial} when it provides no meaningful contribution to resolving the identified error and can be removed without affecting the final correction result. Typical cases include true no-operations, ineffective adjustments that are later reversed, intermediate placements immediately overwritten by subsequent absolute placements, or operations unrelated to the reported construction error. However, reversal or overwrite alone is insufficient for removal: an operation is retained if it reduces an existing error or provides useful intermediate correction experience.

An operation is classified as \texttt{useful\_correction} when it resolves or substantially reduces an existing error, establishes an intermediate state required by later operations, or forms a necessary component of a coordinated multi-object correction. In particular, an operation remains useful even when a later operation further refines its result. Operations whose contribution cannot be reliably determined, for example when the effect of multiple operations cannot be isolated, are labeled as \texttt{uncertain} and conservatively retained.

The verifier returns a structured response containing the predicted label, confidence score, and a brief causal explanation for each candidate operation. It also identifies operations recommended for retention or removal. An operation is considered removable only when it is classified as \texttt{redundant\_trial} with confidence higher than $\tau$. The verifier output is treated as a deletion proposal rather than a final decision; protected operations and operations with downstream dependencies are retained regardless of the predicted label.

\noindent\textbf{Causal Trajectory Consolidation.}
After causal evaluation, we consolidate the original trajectory by removing only operations identified as unnecessary while preserving causally important correction decisions. Specifically, an operation $a_i$ is removed only when it satisfies three conditions: (1) it is identified as a suspicious candidate in the error-aware candidate identification stage; (2) it is classified as \texttt{redundant\_trial} by the causal verifier; and (3) its confidence score is higher than the threshold $\tau$. Formally, the removal set is defined as:

\begin{equation}
\mathcal{R}
=
\left\{
a_i
\mid
C_i=1,\;
y_i=\texttt{redundant\_trial},\;
p_i\geq\tau
\right\},
\end{equation}

where $C_i$ indicates whether operation $a_i$ is selected as a suspicious candidate by the rule-based screening stage, $y_i$ denotes the verifier prediction, and $p_i$ denotes the verifier confidence. In our implementation, we set $\tau=0.8$.

To avoid removing necessary construction decisions, we apply conservative constraints during consolidation. Operations that modify asset existence, including object addition and removal, are always retained. We also preserve operations that reduce construction errors, establish intermediate states required by subsequent operations, or contribute to coordinated multi-object corrections. When multiple operations modify the same asset, the final effective correction is retained. After consolidation, we manually review the trajectories to verify their quality and remove remaining inconsistent cases.

The consolidated trajectory is obtained by removing the selected operations while preserving the temporal order of the remaining operations:

\begin{equation}
T'=\mathrm{Remove}(T,\mathcal{R}).
\end{equation}

\AppendixSideFigure[1]{0.42}{
\includegraphics[width=0.985\linewidth]{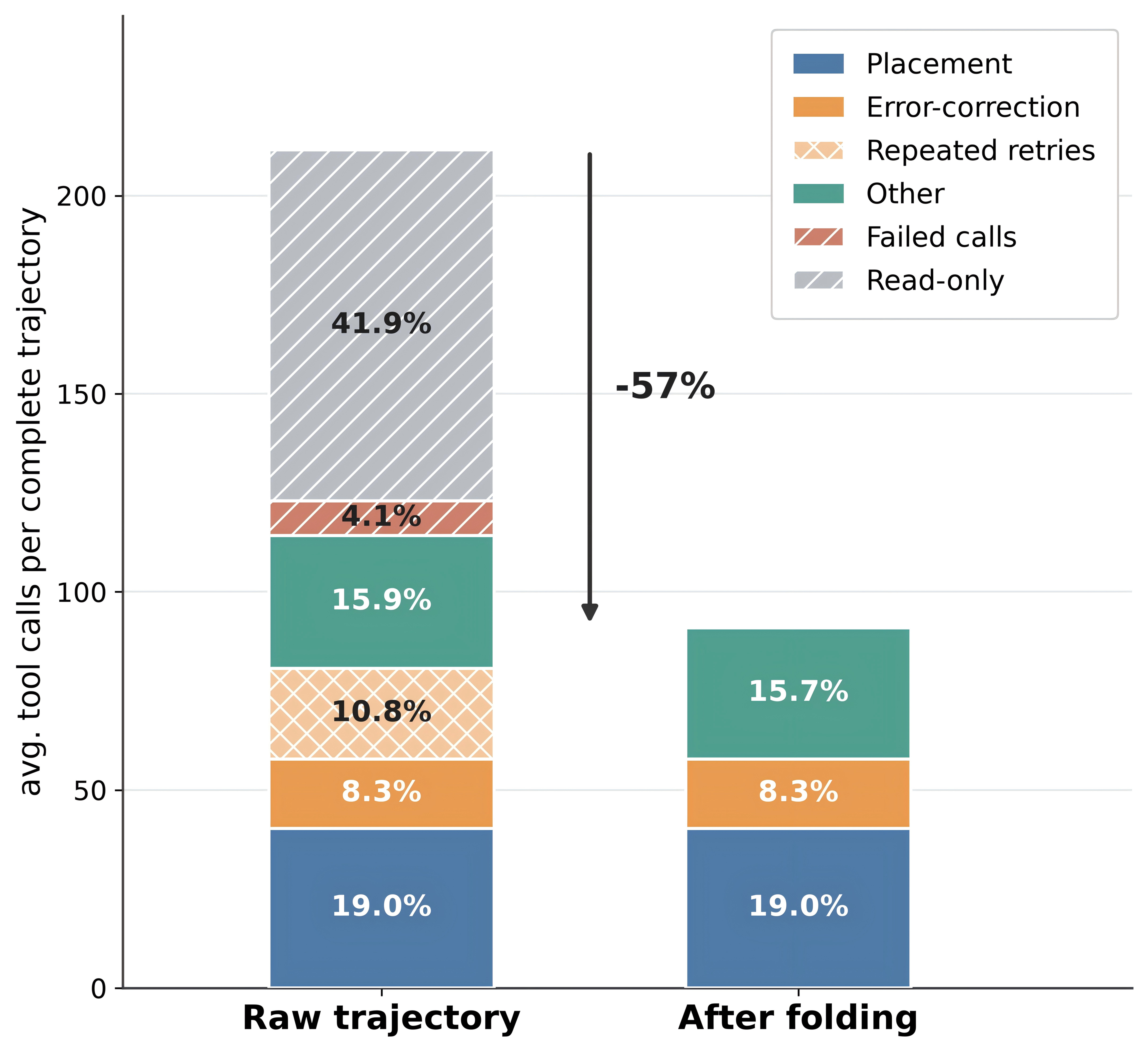}
    \captionof{figure}{
        Effect of Causal Trajectory Consolidation on trajectory composition.
        The consolidation removes read-only calls, failed calls, and repeated retries while preserving essential placement and effective correction operations, reducing the average number of tool calls per trajectory by 57\%.
        }
    \label{fig:ffig12}
}{
\AppendixSideParagraph{
For each removed operation, its corresponding tool call and environment feedback are removed together, while the remaining operations preserve the original temporal order. The consolidated trajectory is finally verified through replay to ensure that it produces the same final scene state as the original trajectory.
}

\AppendixSideParagraph{
\noindent\textbf{Consolidation Statistics.}
To analyze the effect of Causal Trajectory Consolidation, we compare the composition of raw and consolidated trajectories in Fig.~\ref{fig:ffig12}. Beyond removing ineffective trial-and-error corrections, consolidation eliminates several types of low-value operations, including read-only calls that do not modify the scene state and failed tool calls that produce no effective changes. For error-recovery trajectories, we preserve the operation that introduces the error and the final effective correction while removing unnecessary intermediate repair attempts. Efficient correction trajectories remain unchanged.
}

\AppendixSideParagraph{
As shown in Fig.~\ref{fig:ffig12}, Causal Trajectory Consolidation reduces the average number of tool calls per trajectory by 57\%. Specifically, read-only calls, which account for 41.9\% of raw trajectory steps, and repeated retries caused by trial-and-error behaviors are removed, while essential placement operations and effective error-correction operations are preserved. These results demonstrate that consolidation does not simply shorten trajectories, but selectively removes ineffective behaviors while retaining the causal decisions required for successful scene construction.
}
}

To illustrate the design experience captured by Scene-OTA, Fig.~\ref{fig:sceneota_case} presents a representative consolidated construction trajectory. The trajectory records not only the initial scene planning decisions but also intermediate feedback and effective correction decisions, showing how the model progressively improves the scene through iterative design.

\subsection{Scene-OTC Construction}
\label{app:scene-otc}

\noindent\textbf{Overview.}
Scene-OTC provides preference supervision for scene critique learning by applying controlled perturbations to high-quality scenes in Scene-OTA. 
Instead of binary labels, it creates degraded variants with different design qualities, enabling the Critic to learn fine-grained preferences and understand why one scene is preferred over another.
 
We organize perturbations into three increasing levels of severity:
\ding{182}\textbf{mild layout deviations}, which preserve physical validity and functionality but reduce visual harmony or spatial balance;
\ding{183}\textbf{functional degradation}, which disrupts object relations, interaction orientations, or functional grouping while generally maintaining physical validity;
and
\ding{184}\textbf{severe physical violations}, which introduce explicit geometric failures such as collisions and out-of-bound placement.

The perturbation categories and corresponding operations are summarized in Tab.~\ref{tab:perturbation_categories}, and detailed definitions are provided in Tab.~\ref{tab:perturbation_details}.

\begin{table}[H]
\centering
\small
\setlength{\tabcolsep}{3pt}

\begin{tabular}{c|c|l}
\toprule
Severity & Degradation Type & Perturbation Functions \\
\midrule

Tier 1 
& Physical violation 
& \begin{tabular}{l}
move\_blocking\_door,\\
move\_into\_wall, \\
move\_overlap,\\
 rotate\_out\_of\_room,\\
rotate\_into\_collision
\end{tabular}
\\

\midrule

Tier 2 
& Functional degradation 
& \begin{tabular}{l}
face\_wall, rotate\_180,\\
remove\_object, swap\_two
\end{tabular}
\\

\midrule

Tier 3 
& Layout deviation 
& \begin{tabular}{l}
move\_off\_wall,\\
 move\_far\_from\_focal,\\
move\_to\_corner,\\
 rotate\_45\_offset
\end{tabular}
\\

\bottomrule
\end{tabular}

\caption{Overview of perturbation categories in Scene-OTC.}
\label{tab:perturbation_categories}

\end{table}

\begin{table*}[!h]
\centering
\small
\setlength{\tabcolsep}{4pt}
\caption{Detailed perturbation strategies used in Scene-OTC.}
\label{tab:perturbation_details}

\begin{tabular}{c|l|p{0.55\textwidth}}
\toprule
Severity & Perturbation & Description \\
\midrule

\multirow{5}{*}{Tier 1}
& move\_into\_wall 
& Move an object across the room boundary, causing partial or complete out-of-bound placement.
\\
& move\_blocking\_door
& Move an object into the door clearance region, reducing accessibility and blocking movement paths.
\\
& move\_overlap
& Move an object to overlap with another furniture item, creating physical collision.
\\
& rotate\_out\_of\_room
& Rotate an object so that its geometry exceeds the room boundary.
\\
& rotate\_into\_collision
& Rotate an object to introduce collisions with surrounding objects.
\\

\midrule

\multirow{4}{*}{Tier 2}
& face\_wall
& Rotate an oriented object toward the wall, violating its intended functional orientation.
\\
& rotate\_180
& Reverse the original orientation of an object by 180 degrees.
\\
& remove\_object
& Remove an object from the scene, reducing object completeness.
\\
& swap\_two
& Exchange the poses of two objects, disrupting their functional arrangement.
\\

\midrule

\multirow{4}{*}{Tier 3}
& move\_off\_wall
& Move wall-aligned furniture away from the wall, weakening spatial organization.
\\
& move\_far\_from\_focal
& Move an object away from its functional partner, breaking intended object relationships.
\\
& move\_to\_corner
& Move an object into a corner, producing less reasonable spatial distribution.
\\
& rotate\_45\_offset
& Apply an unnatural rotation offset while maintaining physical validity.
\\

\bottomrule
\end{tabular}

\end{table*}

\noindent\textbf{Preference Annotation and Reasoning Generation.}
For each perturbation strategy, we construct preference examples containing $N$ scene renderings. The preference order of accumulating perturbations and different-strength perturbations is automatically determined according to the perturbation severity. For the same perturbation applied to different objects, the preference order is manually annotated because the impact depends on the functional role of each object in the scene.

Given the scene renderings and the corresponding instruction, GPT-5.2 is further used to generate reasoning traces that explain the quality differences and produce complete preference rankings from best to worst. Each example follows the Observation--Thought--Critic format, providing supervision for the model to identify and explain fine-grained differences in physical validity, functional correctness, and layout coherence.

\begin{figure}[!htbp]
    \centering
    \begin{minipage}{0.48\columnwidth}
        \centering
        \includegraphics[width=\linewidth]{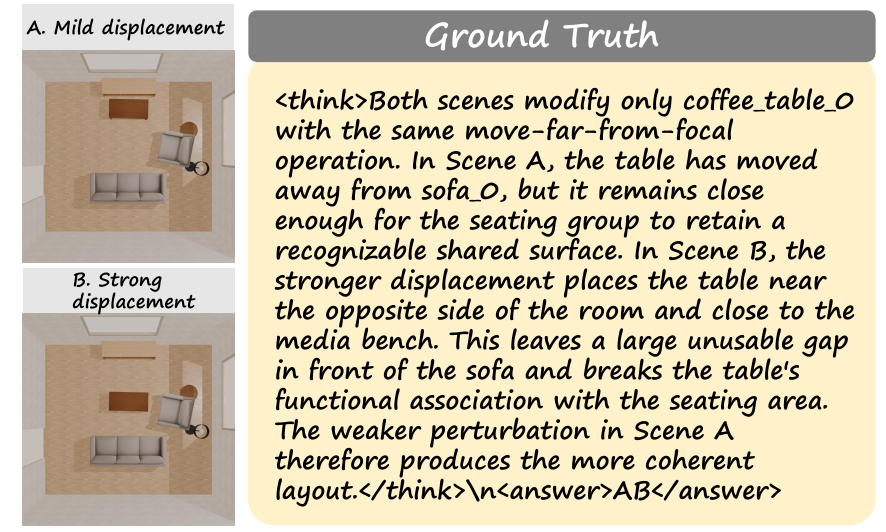}
        \caption*{(b) Different-strength perturbations}
    \end{minipage}
    \hfill
    \begin{minipage}{0.48\columnwidth}
        \centering
        \includegraphics[width=\linewidth]{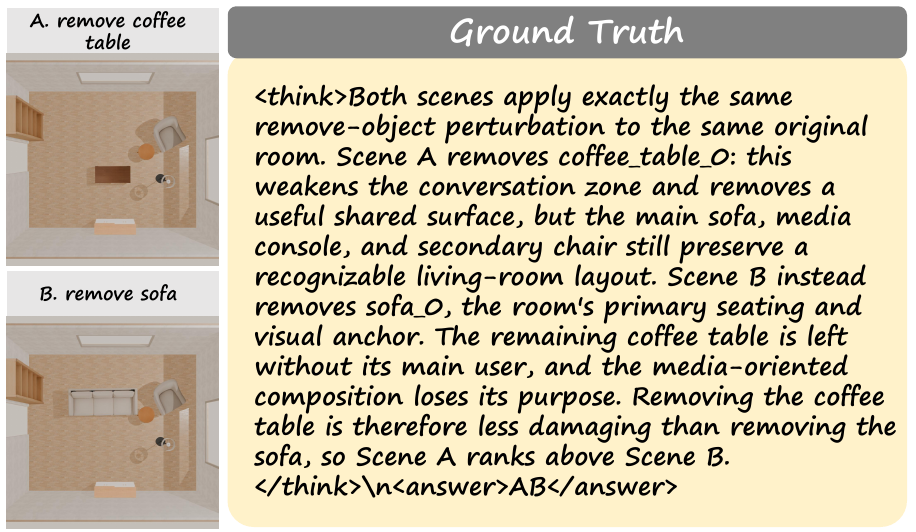}
        \caption*{(c) Same perturbation on different objects}
    \end{minipage}

    \vspace{0.4em}

    \begin{minipage}{0.8\columnwidth}
        \centering
        \includegraphics[width=\linewidth]{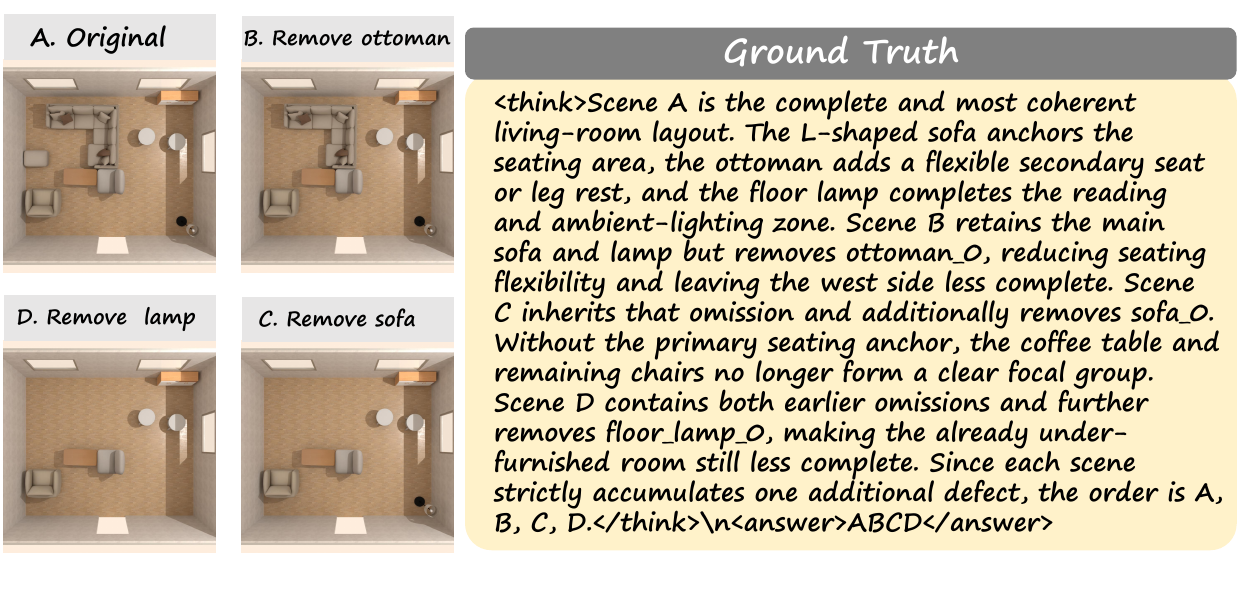}
        \caption*{(a) Accumulating perturbations}
    \end{minipage}

    \caption{
        Representative perturbation strategies in Scene-OTC.
        (a) Accumulating perturbations progressively degrade scene quality by introducing additional defects.
        (b) Different-strength perturbations apply the same modification with varying severity.
        (c) The same perturbation applied to different objects reveals the influence of object functional importance.
        }
    \label{fig:otc_perturbation}
\end{figure}

\AppendixSideFigure[1]{0.39}{
\includegraphics[width=\linewidth]{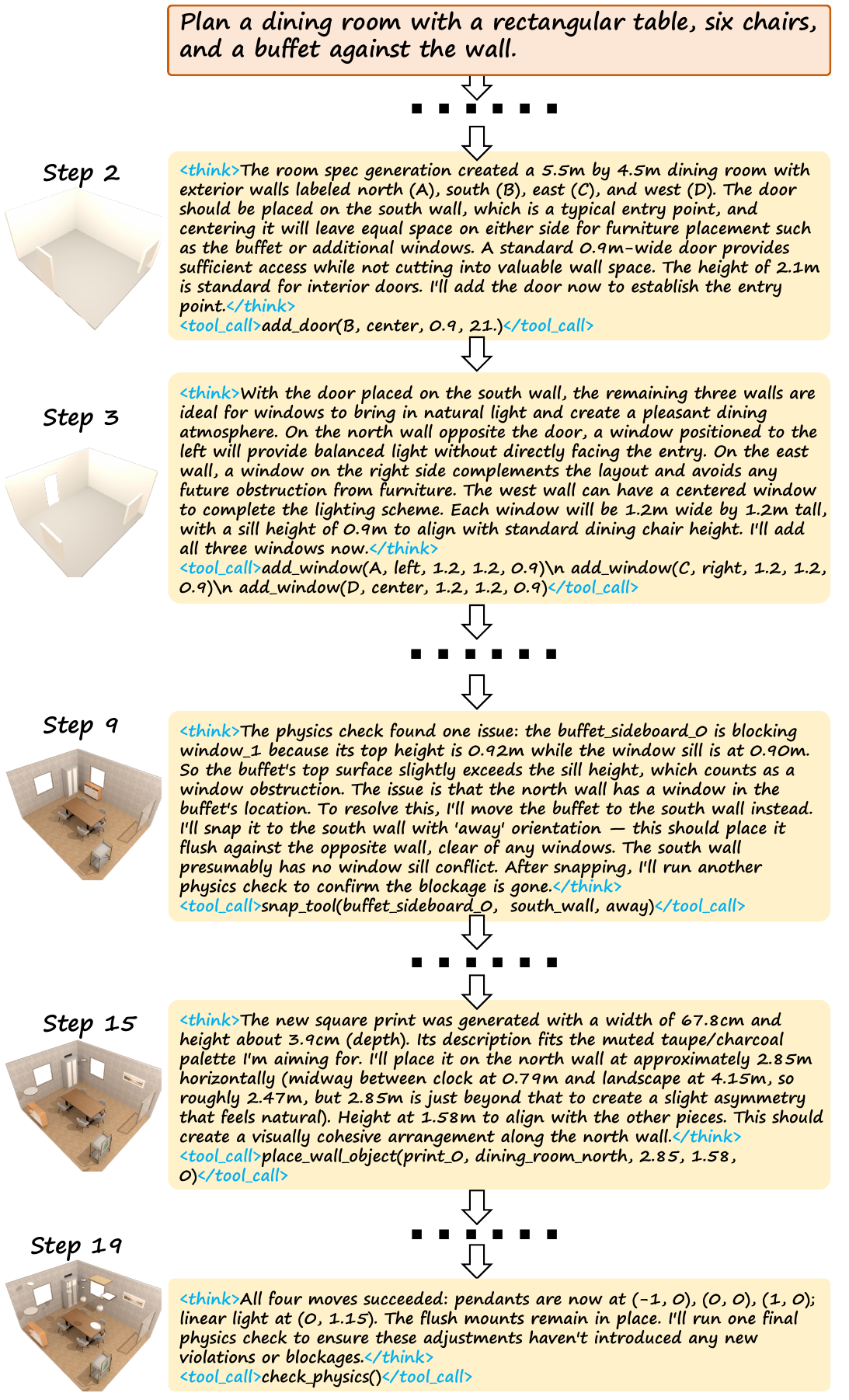}
    \captionof{figure}{
    A representative Scene-OTA trajectory.
    The trajectory records the complete scene construction process
    }
    \label{fig:sceneota_case}

}{
\AppendixSideParagraph{
\noindent\textbf{Perturbation Strategies.}
To construct preference examples with different levels of difficulty, we design three perturbation strategies that capture complementary aspects of scene quality.
}

\AppendixSideParagraph{
First, \textbf{accumulating perturbations} create ordered degradation chains by progressively adding new perturbations to an initially high-quality scene. Each subsequent scene inherits previous defects and introduces additional degradation, resulting in naturally ordered preferences from high to low quality.
}

\AppendixSideParagraph{
Second, \textbf{different-strength perturbations} apply the same modification to the same object with increasing severity levels. These examples teach the Critic to distinguish subtle quality differences, where mild perturbations may only weaken spatial coherence while stronger perturbations significantly damage functional relationships.
}

\AppendixSideParagraph{
Third, \textbf{same perturbation on different objects} applies identical modifications to objects with different functional roles. Since the impact of a perturbation depends on the affected object, these examples require the Critic to consider object importance and scene context rather than relying only on the perturbation type.
}

\AppendixSideParagraph{
Representative examples of the three perturbation strategies are shown in Fig.~\ref{fig:otc_perturbation}.
}

\AppendixSideParagraph{
In Fig.~\ref{fig:otc_perturbation}(a), \textbf{accumulating perturbations} construct an ordered degradation chain from a high-quality scene. Starting from the original layout, each subsequent scene introduces one additional perturbation while preserving previous defects. This creates naturally ranked scene variants, allowing the Critic to understand how accumulated errors progressively reduce overall scene quality.
}

\AppendixSideParagraph{
In Fig.~\ref{fig:otc_perturbation}(b), \textbf{different-strength perturbations} apply the same modification to the same object with different severity levels. For example, gradually moving an object away from its functional partner produces increasingly degraded layouts: a mild displacement may only weaken object relationships, whereas a stronger displacement can disrupt functional coherence. These examples encourage the Critic to capture fine-grained quality differences rather than treating perturbations as binary failures.
}

\AppendixSideParagraph{
In Fig.~\ref{fig:otc_perturbation}(c), \textbf{same perturbation on different objects} applies the same modification to objects with different functional importance. Although identical operations may be applied, their impact on scene quality varies depending on the object's role. For example, removing a primary functional object causes more severe degradation than removing a secondary decorative object. These examples require the Critic to reason about object roles and scene context when producing preferences.
}
}

\noindent\textbf{Dataset Statistics.}
Following the above construction process, we obtain Scene-OTC-15K, containing approximately 15K scene critique examples. Each example contains multiple scene renderings, a complete preference ranking, and a reasoning trace generated by GPT-5.2. In addition, we construct a two-scene comparison test set with 94 examples and a four-scene ranking test set with 55 examples for evaluating scene critique ability.

\noindent\textbf{Critique Capabilities Covered by Scene-OTC.}
The perturbation hierarchy in Scene-OTC is designed to enhance three complementary scene critique abilities. \textbf{Physical validity} focuses on recognizing explicit geometric and environmental violations, including collisions, boundary violations, and accessibility failures. \textbf{Functional correctness} requires understanding object roles, interaction orientations, and functional relationships, where a scene can remain physically valid but fail to support intended usage. \textbf{Layout coherence} captures higher-level scene organization, including spatial balance, object grouping, and global composition quality, where degradation may be subtle and difficult to identify from local object changes. By covering these three dimensions, Scene-OTC enables the Critic to evaluate scenes beyond simple validity checking and develop a more comprehensive understanding of scene quality.

\section{Case Study}
\label{app:case-study}

This section provides additional qualitative analyses to complement the quantitative results in the main paper. We first compare \textbf{Scene-Builder}, \textbf{Scene-Designer}, and \textbf{Scene-Designer with Critic} across representative room types, then examine how the learned Critic guides branch selection during inference, and finally present comparisons between \textbf{Scene-Designer} and existing methods. Together, these cases illustrate the effects of SFT, Scene-GRPO, and test-time Critic guidance on scene quality, construction behavior, and intermediate design decisions.

\subsection{Examples of Scene-Builder, Scene-Designer, and Scene-Designer with Critic}

\AppendixSideFigure[1]{0.60}{
\includegraphics[width=0.90\linewidth]{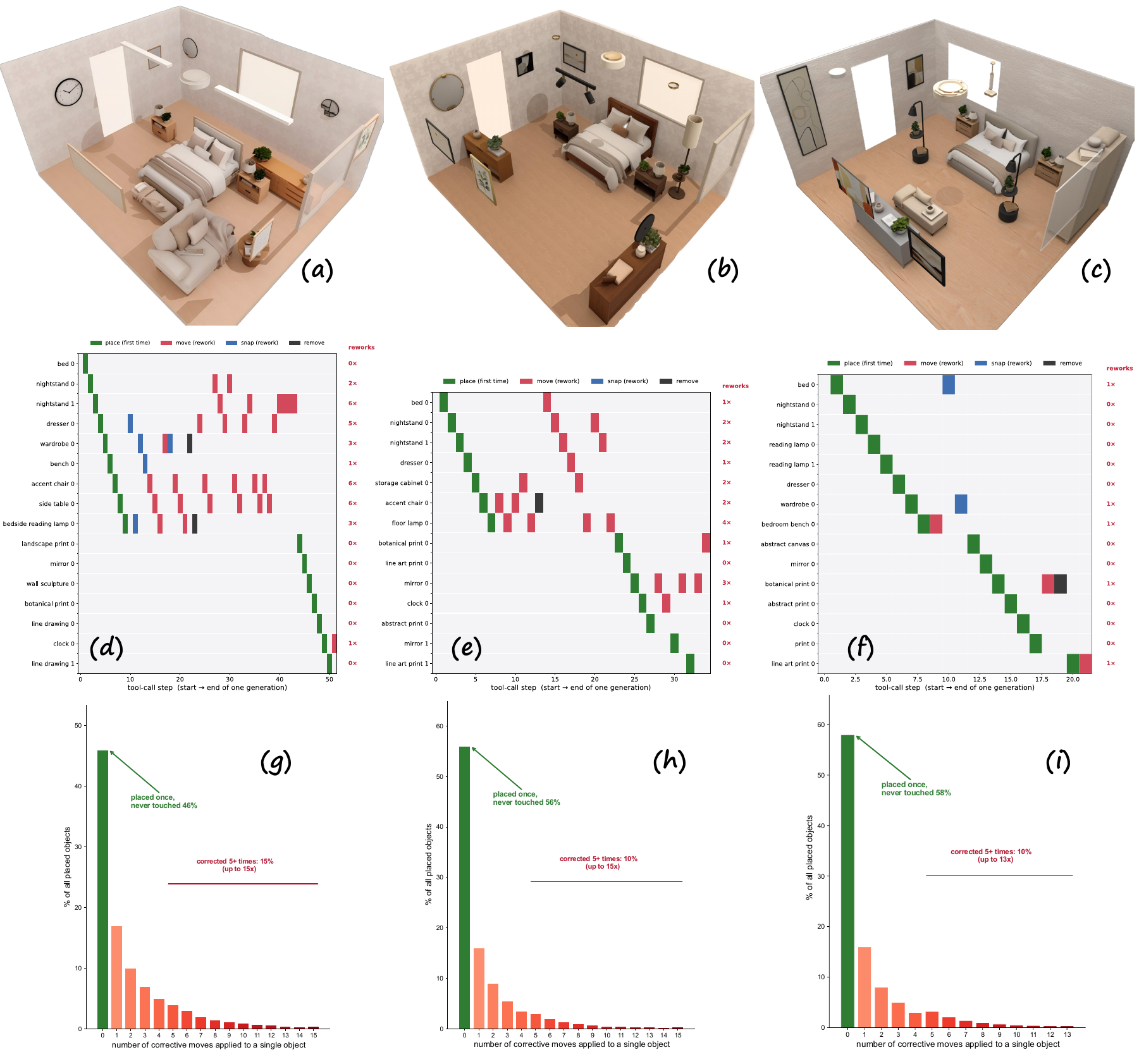}
    \captionof{figure}{Trajectory analysis under the instruction: “Design a serene master bedroom centered on a queen bed with matching nightstands and reading lamps, with a dresser placed against the opposite wall to create a calm and relaxing atmosphere.” (a,d,g), (b,e,h), and (c,f,i) correspond to Scene-Builder, Scene-Designer, and Scene-Designer w/ Critic, respectively. (a--c) Generated scenes under the same instruction. (d--f) Editing trajectories of major assets during construction. (g--i) Distributions of corrective edits per object over 100 generated scenes.}
    \label{fig:ffig5}
}{
\AppendixSideParagraph{
We provide additional qualitative comparisons to complement the quantitative results in the main paper. Each case compares \textbf{Scene-Builder}, \textbf{Scene-Designer}, and \textbf{Scene-Designer with Critic} under the same instruction, together with their corresponding construction trajectories. These examples illustrate how SFT provides basic scene construction ability, while RL improves functional organization and Critic guidance further enables more deliberate intermediate decision making.
}

\AppendixSideParagraph{
As shown in Fig.~\ref{fig:ffig5}(a-c), Scene-Builder can generate physically valid scenes but often lacks coherent functional organization, resulting in less structured layouts under the same instruction. While Scene-Builder can generate physically valid scenes, it often lacks consideration of functional correctness and layout coherence. In contrast, Scene-Designer and Scene-Designer w/ Critic produce scenes that better satisfy all three design requirements, including physical validity, functional correctness, and layout coherence, indicating that Scene-GRPO enables the model to better reason about design requirements before execution. We further analyze the editing trajectories of major assets in Fig.~\ref{fig:ffig5}(d--f). Scene-Builder frequently relies on trial-and-error corrections, where objects are repeatedly adjusted after suboptimal initial decisions. By learning sequential construction experience and optimizing editing choices through Critic-guided GRPO, Scene-Designer performs more deliberate modifications and avoids unnecessary revisions. With additional test-time Critic guidance, Scene-Designer w/ Critic further improves intermediate decision selection by filtering inferior editing alternatives before execution.
}

\AppendixSideParagraph{
Fig.~\ref{fig:ffig5}(g-i) summarizes the distribution of object-level corrective edits over 100 generated scenes. Compared with Scene-Builder, Scene-Designer increases the proportion of objects placed once and never modified from 46\% to 56\%, while reducing heavily corrected objects (more than five corrections) from 15\% to 10\%. Scene-Designer w/ Critic further improves these trends, achieving 58\% untouched objects and reducing heavily corrected cases to only 5\%. These results show that Critic-guided optimization does not simply shorten trajectories; instead, it enables the model to make more reliable design decisions before execution, leading to more efficient and coherent scene construction.
}
}

Additionally, we provide several more representative cases as follows.

\AppendixSideFigure[1]{0.53}{
\includegraphics[width=\linewidth]{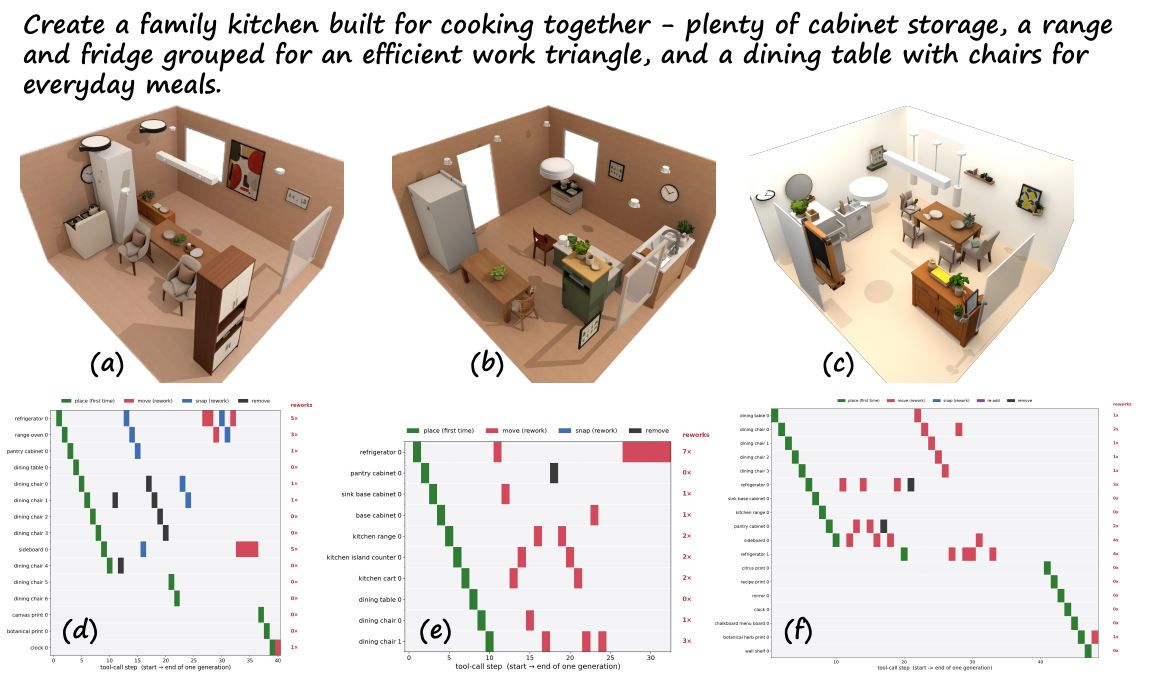}
    \captionof{figure}{
The evaluated instruction is: ``Create a family kitchen built for cooking together---plenty of cabinet storage, a range and fridge grouped for an efficient work triangle, and a dining table with chairs for everyday meals.''
Qualitative comparison across SFT, RL, and test-time Critic guidance.
(a,d), (b,e), and (c,f) correspond to Scene-Builder, Scene-Designer, and Scene-Designer with Critic, respectively.
}
    \label{fig:ffig2}
}{
\AppendixSideParagraph{
Fig.~\ref{fig:ffig2} presents a family kitchen example requiring cabinet storage, an efficient cooking workflow among the refrigerator, range, and preparation area, as well as a dining zone for everyday meals. Scene-Builder produces a physically valid kitchen and satisfies most object requirements, but the overall arrangement is relatively object-centric: the cooking components and dining furniture are less clearly organized into functional zones, and several assets require repeated adjustments during construction. After RL optimization, Scene-Designer generates a more structured layout, where the kitchen appliances, storage units, and dining table form clearer functional groups. Scene-Designer with Critic further improves the result by selecting better intermediate branches during construction, preserving circulation space and producing a more coherent relationship between cooking and dining areas. The trajectory comparison shows that Critic guidance helps avoid unnecessary corrections and leads to more purposeful refinement.
}
}

\AppendixSideFigure[1]{0.53}{
\includegraphics[width=\linewidth]{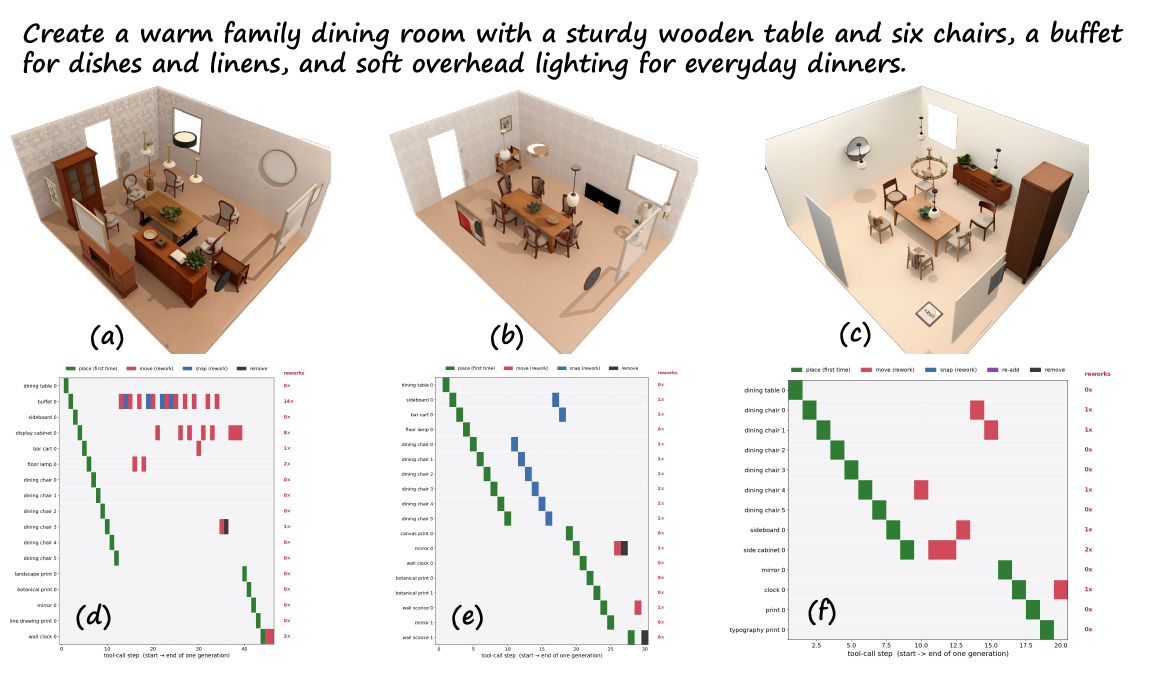}
    \captionof{figure}{
The evaluated instruction is: ``Create a warm family dining room with a sturdy wooden table and six chairs, a buffet for dishes and linens, and soft overhead lighting for everyday dinners.''
Qualitative comparison across SFT, RL, and test-time Critic guidance.
(a,d), (b,e), and (c,f) correspond to Scene-Builder, Scene-Designer, and Scene-Designer with Critic, respectively.
}
    \label{fig:ffig3}
}{
\AppendixSideParagraph{
Fig.~\ref{fig:ffig3} shows a family dining-room example requiring a wooden table with six chairs, a buffet for storage, and soft overhead lighting. Scene-Builder generates a reasonable dining environment but exhibits weaker functional grouping, with furniture elements lacking a clear relationship centered around the dining activity. Scene-Designer improves the spatial organization by placing the table, chairs, buffet, and lighting elements into a more coherent dining composition, better matching the intended family-meal scenario. With test-time Critic guidance, Scene-Designer further refines intermediate choices by favoring layouts with stronger functional connections and balanced spatial distribution. As reflected in the construction trajectories, Critic-guided generation requires fewer unnecessary modifications while maintaining a more consistent design process.
}
}

\begin{figure}[!htbp]
    \centering
    \includegraphics[width=0.82\columnwidth]{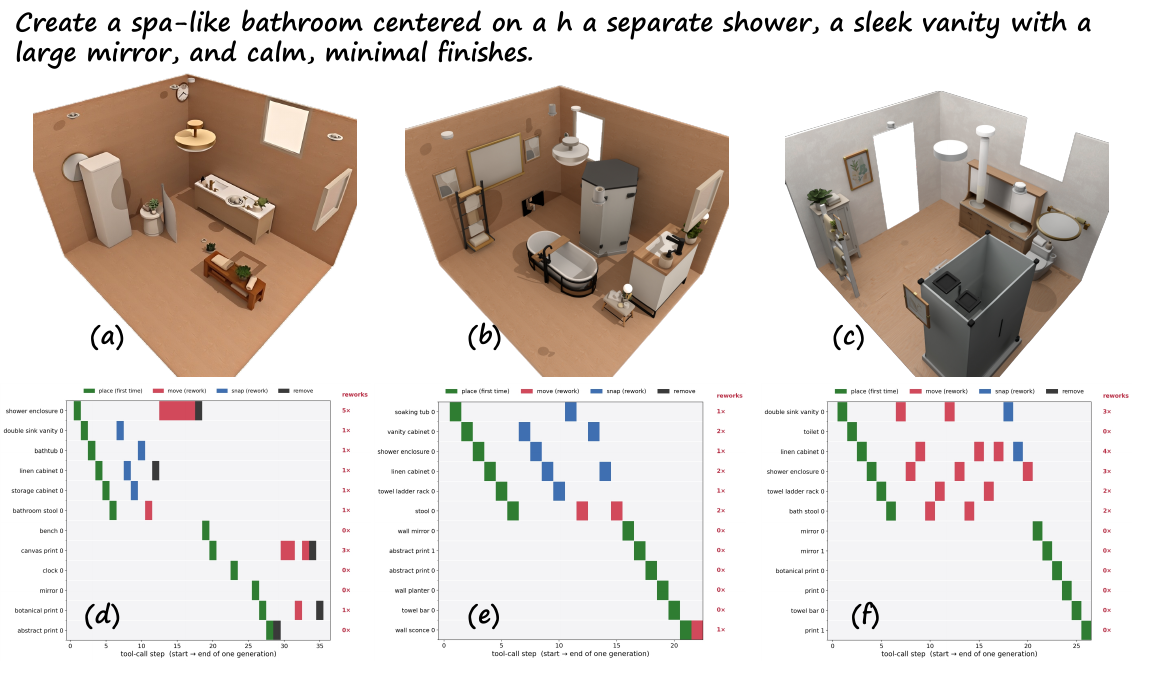}
    \caption{
The evaluated instruction is: ``Create a spa-like bathroom Create a spa-like bathroom with a separate shower, a sleek vanity with a large mirror, and calm, minimal finishes.''
Qualitative comparison across SFT, RL, and test-time Critic guidance.
(a,d), (b,e), and (c,f) correspond to Scene-Builder, Scene-Designer, and Scene-Designer with Critic, respectively.
}
    \label{fig:ffig4}
\end{figure}

Fig.~\ref{fig:ffig4} presents a spa-like bathroom example involving a separate shower, a vanity with a large mirror, and calm minimal finishes. Scene-Builder can construct the required bathroom elements, but the generated layout relies more on local placement decisions and may contain weaker relationships among major functional components. Scene-Designer improves the arrangement by learning higher-level spatial organization, creating clearer separation between the shower area, vanity zone, and remaining functional space. Scene-Designer with Critic further enhances the final layout by filtering less suitable construction alternatives, resulting in a more balanced and coherent bathroom design. This case demonstrates that Critic guidance is beneficial not only for avoiding physical errors but also for improving subtle functional and aesthetic decisions during scene creation.

\subsection{Test-Time Critic Guidance Examples}
\begin{figure*}[!htbp]
    \centering
    \includegraphics[width=0.88\textwidth]{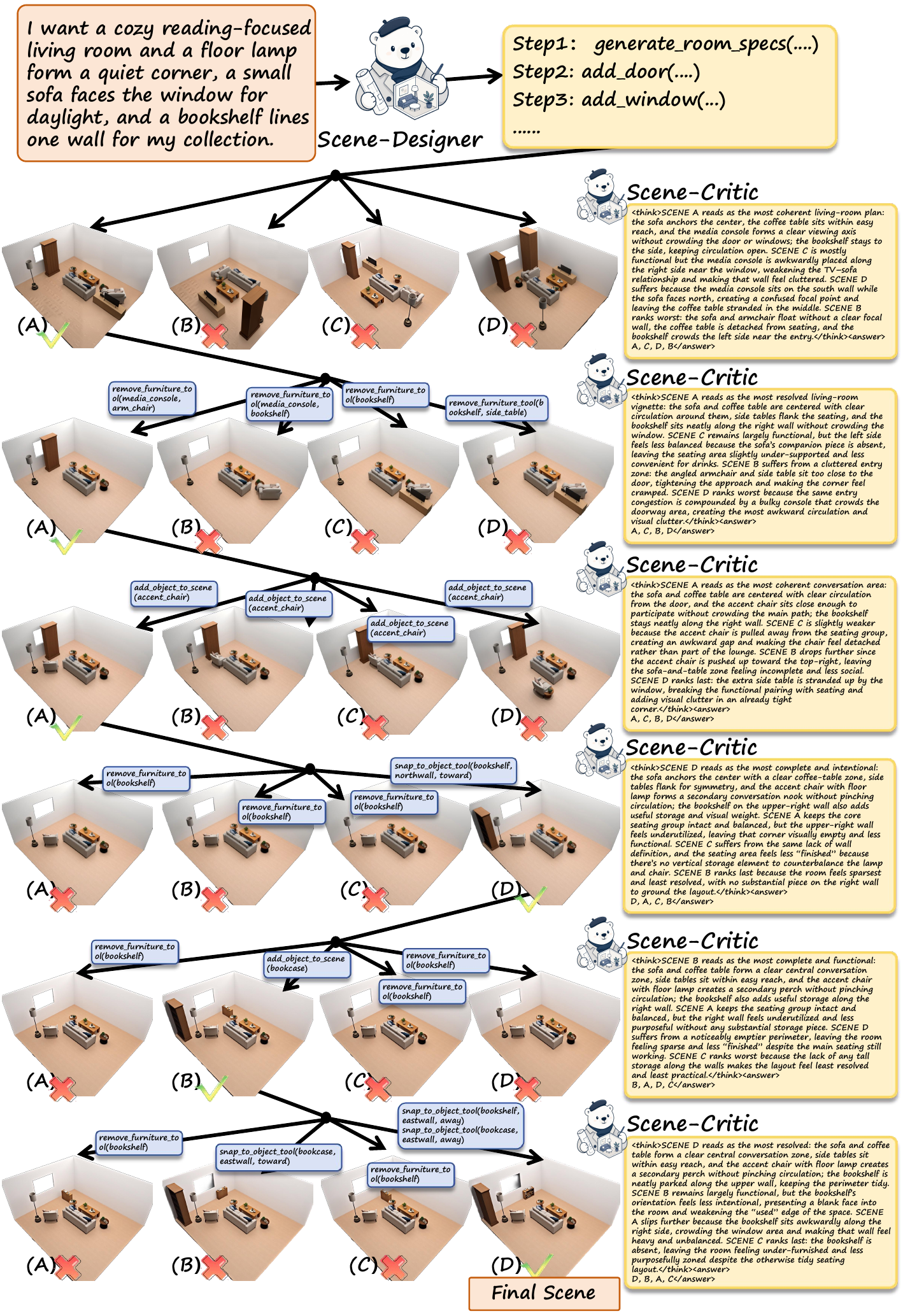}
    \caption{
        Test-time Critic guidance during scene construction.
        The Scene-Designer samples multiple candidate branches at each step, and the Scene-Critic ranks intermediate layouts based on physical validity, functional correctness, and layout coherence. The selected branches progressively refine the scene toward the final design.
        }
    \label{fig:ffig1}
\end{figure*}

To better understand how the learned Critic guides scene generation during inference, we provide a representative construction example in Fig.~\ref{fig:ffig1}. Given the instruction of creating a cozy reading-focused living room, the Scene-Designer policy performs iterative construction through multiple editing steps. At each step, instead of directly executing a single action, the policy samples multiple candidate branches with different object additions, removals, or modifications. The learned Critic then evaluates these intermediate layouts based on physical validity, functional correctness, and layout coherence, and selects the branch with the highest design quality for further construction.

In the first stage, the policy generates four candidate layouts after the initial construction. The Critic ranks Scene A as the preferred choice because it better satisfies the design requirements: the sofa serves as the central anchor, the coffee table remains within reach of the seating area, the media console maintains a clear viewing relationship, and the bookshelf is positioned without obstructing circulation. In contrast, other candidates exhibit weaker functional correctness and layout coherence. For example, Scene C places the media console near the window, weakening the relationship between the viewing area and seating zone, while Scene D creates a mismatched focal arrangement between the sofa and media console. Scene B further suffers from a disconnected seating arrangement, where the sofa, armchair, and coffee table fail to form a coherent functional group. By identifying these differences, the Critic guides the policy toward a better intermediate scene state.

As construction continues, the Critic repeatedly filters out inferior editing decisions and encourages modifications that improve the overall scene quality. In later steps, candidate branches involve operations such as removing unnecessary furniture, adding missing assets, or adjusting object configurations. The Critic does not simply favor scenes with more objects; instead, it evaluates whether each modification improves functional correctness and layout coherence while maintaining physical validity. For example, when selecting among branches with different furniture removals, the Critic prefers layouts that preserve essential functional elements, maintain balanced seating relationships, and avoid circulation conflicts. Similarly, when evaluating object additions, it favors branches where newly introduced furniture strengthens existing functional groups rather than creating isolated or redundant elements.

After several rounds of Critic-guided selection, the final scene achieves a more coherent reading-oriented living environment. The sofa, coffee table, accent chair, floor lamp, and bookshelf form meaningful spatial relationships, creating both a primary conversation area and a secondary reading corner. Compared with unguided trial-and-error exploration, this process avoids unnecessary corrections by selecting promising intermediate states before committing to further construction.

This example highlights that the learned Critic is not merely a reward estimator used during GRPO optimization. Instead, it captures transferable scene design preferences regarding physical validity, functional correctness, and layout coherence, allowing it to serve as an inference-time decision assistant. By evaluating alternative construction branches and guiding the policy toward better intermediate layouts, the Critic enables more deliberate and efficient scene generation.

\begin{figure}[!htbp]
\centering

\includegraphics[width=0.92\textwidth]{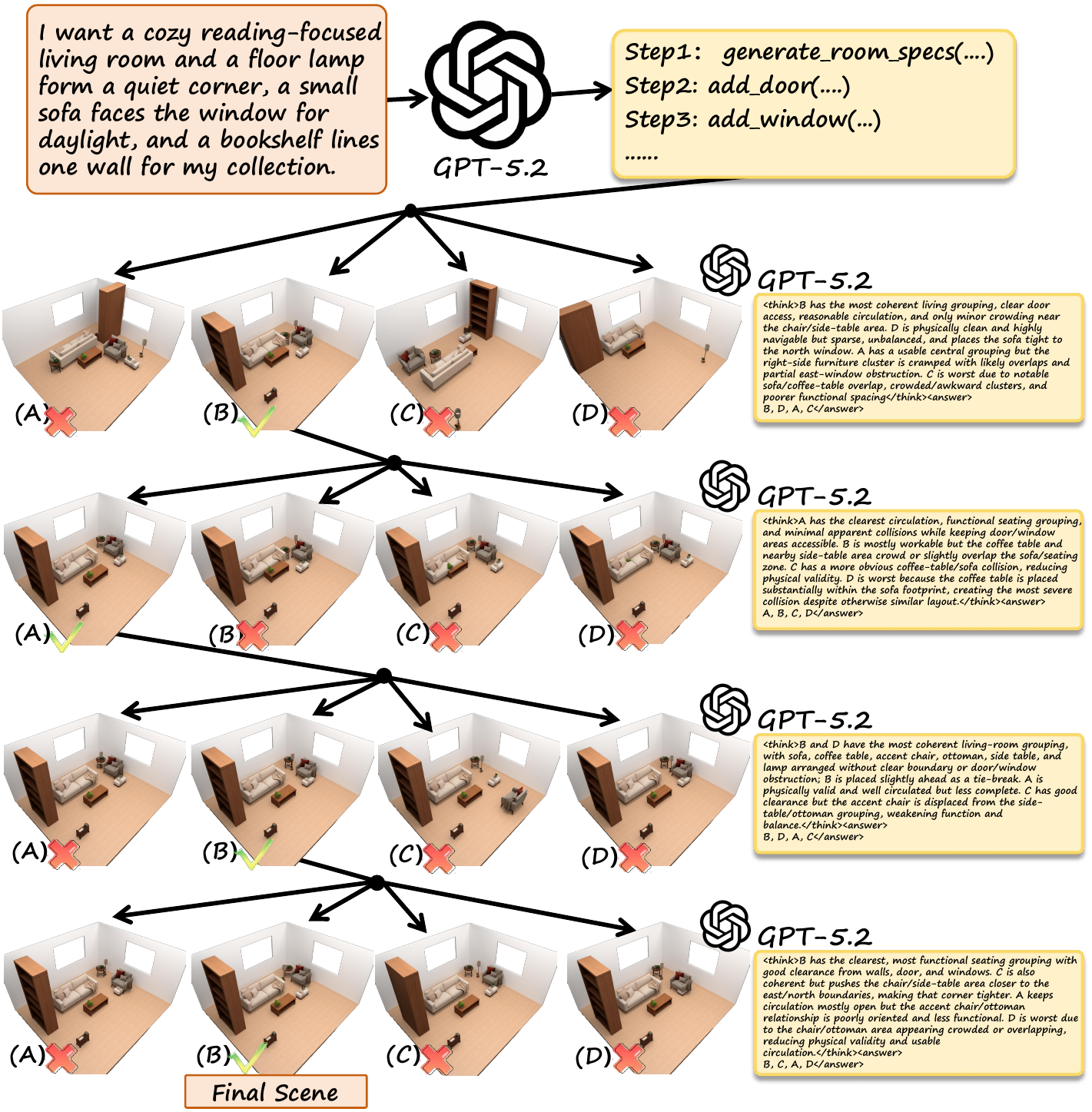}
    \caption{A representative failure case of GPT-5.2 as a test-time Critic during scene construction. GPT-5.2 often provides plausible rationales but fails to consistently evaluate physical validity, functional correctness, and layout coherence when selecting among candidate branches.}
    \label{fig:ffig13}

\end{figure}

In addition, Fig.~\ref{fig:ffig13} presents a representative example of using GPT-5.2 as a test-time critic for guiding scene construction. Although GPT-5.2 provides detailed rationales for each branch selection, its judgments are often inconsistent with the actual scene quality. In the first selection step, GPT-5.2 prefers Scene B by emphasizing a balanced seating arrangement and clear circulation. However, this judgment overlooks that Scene B provides weaker functional organization: the sofa, chair, and coffee table do not form as coherent a reading and conversation area as Scene A. Similarly, in later decisions, GPT-5.2 tends to focus on isolated visual cues, such as open space or apparent object alignment, while missing more important design factors including object relationships and functional usability.

In the third branch selection, GPT-5.2 selects Scene B as the best candidate while overlooking a clear physical validity issue: the table and floor lamp are placed in collision. In contrast, Scene C avoids this collision and maintains a more reasonable spatial arrangement. This indicates that GPT-5.2 may recognize general layout patterns but fails to consistently verify explicit geometric constraints when multiple factors compete. In the final branch selection, GPT-5.2 again prefers Scene B, although Scene C provides a better overall composition. Scene B places furniture too densely in the corner and reverses the functional orientation of the cabinet, weakening both layout coherence and functional correctness. These examples show that general-purpose models may produce plausible explanations but still lack reliable scene-specific criteria for evaluating physical validity, functional correctness, and layout coherence.

These observations reveal that general-purpose closed-source models lack stable evaluation criteria across the three key dimensions of scene quality. Although GPT-5.2 can generate reasonable explanations based on visual observations, its preferences may be inconsistent with actual design requirements: it can overlook explicit physical violations, focus on local object arrangements rather than functional relationships, or favor visually open layouts while ignoring global spatial organization. Therefore, directly using such models as test-time critics can provide unreliable guidance, causing the policy to select inferior intermediate states and limiting the effectiveness of inference-time scene optimization. This motivates the need for a domain-specific Critic trained with structured scene-quality supervision to provide more consistent guidance during scene construction.

\begin{figure*}[!htbp]
    \centering
    \includegraphics[width=\textwidth]{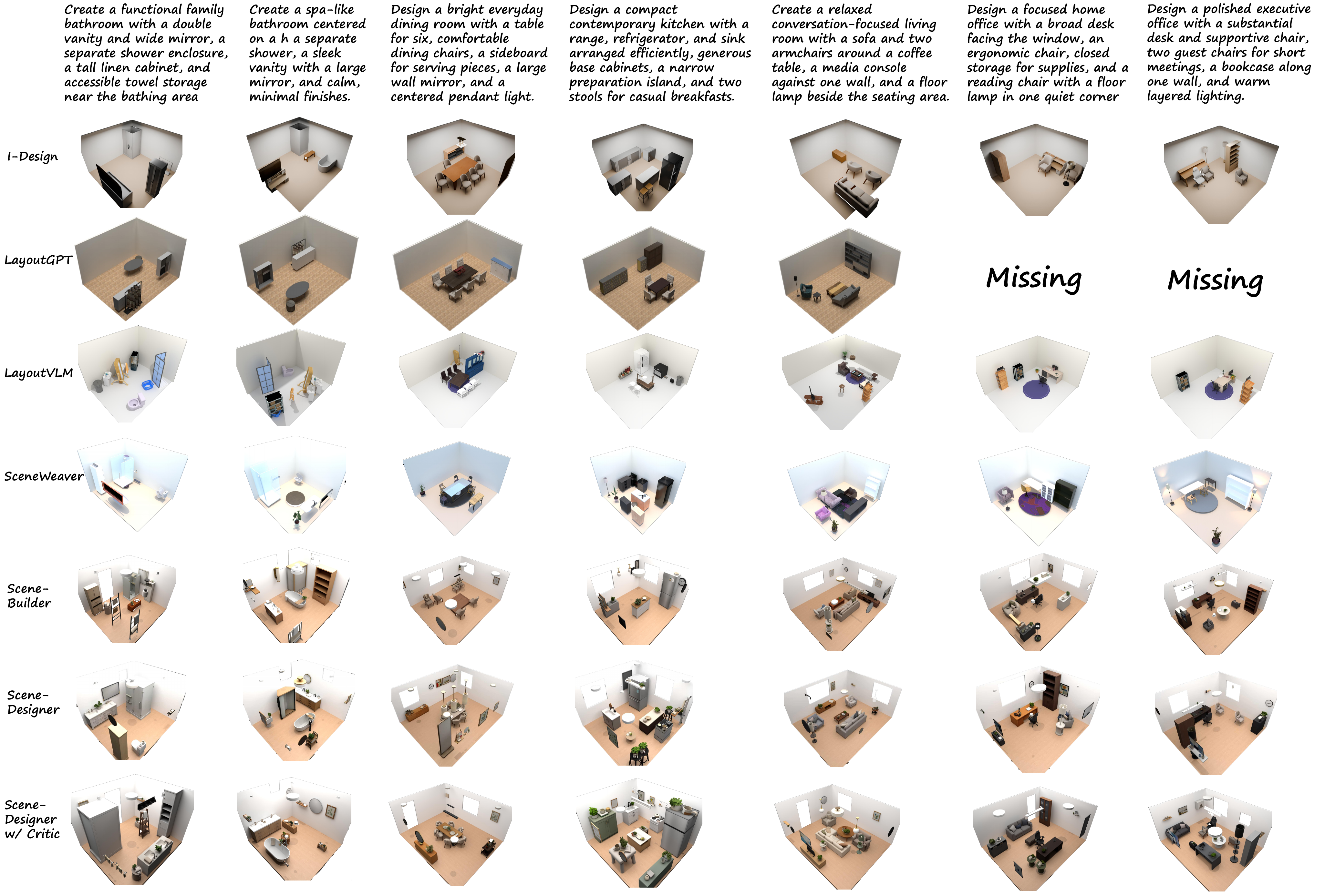}
    \caption{
    Qualitative comparison between Scene-Designer and existing scene generation methods.
    Existing methods often generate incomplete scenes or suffer from physical violations, poor functional organization, and incoherent layouts.
    Scene-Builder produces physically valid scenes, while Scene-Designer further improves functional correctness and layout coherence through learned design strategies. 
    Adding the Critic during inference provides additional guidance and further enhances overall scene quality.
    }
    \label{fig:ffig15}
\end{figure*}

\subsection{Examples of Scene-Designer and Other Methods}
Figure~\ref{fig:ffig15} presents representative comparisons between our Scene-Designer and existing scene generation methods across diverse indoor environments, including bathrooms, dining rooms, kitchens, living rooms, and offices. 
Existing methods often struggle to generate complete and functionally coherent scenes. 
For example, LayoutGPT and LayoutVLM frequently produce incomplete environments or only arrange a subset of major furniture while missing important structural elements such as walls, windows, mirrors, wall-mounted objects, and lighting fixtures. 
In several cases, generated layouts also contain obvious physical issues, including object collisions, unreasonable object density, and insufficient circulation space. 
For instance, in the dining room and living room cases, some methods place furniture without considering functional grouping, resulting in isolated objects and poorly organized activity areas. 
Similarly, in kitchen and office scenarios, objects are often arranged without considering practical usage relationships, leading to layouts that are physically plausible but functionally ineffective.

Compared with these approaches, Scene-Builder consistently generates complete scenes with valid room structures and satisfies basic physical constraints. 
However, as shown in the examples, its layouts may still lack higher-level design considerations, such as optimal functional organization and spatial balance. 
After applying Scene-GRPO, Scene-Designer further improves scene quality by learning effective construction and correction strategies from design trajectories. 
It produces more coherent arrangements, such as grouping functionally related objects, maintaining appropriate interaction spaces, and balancing furniture distribution within the room. 
Moreover, incorporating the learned Critic during inference further enhances the generated scenes by providing additional quality guidance, leading to improvements across physical validity, functional correctness, and layout coherence.

\section{Prompts}

\subsection{System Prompts}
We adopt the scene-construction scaffold introduced by SceneSmith~\cite{pfaff2026scenesmithagenticgenerationsimulationready}. Since the scaffold provides a large number of tools for different construction operations, following SceneSmith, we decompose the generation process into multiple stages: \ding{182} defining the overall room structure, including room dimensions, door and window placement, and material settings; \ding{183} placing large floor-standing assets; \ding{184} adding wall-mounted assets; and \ding{185} adding ceiling-mounted assets. During inference, each stage is associated with a different system prompt that specifies its current construction objective and available operations. When the model produces only a reasoning trace without an executable tool call, we interpret this as the completion of the current stage. In the following step, the system prompt is replaced with the predefined prompt for the next stage. Although the construction process is divided into multiple stages, we treat all stages as a unified generation process and record them as a single Observation--Thought--Action trajectory. We next introduce the system prompt used at each stage.

\AppendixSideFigure[1]{0.56}{
\includegraphics[width=0.8\linewidth]{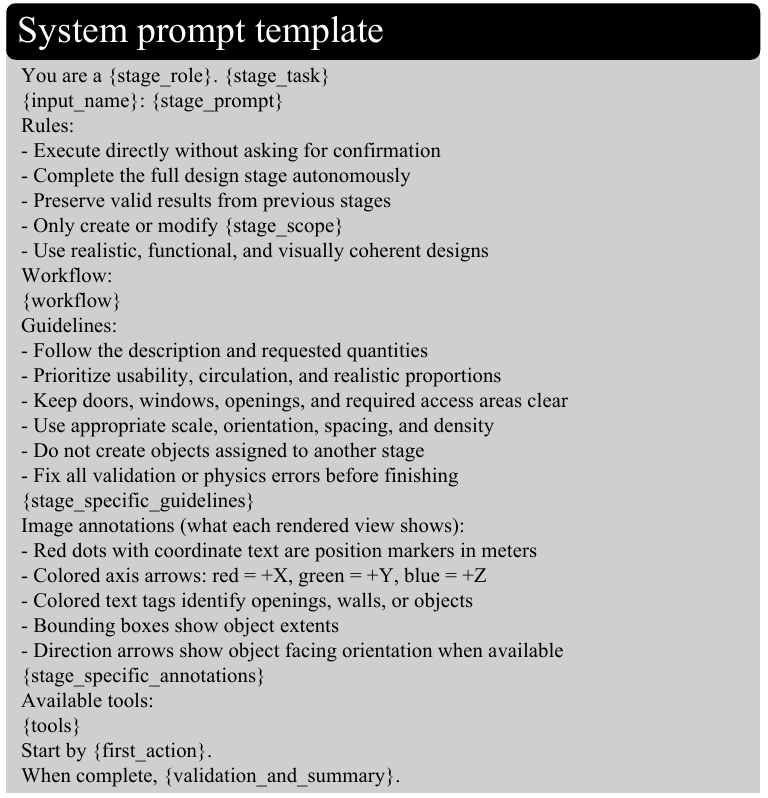}
    \captionof{figure}{System prompt template.}
    \label{fig:prompt1}
}{
\AppendixSideParagraph{
The system prompts follow a unified template across different construction stages, where only stage-specific fields are replaced according to the current construction objective. Specifically, \texttt{{stage\_role}} defines the role of the model in the current stage, \texttt{{stage\_task}} and \texttt{{stage\_prompt}} specify the construction objective, and \texttt{{stage\_scope}} restricts the assets that can be created or modified. The \texttt{{workflow}} field provides the recommended construction procedure, while \texttt{{stage\_specific\_guidelines}} and \texttt{{stage\_specific\_annotations}} incorporate stage-dependent constraints and rendering interpretations. The available tools are also dynamically specified according to each stage. Therefore, although different stages use different system prompts, they share the same overall structure and only differ in stage-specific configurations.
}
}

\vspace{-0.75\baselineskip}
\subsection{Memory Snapshot}

\AppendixSideFigure{0.5}{
\includegraphics[width=\linewidth]{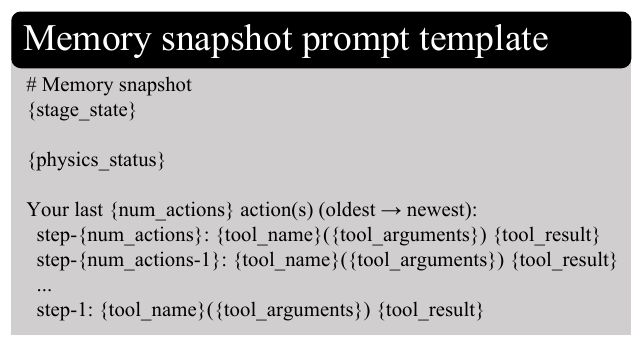}
    \captionof{figure}{Unified memory snapshot template used by the policy model across different construction stages.}
    \label{fig:memory_snapshot}
}{
\AppendixSideParagraph{
In our framework, $M_t$ denotes a structured memory snapshot provided to the policy at each reasoning step. Instead of storing the complete interaction trajectory, $M_t$ summarizes the essential information required for continued scene construction. As shown in Fig.~\ref{fig:memory_snapshot}, the memory snapshot contains three components: the current stage state, physical validation status, and recent tool interactions.
}

\AppendixSideParagraph{
The \texttt{stage\_state} records the scene information relevant to the current construction stage. Its content changes according to the stage objective: the floor-plan stage describes the room structure, doors, windows, and materials; the furniture stage describes placed assets and their configurations; and the wall and ceiling stages record the corresponding mounted objects and placement states. The \texttt{physics\_status} provides the latest environment validation results, including detected physical violations that require correction. The final component maintains the most recent tool interactions, including executed tool calls and their returned results, allowing the model to track recent decisions without repeatedly revisiting the full trajectory history.
}
}

At each step, the rendered memory snapshot is provided together with the system prompt, task instruction, and current scene observation. After a successful tool execution, the memory is updated according to the returned tool result:
\begin{equation}
M_{t+1}=Update(M_t,a_t,f_t),
\end{equation}
where $a_t$ denotes the executed tool call and $f_t$ denotes the corresponding tool feedback.

\subsection{Correction-Effect Evaluation Prompt}
For correction-effect evaluation, we provide GPT-5.2 with a structured JSON input describing each candidate correction segment. The placeholder \texttt{\{SEGMENT\_JSON\}} is dynamically replaced by the candidate segment data extracted from the original trajectory. It contains segment-level information, operation-level states, geometric progress signals, feedback changes, and object lifecycle context. Each operation record includes the executed tool call and arguments, object identity, scene states before and after the operation, final accepted state, final-state progress, and screening signals. The feedback fields record error changes from environment validation and semantic revision feedback, while the object lifecycle context provides nearby operations to identify potential dependencies.

Given this structured input, GPT-5.2 predicts an operation-level label:

\begin{equation}
y_i \in
\left\{
\begin{aligned}
&\texttt{redundant\_trial},\\
&\texttt{useful\_correction},\\
&\texttt{uncertain}.
\end{aligned}
\right.
\end{equation}

\begin{figure}[!t]
    \centering
    \includegraphics[width=0.71\columnwidth]{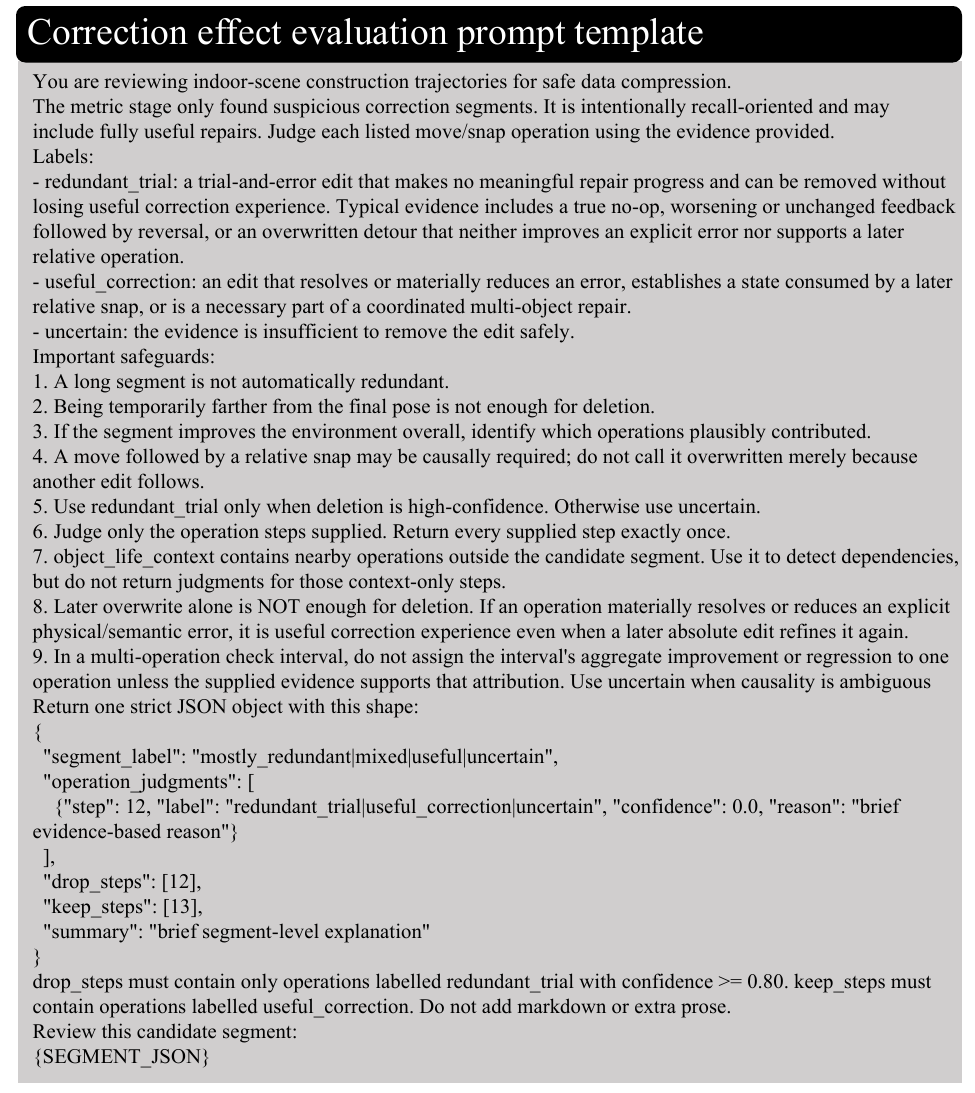}
    \caption{Prompt template for correction-effect evaluation. The verifier receives structured trajectory evidence and outputs operation-level causal judgments.}
    \label{fig:Correctioneffecttemplate}
\end{figure}

The output is formatted as a strict JSON object containing the segment-level judgment, operation-level labels, confidence scores, and brief causal explanations. The complete prompt template is shown in Fig.~\ref{fig:Correctioneffecttemplate}.

\section{Additional Experiments}
Due to space limitations in the main paper, we present several additional experiments and analyses in this section to further evaluate our method from complementary perspectives.

\subsection{Quantitative Evaluation of Scene Design Quality}

\begin{table*}[!h]
\centering
\small
\setlength{\tabcolsep}{5pt}
\caption{Quantitative evaluation of scene accessibility and spatial organization. Higher values are better for RR, mRR, FR, DAR, and WAR. NAR measures the proportion of navigable area and is not a strictly monotonic quality metric.}
\label{tab:objective_scene_metrics}
\resizebox{\textwidth}{!}{%
\begin{tabular}{lcccccccc}
\toprule
Method  & RR@0.2 $\uparrow$ & RR@0.3 $\uparrow$ & RR@0.5 $\uparrow$ & mRR $\uparrow$ & FR@0.3 $\uparrow$ & NAR@0.3 & DAR $\uparrow$ & WAR $\uparrow$ \\
\midrule
Scene-Builder & 0.92 & 0.98 & 0.89 & 0.96 & 0.72 & 0.53 & \textbf{1.0} & 0.96 \\
Scene-Designer & 0.97 & \textbf{1.0} & 0.93 & 0.96 & 0.83 & 0.53 & \textbf{1.0} & 0.96 \\
Scene-Designer w/ Critic& \textbf{1.0} & \textbf{1.0} & \textbf{0.94} & \textbf{0.98} & \textbf{1.0} & \textbf{0.54} & \textbf{1.0} & \textbf{1.0} \\
\bottomrule
\end{tabular}%
}
\end{table*}

Beyond the GPT-based evaluation and basic physical metrics reported in the main experiments, we further evaluate the three variants of our framework using deterministic geometry-based metrics. These metrics are computed directly from room geometry and final asset placements without using GPT or rendered images, providing a more objective assessment of scene accessibility and spatial organization.

\noindent\textbf{Reachability Ratio.}
This metric evaluates whether the free space in a scene is well connected. Intuitively, it asks: if an agent starts from the main open area of the room, how much of the remaining free space can it reach without passing through furniture or walls?

For an agent with width $w$, we shrink the room boundary by $w/2$ and expand each furniture footprint by $w/2$, so that the agent is treated as having a non-zero body width rather than as a point. Let $\mathcal{F}_w$ denote the resulting free space, and let $\mathcal{F}_w^{\max}$ denote its largest connected component. We define

\begin{equation}
\mathrm{RR}@w
=
\frac{\operatorname{Area}(\mathcal{F}_w^{\max})}
{\operatorname{Area}(\mathcal{F}_w)}.
\end{equation}

A high RR means that most free space belongs to one connected region, whereas a low RR indicates that furniture divides the room into isolated or difficult-to-reach areas. To reduce sensitivity to a single agent width, we additionally report

\begin{equation}
\mathrm{mRR}
=
\frac{
\mathrm{RR}@0.2+
\mathrm{RR}@0.3+
\mathrm{RR}@0.5
}{3}.
\end{equation}

The three widths represent different assumptions about the space required for movement, and their average provides a more stable measure of overall reachability.

\noindent\textbf{Full Reachability.}
While RR measures the degree of connectivity, Full Reachability provides a stricter binary test. It asks whether all free space in the room forms exactly one connected region, without any isolated pockets.

Let $K_w$ denote the number of connected components in $\mathcal{F}_w$. We define

\begin{equation}
\mathrm{FR}@w
=
\begin{cases}
1, & K_w=1,\\
0, & K_w\neq 1.
\end{cases}
\end{equation}

For an individual scene, FR is either 1 or 0. An FR value of 1 means that all free space is mutually reachable, while 0 indicates that at least one disconnected region exists. For a group of scenes, FR is reported as the proportion of scenes satisfying full connectivity.

\noindent\textbf{Navigable Area Ratio.}
This metric measures how much of the original room remains available for movement after accounting for furniture placement and agent width. In simple terms, it reflects how open or crowded the scene is.

Let $A_{\mathrm{floor}}$ denote the original floor area. We define

\begin{equation}
\mathrm{NAR}@w
=
\frac{\operatorname{Area}(\mathcal{F}_w)}
{A_{\mathrm{floor}}}.
\end{equation}

A higher NAR indicates that a larger portion of the room is navigable, whereas a lower value suggests a denser or more constrained layout. However, NAR is not a strictly monotonic scene-quality metric, because an under-furnished or nearly empty room may obtain a high NAR despite failing to satisfy the design instruction. Therefore, we use NAR mainly to characterize spatial congestion rather than overall scene quality.

\noindent\textbf{Door Accessibility Rate.}
DAR evaluates whether doors remain usable after furniture placement. It asks a simple question: how many doors still have an unobstructed clearance area?

Let $N_D$ denote the total number of doors and $N_D^{\mathrm{clear}}$ denote the number of doors whose clearance regions are not occupied by furniture. We define

\begin{equation}
\mathrm{DAR}
=
\frac{N_D^{\mathrm{clear}}}{N_D}.
\end{equation}

A higher DAR indicates that the generated layout better preserves entry, exit, and circulation around doors.

\noindent\textbf{Window Accessibility Rate.}
WAR evaluates whether windows remain accessible rather than being blocked by tall furniture. It measures the proportion of windows whose clearance regions are not occupied by furniture extending above the corresponding window sill.

Let $N_W$ denote the total number of windows and $N_W^{\mathrm{clear}}$ denote the number of accessible windows. We define

\begin{equation}
\mathrm{WAR}
=
\frac{N_W^{\mathrm{clear}}}{N_W}.
\end{equation}

Furniture below the window sill is not treated as an obstruction, allowing reasonable arrangements such as low cabinets or benches beneath windows. A higher WAR indicates better preservation of window access and the surrounding usable space.

All metrics are computed independently for each scene and then macro-averaged over the evaluation set, so that every scene contributes equally regardless of its room size or number of assets. The same geometric rules, thresholds, and agent widths are used for all methods.

\noindent\textbf{Analysis of Results.}
The results demonstrate that the proposed training strategy consistently improves the geometric quality of generated scenes. Scene-Designer already achieves strong accessibility performance compared with Scene-Builder, with higher RR@0.3, RR@0.5, and FR@0.3, indicating that Scene-GRPO enables the model to better organize objects and preserve connected free space during construction. After introducing the Critic, Scene-Designer w/ Critic further improves the overall spatial organization, achieving the highest RR across different agent widths, the best mRR (0.98), complete full reachability (FR@0.3=1.0), and the highest navigable area ratio. These results suggest that Critic-guided optimization helps the model avoid locally plausible but spatially inefficient arrangements and select layouts with better global accessibility.

The door and window accessibility metrics further show that all proposed models maintain strong environmental constraints. All methods achieve perfect DAR, while Scene-Designer w/ Critic additionally achieves the highest WAR, indicating that Critic guidance helps preserve accessibility around important environmental structures such as windows. Overall, these geometry-based metrics provide complementary evidence that Scene-Designer and Scene-Designer w/ Critic generate scenes with improved spatial organization and accessibility, beyond the GPT-based quality evaluations and basic collision statistics reported in the main experiments.

\end{document}